\documentclass{article}

\usepackage[final]{neurips_2022}

\usepackage[utf8]{inputenc} 
\usepackage[T1]{fontenc}    
\usepackage{hyperref}       
\usepackage{url}            
\usepackage{booktabs}       
\usepackage{amsmath}
\usepackage{amsfonts}       
\usepackage{nicefrac}       
\usepackage{microtype}      
\usepackage{xcolor}         

\usepackage{amssymb,amsmath,mathtools}
\usepackage{amsfonts}
\usepackage{bm,dsfont}
\usepackage{url}
\usepackage{pifont}

\usepackage{xspace}
\usepackage{stmaryrd}
\usepackage{graphicx}
\usepackage{subcaption}
\usepackage[mathcal]{eucal}
\usepackage{enumerate}
\usepackage{float}
\usepackage{booktabs}
\usepackage{tikz-cd} 
\usepackage{wrapfig}
\usepackage{tabularx}

\hypersetup{
  colorlinks   = true, 
  urlcolor     = cyan, 
  linkcolor    = blue, 
  citecolor    = blue, 
}

\usepackage{amsthm}
\theoremstyle{plain}
\newtheorem{theorem}{Theorem}[section]
\newtheorem{proposition}[theorem]{Proposition}
\newtheorem{lemma}[theorem]{Lemma}

\newtheorem{definition}[theorem]{Definition}

\newcommand{\BEAS}{\begin{eqnarray*}}
\newcommand{\EEAS}{\end{eqnarray*}}
\newcommand{\BEA}{\begin{eqnarray}}
\newcommand{\EEA}{\end{eqnarray}}
\newcommand{\BEQ}{\begin{equation}}
\newcommand{\EEQ}{\end{equation}}
\newcommand{\BIT}{\begin{itemize}}
\newcommand{\EIT}{\end{itemize}}
\newcommand{\BNUM}{\begin{enumerate}}
\newcommand{\ENUM}{\end{enumerate}}
\newcommand{\BA}{\begin{array}}
\newcommand{\EA}{\end{array}}

\usepackage{todonotes}

\newcommand{\RevDiff}[1]{#1}
\newcommand{\FinDiff}[1]{#1}

\usepackage{bm,dsfont}

\newcommand{\argmin}{\mathop{\rm argmin}}
\newcommand{\argmax}{\mathop{\rm argmax}}

\renewcommand{\d}{\,\mathrm{d}} 

\newcommand{\Loja}{\L{}ojasiewicz}

\def \R{{\mathbb R}}

\newcommand{\inner}[1]{\langle \cdot, \cdot \rangle_{#1}}

\newcommand{\overbar}[1]{#1^\star}

\DeclareMathOperator{\sinc}{sinc}

\title{Convergence beyond the over-parameterized regime using Rayleigh quotients}

\author{%
  David A. R. Robin\\
  INRIA\\
  École Normale Supérieure\\
  PSL Research University \\
  \texttt{david.a.r.robin@gmail.com}\\
  \And
  Kevin Scaman\\
  INRIA\\
  École Normale Supérieure\\
  PSL Research University \\
  \texttt{kevin.scaman@inria.fr}\\
  \And
  Marc Lelarge\\
  INRIA\\
  École Normale Supérieure\\
  PSL Research University \\
  \texttt{marc.lelarge@ens.fr}\\
}

\begin{document}

\maketitle

\begin{abstract}
\looseness=-1
In this paper, we present a new strategy to prove the convergence of deep learning architectures to a zero training (or even testing) loss by gradient flow. Our analysis is centered on the notion of Rayleigh quotients in order to prove
Kurdyka-\Loja{} inequalities for a broader set of neural network architectures and loss functions.
We show that Rayleigh quotients provide a unified view for several convergence analysis techniques in the literature. Our strategy produces a proof of convergence for various examples of parametric learning. In particular, our analysis does not require the number of parameters to tend to infinity, nor the number of samples to be finite, thus extending to test loss minimization and beyond the over-parameterized regime.

\end{abstract}

\section{Introduction}

In order to understand the performance of vastly over-parameterized networks, various works have investigated the properties of neural tangent kernels (NTK, see \citealp{jacot2018neural}) and their eigenspaces.
While the study of these spectra has led to proofs of convergence to global minima despite the non-convexity of the problem, these analyses typically rely on an over-parameterization assumption, or even infinite-width limits, casting a shadow on their applicability.
Positive-definiteness of the NTK in particular, granted by the infinite-width limit, does not hold with finite width and a growing number of samples, despite observed successes of neural networks in this regime.
We provide a (toy) counter-example in dimension two to better outline this issue, and fix this flaw by re-centering the discussion on Rayleigh quotients\RevDiff{, corresponding to fixed directions,} rather than \RevDiff{positive definiteness, i.e.\ uniformly bounding in all directions}.
We give several ideas to obtain bounds on Rayleigh quotients, and provide non-trivial examples for each of the presented ideas, including a recovery of known results, but also a new convergence speed guarantee for the multi-class logistic regression.

\paragraph{Overview.}
In a typical supervised learning task, one is given a training dataset of $n \in \mathbb{N}$ labeled samples $\mathcal{D} = ((x_i, y_i) \in \mathbb{R}^d \times \mathbb{R})_{i \in [n]}$, and a parametric model with $m \in \mathbb{N}$ parameters, ${f : \mathbb{R}^m \times \mathbb{R}^d \to \mathbb{R}}$.
The task is to find parameters fitting the training data, i.e.\ find $\theta^* \in \mathbb{R}^m$ such that $\forall i \in [n], f(\theta^*; x_i) \approx y_i$.
Aggregating these into a single vector $F : \theta \mapsto f_\theta = (f(\theta; x_i))_{i \in [n]}$, this becomes a satisfaction of a system of equations $F(\theta) \approx y \in \mathbb{R}^n$.
After choosing a functional loss $\ell : \mathbb{R}^n \to \mathbb{R}_+$, one can learn the associated parameters by gradient flow $\partial_t \theta = - D F(\theta)^T \cdot \nabla \ell(F(\theta))$, where the jacobian of the parameterization $F$ is a matrix $D F(\theta) \in \mathbb{R}^{n \times m}$.
This corresponds exactly to the usual practice of defining a parametric function $F$, a functional loss $\ell$, and training by gradient flow on the parameters to minimize the parametric loss $\mathcal{L} = \ell \circ F$.
The question is then when does this algorithm converge, and how fast ?
Our focus is on the regime of finitely many parameters ($m \in \mathbb{N}$) and large data ($n \to +\infty$)
, where the over-parameterization arguments ($m \gg n$) are insufficient. 


\paragraph{Context.}
Early arguments for the proof of convergence of this system to a loss of zero revolved around strong convexity hypotheses on the loss \citep[see][Section 9.3.1]{boyd04convex}. However
the parameterization $F$, typically as a neural network,
 leads to non-convex parametric losses $\mathcal{L}$ even when the functional loss $\ell$ is convex, sometimes even parametric losses that are not locally quasi-convex \citep[for details, see][]{liu2022loss}.
Recently, a common solution has been the leverage of Polyak-\Loja{} inequalities
$\lVert \nabla \mathcal{L}(\theta) \rVert_2^2 \geq \mu \, \mathcal{L}(\theta)$,
which grant linear convergence by integrating with Grönwall's lemma
since for gradient flows it holds $- \partial_t \mathcal{L}(\theta) = \lVert \nabla \mathcal{L} (\theta) \rVert_2^2$
$(\text{thus for}\, \mu \in\mathbb{R}_+^*, - \partial_t \mathcal{L}(\theta) \geq \mu \, \mathcal{L}(\theta) \Rightarrow \mathcal{L}(\theta_t) \leq \mathcal{L}(\theta_0) \, \exp(- \mu t))$.
For examples in continuous time, see \citet[Theorem 3.3 and 3.4]{chizat20:hal-sparseoptimization}.
Other results with discrete time include
\citet[Theorem 4.1]{arora19finegrained},
\citet[Theorem 2.1]{oymak19overparameterized},
\citet[Theorem 5.1]{liu2020linearity} and
\citet[Eq (3)]{liu2022loss}
.
Generally speaking, discretized versions with sufficiently small learning rate have very similar dynamics, at the cost of some local smoothness assumption, and similarly, stochastic versions can leverage the same \Loja{} inequalities to prove convergence rates, so the continous-time dynamics proof can be viewed as a first step in the analysis of these more complex cases.
These inequalities ensure that there are no critical points that are not global minima, and can hold
even for non-convex losses $\mathcal{L}$, although they can be hard to prove.

The behavior of the dynamical system $\partial_t \theta = - \nabla \mathcal{L}(\theta)$ 
has been shown to be closely tied with the eigenspaces of the Neural Tangent Kernel (NTK) matrix $K(\theta) = D F(\theta) \cdot D F(\theta)^T \in \mathbb{R}^{n \times n}$, introduced in \citet[Section~4]{jacot2018neural}.
More precisely, the local decrease of the loss is $- \partial_t \mathcal{L}(\theta) = \nabla \ell(f_\theta)^T \cdot K(\theta) \cdot \nabla \ell(f_\theta)$.
As an example, for the quadratic loss, the gradient satisfies $\lVert \nabla \ell(f_\theta) \rVert_2^2 = 4 \ell(f_\theta) = 4 \mathcal{L}(\theta)$, such that a positive definiteness condition $K(\theta) \succeq \mu > 0$ guarantees the Polyak-\Loja{} condition $-\partial_t \mathcal{L}(\theta) \geq 4 \mu \, \mathcal{L}(\theta)$, and thus by integration, convergence to zero with a linear convergence speed.
Several works, starting with \citet[Proposition 2]{jacot2018neural} but also \citet{du2018gradient}, have shown that the smallest eigenvalue of this $K(\theta)$ operator is indeed strictly positive if the network is sufficiently overparameterized ($m \gg n$). Subsequent papers have also anayzed \textit{how overparameterized} the network needs to be for this argument to hold, with interesting asymptotic bounds on the number of parameters required \citep{ji2020polylogarithmic,chen2021how}.

\paragraph{Challenges.}
However, this argument for convergence is bound to fail when there are fewer parameters than datapoints ($m < n$).
In particular, for a fixed number of parameters $m \in \mathbb{N}$, it is impossible to have both $n \to +\infty$ and $\lambda_{\min}(K(\theta)) > 0$,
since $K(\theta) \in \mathbb{R}^{n \times n}$ has rank $m < n$ by definition.
As argued by \citet[Proposition~3]{liu2022loss} for the quadratic loss ($\ell : f \mapsto \lVert f - y \rVert_2^2$, satisfying $\nabla \ell(f_\theta) = 2 (f_\theta - y) \in \mathbb{R}^n$), this implies that for underparameterized systems, the \Loja{} condition cannot be satisfied for all $y$, since $\inf_{u \in \mathbb{R}^n} u^T K(\theta) u / u^T u = \lambda_{\min}(K(\theta)) = 0$.
Nonetheless, if some knowledge $y_i = f^*(x_i)$ for some $f^* \in \mathcal{F}_0$ is available, then it is sufficient to show that $\inf_{u \in \mathcal{Y}_0} u^T K(\theta) u / u^T u > 0$, where $\mathcal{Y}_0 = \{ (f^*(x_i) - f_\theta(x_i))_i, f^* \in \mathcal{F}_0 \} \subseteq \mathbb{R}^n$ is only a subset of the responses $\mathbb{R}^n$ on which the smallest eigenvalue of the NTK might be positive.
Bounding the eigenvalues of the NTK away from zero is sufficient, but not necessary,
and for cases where the smallest eigenvalue is zero, one can bound the Rayleigh quotient of the gradient and enjoy similar guarantees despite the null eigenvalue(s).
Although stated differently in their respective context, previous uses of this restricted eigenvalue argument can be found for instance in 
\citet[Assumption A4: response is NTK-separable]{nitanda2019refined},
or \citet[Section 6, bounded inverse-NTK response]{arora19finegrained}
.
We show how the argument used in these particular cases can be extended to a broader setting, and introduce tools to make calculations easier and obtain such guarantees.

Rayleigh quotient bounds enable convergence guarantees in the underparameterized regime ($m < n$) and in particular, for fixed number of parameters $m$, the guarantees hold even when the number of datapoints grows ($n \to +\infty$) and the domain becomes continuous.
Letting $n \to +\infty$ requires a slightly different formalism than the vectors and matrices used in this introduction, we will therefore use functional spaces in the following, and the usual notations of differential geometry, with parameters in indices for instance.
%
Contrary to results such as \citet{arora19finegrained, du2019gradient}, the formulation using functional spaces, from \citet{jacot2018neural}, extends to the case where datapoints are arbitrarily close and even identical, allowing guarantees on the expected loss with respect to a continuous distribution and not just the empirical loss measured on finitely many well-separated samples. In particular, these conditions need not rely on properties satisfied only with high-probability by random initialization when $m \to +\infty$, they can be proven even for fixed initialization and $m \in \mathbb{N}$.

Lastly, our analysis ties together in a more general framework the convergence arguments formulated in the functional space \citep{du2018gradient,du2019gradient} studying dynamics of the network response, and similar arguments formulated in the parameter space \citep{li2018learning,zou20gradient}, by centering the work on the singular values of the network differential $D F(\theta) \in \R^{n \times m}$ rather than the functional-space $D F(\theta) \cdot D F(\theta)^T \in \R^{n \times n}$ or parameter-space $D F(\theta)^T \cdot D F(\theta) \in \R^{m \times m}$ kernels.


\paragraph{Contributions.}
We provide definitions in Sec.~\ref{sec:setup-and-notations},
then present Kurdyka-\Loja{} inequalities, Rayleigh quotients, and their link in Sec.~\ref{sec:tools-to-obtain-guarantees}.
We show in Sec.~\ref{sec:linear-models} that this recovers previously known linear bounds for the quadratic case.
We illustrate a two-dimensional counterexample to the NTK positive-definiteness in Sec.~\ref{sec:lemniscate}, and how to overcome it with Rayleigh quotients.
In Sec.~\ref{sec:logistic-regression} we prove a new bound on logistic regression obtained by the same technique. In Sec.~\ref{sec:relu} and Sec.~\ref{sec:periodic}, we outline arguments of convergence in more realistic settings and highlight future challenges.

\section{Definitions for gradient flows and neural tangent kernels}\label{sec:setup-and-notations}


Let $\mathcal{X}$ be a set with no particular structure.
We consider the problem of learning a target function $f^* : \mathcal{X} \to \mathbb{R}$, by having access only to samples $(x, f^*(x)) \in \mathcal{X} \times \mathbb{R}$, where $x \sim \mathcal{D}$ are random samples from a probability distribution $\mathcal{D}$ on $\mathcal{X}$.
Let $\mathcal{F} = \mathbb{R}^\mathcal{X}$ be the vector space of functions from $\mathcal{X}$ to $\mathbb{R}$. The setting presented in the introduction corresponds to $\mathcal{X}$ being finite containing the examples $x_i$ so that functions are represented as vectors $f = (f(x_i))_{i \in [n]}$ and $\mathcal{D}$ is the empirical measure on $\mathcal{X}$.

\begin{definition}
A network map is a function $F : \Theta \to \mathcal{F}$, from $\Theta$ a vector space of finite dimension equipped with an inner product $\langle \cdot, \cdot \rangle_\Theta$, to $\mathcal{F}$ equipped with the topology of pointwise convergence.
\end{definition}

To avoid confusions as much as possible, we will reserve lowercase letters $(f,g,h)$ for functions in $\mathcal{F}$, and the uppercase $F$ for network maps.
We will usually put the parameters in index, and inputs between parenthesis, so that for $\theta \in \Theta$, the function $f_\theta : \mathcal{X} \to \mathbb{R}$ sends inputs $x \in \mathcal{X}$ to outputs $f_\theta(x) \in \mathbb{R}$.
Readers familiar with differential geometry will note that the assumption that $\Theta$ is a vector space is a simplification, and could be relaxed for instance to a differentiable manifold. However, we are interested in easily readable results closest to applications, and this assumption will avoid cumbersome discussions on the parameter manifold's tangent space, and keep results readable with only some background in linear algebra. In all the examples, it is sufficient for our needs to set $\Theta = \mathbb{R}^m$ with canonical inner product and $\lVert \cdot \rVert_\Theta = \lVert \cdot \rVert_2$, for some number of parameters $m \in \mathbb{N}$.

\begin{definition}[$\mathcal{D}$-seminorm]\label{def:seminorm}
Any probability distribution $\mathcal{D}$ on $\mathcal{X}$ induces on $\mathcal{F}$ a
bilinear symmetric positive semi-definite form
$\langle \cdot, \cdot \rangle_\mathcal{D} : \mathcal{F} \times \mathcal{F} \to \mathbb{R}$,
defined for $(g,h) \in \mathcal{F} \times \mathcal{F}$ as
\[ \langle g, h \rangle_\mathcal{D} = \mathbb{E}_{x \sim \mathcal{D}} \left[ g(x) h(x) \right] \]
The associated seminorm $\lVert \cdot \rVert_\mathcal{D} : \mathcal{F} \to \mathbb{R}_+$ is defined as $\lVert g \rVert_\mathcal{D}^2 = \langle g, g \rangle_\mathcal{D} = \mathbb{E}_{x \sim \mathcal{D}} \left[ g(x)^2 \right] $.
\end{definition}

This seminorm does not in general separate points, it is therefore not a norm on $\mathcal{F}$. In particular, if $\mathcal{D}$ does not have full support, then there are non-null functions $g \in \mathcal{F}$ with null seminorm $\lVert g \rVert_\mathcal{D} = 0$.

\begin{definition}[Gradient flow]
A gradient flow with respect to the differentiable loss $\mathcal{L} : \Theta \to \mathbb{R}_+$ is an absolutely continuous curve $\theta : \mathbb{R}_+ \to \Theta$ satisfying the differential equation
$ \partial_t \theta = - \nabla \mathcal{L}(\theta) $.
Additionally, we say that a gradient flow is trivial if $\mathcal{L}(\theta_0) = 0$, since it implies that for all $t, \, \theta_t = \theta_0$.
For $\mathcal{U} \subseteq \Theta$, if $\theta : \mathbb{R}_+ \to \Theta$ is a gradient flow such that $\theta(\mathbb{R}_+) \subseteq \mathcal{U}$ then we write just $\theta : \mathbb{R}_+ \to \mathcal{U}$.
\end{definition}
A common choice for regression with target $f^* \in \mathcal{F}$ is the quadratic loss $\mathcal{L} : \theta \mapsto \lVert F(\theta) - f^* \rVert_\mathcal{D}^2$.


If a network map $F : \Theta \to \mathcal{F}$ is differentiable for the pointwise convergence, we will write $\d F_\theta : \Theta \to \mathcal{F}$ for the differential of $F$ at $\theta \in \Theta$, with parameters in index for shortness.
Evaluation at $x \in \mathcal{X}$ and derivation with respect to $\theta \in \Theta$ commute, easing computations (see Appendix~\ref{appendix:eval-deriv-commutation}).
We write the corresponding gradient $\nabla F_\theta : \mathcal{X} \to \Theta$, defined by $\langle \nabla F_\theta(x), \nu \rangle_\Theta = (\d F_\theta \cdot \nu)(x)$ for all $x \in \mathcal{X}$ and $\nu \in \Theta$.

\begin{definition}[Neural Tangent Kernel, NTK form]\label{def:NTK-primal}
A differentiable network map $F : \Theta \to \mathcal{F}$ defines at every point $\theta \in \Theta$ a kernel function $K_\theta : \mathcal{X} \times \mathcal{X} \to \mathbb{R}$ as
\[ K_\theta : (x, x') \mapsto \left\langle \nabla F_\theta(x), \nabla F_\theta(x') \right\rangle_\Theta \]
This function induces a  bilinear symmetric positive semi-definite form 
$\overbar{K_\theta} : \mathcal{F} \times \mathcal{F} \to \mathbb{R}$
as
\[ \overbar{K_\theta}(g,h) = \mathbb{E}_{x \sim \mathcal{D}, x' \sim \mathcal{D}} \left[ g(x) K_\theta(x, x') h(x') \right] \]
\end{definition}

In exponent notation, this bilinear form has signature
$\overbar{K_\theta} : \mathbb{R}^\mathcal{X} \times \mathbb{R}^\mathcal{X} \to \mathbb{R}$,
while the kernel $K_\theta \in \mathbb{R}^{\mathcal{X} \times \mathcal{X}}$ is an $n \times n$ matrix when $\mathcal{X}$ is finite with $n \in \mathbb{N}$ elements.
Importantly, the (primal) kernel $K_\theta$ is independent of the distribution $\mathcal{D}$, while the (dual) kernel form $\overbar{K_\theta}$ changes with $\mathcal{D}$.

\begin{definition}[$\mathcal{D}$-compatibility, functional gradient]\label{def:functional-gradient}
A function $\ell : \mathcal{F} \to \mathbb{R}_+$ is said $\mathcal{D}$-compatible if $\forall (f,g) \in \mathcal{F} \times \mathcal{F}$, it holds that $(f=g)$ $\mathcal{D}$-almost everywhere implies $\ell(f) = \ell(g)$.

Moreover, if $\ell$ is $\mathcal{D}$-compatible and differentiable, we say $\nabla \ell : \mathcal{F} \to \mathcal{F}$ is a gradient of $\ell$ if it satisfies
\[ \forall (f, g) \in \mathcal{F} \times \mathcal{F}, \> \langle \nabla \ell_f, g \rangle_\mathcal{D} = \d \ell_f (g) \]
\end{definition}

This formalizes the idea that the loss depends \textit{only} on the training samples,
and the use of a gradient simplifies the following statements.
When it exists, the functional gradient is usually not unique, for it is defined only $\mathcal{D}$-almost everywhere.
See Appendix~\ref{appendix:loss-gradient} for some examples of conditions under which it is well defined (for instance $\mathcal{D}$ has finite support, or $\ell$ is the expectation of a pointwise loss).

\section{Rayleigh quotients to obtain Kurdyka-\Loja{} inequalities}\label{sec:tools-to-obtain-guarantees}

\subsection{Context: Kurdyka-\Loja{} inequalities for convergence}

All convergence proofs presented in this paper rely on inequalities introduced by \citet{Kurdyka1998} of the form of Proposition \ref{prop:KLoja-direct}.
These are used for instance to prove finite length of trajectories in dynamical systems (see e.g.\ \citet[Corollary~4.1]{bolte07}), and sufficient to prove convergence to a loss of zero even for non-convex losses. We will therefore direct all later efforts to the construction of such inequalities.
This was introduced as an extension to the Polyak-\Loja{} inequalities for linear convergence \RevDiff{\citep[see e.g.][Section 1.3 for examples]{nguyen2017inegalites}}, to more general dynamics, and the proof of the following proposition is a simple application of the chain rule to $\varphi \circ \mathcal{L}$ (see \ref{appendix:KL-direct-proof}).

\begin{proposition}[Convergence by Kurdyka-\Loja{} inequality]\label{prop:KLoja-direct}
Let $\mathcal{U} \subseteq \Theta$.
If $\mathcal{L} : \mathcal{U} \to \mathbb{R}_+$ is such that there exists $\mu \in \mathbb{R}_+^*$ and a strictly increasing differentiable function $\varphi : \mathbb{R}_+^* \to \mathbb{R}$ satisfying
\[ \forall \theta \in \mathcal{U}, \> \mathcal{L}(\theta) \neq 0 \Rightarrow \d \varphi_{\mathcal{L}(\theta)} \, \left\langle \nabla \mathcal{L}(\theta) , \nabla \mathcal{L}(\theta) \right\rangle_\Theta \geq  \mu \]
Then all non-trivial gradient flows $\theta : \mathbb{R}_+ \to \mathcal{U}$ of $\mathcal{L}$ satisfy
$ \forall t \in \mathbb{R}_+, \, \mathcal{L}(\theta_t) \leq \varphi^{-1}\left( \varphi(\mathcal{L}(\theta_0)) - \mu t \right) $
\end{proposition}

Moreover, if such a flow exists, then $\inf_\theta \mathcal{L}(\theta) = 0$ and $\varphi(u) \to - \infty$ if $u \to 0$ (see Appendix~\ref{appendix:KL-details}).

The central idea, similar to the one used in the following sections, is that a
desingularizing function $\varphi : \R_+^* \to \R$ transports the loss evolution
$\mathcal{L}(\theta) : I \to \R_+^*$ in $\operatorname{dom}(\varphi) = \R_+^*$
to the space $\operatorname{Im}(\varphi) = \R$ where the evolution is easy to understand,
since $(\varphi \circ \mathcal{L})(\theta)$ is bounded by an affine function of time.
The desingularizing function provides a way to transfer the understanding of the convergence in
the image of $\varphi$ back to the domain of $\varphi$, where the loss evolution is a little more complicated.
The condition is also sometimes written $\nabla \mathcal{L} \cdot \nabla \mathcal{L} \geq \psi(\mathcal{L})$, where $\psi : \R_+ \to \R_+$ is $(\psi(u))^{-1} = \d \varphi_u$.

For the case of a linear convergence speed guarantee, the Polyak-\Loja{} condition from the introduction (i.e. $- \partial_t \mathcal{L}(\theta) = \lVert \nabla \mathcal{L}(\theta) \rVert_2^2 \geq \mu \, \mathcal{L}(\theta)$) corresponds to the choice $\varphi : u \mapsto \log(u)$.
To accurately describe systems with more intricate dynamics, more complicated choices of $\varphi$ may be necessary, see the case of logistic regression in Sec.~\ref{sec:logistic-regression} for one such example.

\subsection{Contribution: Kurdyka-\Loja{} inequalities by composition}

\begin{definition}[Rayleigh quotients of bilinear maps]\label{def:rayleigh-quotient}
Let $(V, \lVert \cdot \rVert_V)$ and $(W, \lVert \cdot \rVert_W)$ be two vector spaces equipped with seminorms,
and let $A : V \times W \to \mathbb{R}$ be a bilinear map.
Then for $(x,y) \in V \times W$ such that $(\lVert x \rVert_V \in \mathbb{R}_+ \setminus \{0\})$, and $(\lVert y \rVert_W \in \mathbb{R}_+ \setminus \{0\})$,
define the Rayleigh quotient
\[ \mathrm{R}(A; x, y) = \frac{A(x, y)}{\lVert x \rVert_V  \lVert y \rVert_W} \]
\end{definition}

With a symmetric map $A : V \times V \to \mathbb{R}$, the Rayleigh quotient $\mathrm{R}(A; x,x)$ is a convex combination of the eigenvalues of $A$ (which are real-valued), whose weighting depends on $x$.
Moreover, the minimal value is attained when $x$ is an eigenvector corresponding to the minimal eigenvalue,
and $\lambda_{\min}(A) = \inf_{x \in V \setminus \{0\}} R(A; x, x)$.
Lastly, when the map is an inner product, then the Rayleigh quotient $R(\inner{\Theta}; a, b) = \langle a, b \rangle_\Theta / \lVert a \rVert_\Theta \lVert b \rVert_\Theta$ is a form of cosine similarity.
The most common usage is with $x=y$, but the asymmetric definition will be necessary later for the variational bound.

\begin{proposition}[Kurdyka-\Loja{} inequality by composition]\label{prop:KLoja-comp}
Let $F : \Theta \to \mathcal{F}$ be a differentiable network map, and $K_\theta$ the associated neural tangent kernel (by Def~\ref{def:NTK-primal}).
Let $\mathcal{U} \subseteq \Theta$ be a subset of parameters and $\mathcal{F}_\mathcal{U} = F(\mathcal{U}) \subseteq \mathcal{F}$ its image by $F$.
Let $\ell : \mathcal{F}_\mathcal{U} \to \mathbb{R}_+$ be a $\mathcal{D}$-compatible differentiable loss with gradient $\nabla \ell : \mathcal{F}_{\mathcal{U}} \to \mathcal{F}$ whose seminorm is finite $\forall f \in \mathcal{F}_\mathcal{U}, \lVert \nabla \ell_f \rVert_\mathcal{D} < + \infty$.\newline
Assume that there exists a strictly increasing differentiable $\varphi : \mathbb{R}_+^* \to \mathbb{R}$ satisfying
\[ \forall f \in \mathcal{F}_\mathcal{U},\, \ell(f) \neq 0 \Rightarrow \d \varphi_{\ell(f)} \left\langle \nabla \ell_f , \nabla \ell_f \right\rangle_\mathcal{D} \geq 1 \]
If the $\overbar{K_\theta}$-Rayleigh quotient of the gradient of $\ell$ is bounded below, i.e. if there exists $\mu \in \mathbb{R}_+^*$ such that
\[ \forall \theta \in \mathcal{U}, \, \ell(F(\theta)) \neq 0 \Rightarrow
\mathrm{R}\left(\overbar{K_\theta}; \nabla \ell_{F(\theta)}, \nabla \ell_{F(\theta)} \right)
\geq \mu \]
Then, for $\mathcal{L} = (\ell \circ F) : \mathcal{U} \to \mathbb{R}_+$, it holds
\[\forall \theta \in \mathcal{U},\, \mathcal{L}(\theta) \neq 0 \Rightarrow \d \varphi_{\mathcal{L}(\theta)} \left\langle \nabla \mathcal{L}(\theta), \nabla \mathcal{L}(\theta) \right\rangle_\Theta \geq \mu \]
\end{proposition}

The proof of this statement is deferred to Appendix~\ref{proof:composition}, and similar to the usual NTK arguments.
If $\overbar{K_\theta}$ is $\mu$-uniformly conditioned, then in particular $\overbar{K_\theta}(\nabla \ell_f, \nabla \ell_f) \geq \mu \langle \nabla \ell_f, \nabla \ell_f \rangle_\mathcal{D}$, which is exactly the Rayleigh quotient condition.
The main difference is that it is not necessary to require \textit{uniform} conditioning, it is sufficient for this property to hold on any subspace containing the gradient (and in particular the one-dimensional subspace defined by the gradient, i.e. the Rayleigh quotient).

Kurdyka-\Loja{} (K\L{}) inequalities provide a reasonable path to convergence bounds, outside the usual convex framework. However, they can still be very difficult to obtain.
This proposition splits the parametric-space K\L{} inequality into a functional-space K\L{} inequality which is easier to obtain (trivial for quadratic losses, see Sec.~\ref{sec:linear-models}; available for cross-entropy for instance, see Sec.~\ref{sec:logistic-regression}) and a Rayleigh quotient bound, which is the focus of the following propositions.
Similarly, we provide hereafter several variational forms that can help break the Rayleigh quotient bounding problem down into smaller blocks that can be easier to compute independently before reassembling.

\begin{proposition}[Variational bound]\label{prop:variational}
Let $F : \Theta \to \mathcal{F}$ be a differentiable network map,
$K_\theta$ the associated neural tangent kernel (by Def~\ref{def:NTK-primal}), and $\theta \in \Theta$.
If $h \in \mathcal{F}$ satisfies $\lVert h \rVert_\mathcal{D} \neq 0$, then it holds
\[ \mathrm{R}(\overbar{K_\theta}; h, h) = \sup_{\nu \in \Theta \setminus \{0\}} \mathrm{R}(\overbar{\d F_\theta}; \nu, h)^2 \]
Where $\overbar{\d F_\theta}$ is the bilinear form $(\nu, h) \mapsto \langle \d F_\theta \cdot \nu, h \rangle_\mathcal{D}$ associated with the linear operator $\d F_\theta$.
\end{proposition}

This property is particularly useful to avoid dealing with the square of the differential, and instead obtain lower-bounds on the Rayleigh quotient by carefully selecting (suboptimal) inputs $\nu \in \Theta \setminus \{0\}$.

\begin{proposition}[Split cosine - singular value]\label{prop:cosine-singular-split}
Let $F : \Theta \to \mathcal{F}$ be a differentiable network map, $K$ the associated neural tangent kernel, $\theta \in \Theta$, and $h \in \mathcal{F}$ such that $\lVert h \rVert_\mathcal{D} \neq 0$.
If there exists a subspace $\Theta_0 \subseteq \Theta$ and some $\mu \in \mathbb{R}_+^*$ such that
there exists $\nu \in \Theta_0$ satisfying $\mathrm{R}(\inner{\mathcal{D}}; \d F_\theta \cdot \nu, h) \geq \mu$,
then for $\lambda = \inf_{\nu \in \Theta_0} \lVert \mathrm{d} F_\theta \cdot \nu \rVert_\mathcal{D}^2 / \lVert \nu \rVert_\Theta^2 \in \mathbb{R}_+$, it holds $\mathrm{R}(\overbar{K_\theta}; h, h) \geq \mu^2 \, \lambda$.
\end{proposition}

This proposition is a trivial consequence of the following one, but is easier to parse while still making
apparent the distinction between a geometric quantity $\mu$ and the singular value $\lambda$.
See Sec.~\ref{sec:lemniscate} for an example in dimension two, where $\mu$ is defined only by the angle between the gradient and the lemniscate's tangent, independently of the parameterization. Observe on the other hand that as $\lambda$, the speed at which the lemniscate is traveled, changes, so does the gradient flow's convergence speed.

\begin{proposition}\label{prop:shattering}
Let $F : \Theta \to \mathcal{F}$ be a differentiable network map, $\theta \in \Theta$, and $h \in \mathcal{F}$ s.t. $\lVert h \rVert_\mathcal{D} \neq 0$.

Let $k \in \mathbb{N}^*$. Let $(a_i)_{i \in [k]} \in (\Theta \setminus \{0\})^k$ and $(g_i)_{i \in [k]} \in (\mathcal{F} \setminus (\lVert \cdot \rVert_\mathcal{D})^{-1}(0) )^k$. If $h \in \mathrm{Span}(g)$, then
%
\[ \max_{\nu \in \mathrm{Span}(a) \setminus \{0\}} \mathrm{R}(\overbar{\d F_\theta}; \nu, h) \geq
\frac{\lambda_{\min}\left( {\mathrm{R}\left(\inner{\mathcal{D}}; \d F_\theta \cdot a_i, g_j \right)}_{i,j} \right)  \min_{i \in [k]} \lVert\mathrm{d} F_\theta \cdot a_i \rVert_\mathcal{D} / \lVert a_i \rVert_\Theta}{\sqrt{\lambda_{\max} \left( {\mathrm{R}\left(\inner{\Theta}; a_i, a_j\right)}_{i,j}\right) \lambda_{\max}\left( {\mathrm{R}\left(\inner{\mathcal{D}}; g_i, g_j \right)}_{i,j} \right)}} \]
where the smallest singular value of $A \in \mathbb{R}^{k \times k}$ is $\lambda_{\min}(A) = \min_{u \neq 0} u^T A u / u^T u$ (resp. $\max$).
\end{proposition}

If the vectors $(a,g)$ are taken orthogonal and such that $\d F_\theta \cdot a_i = \sigma_i g_i$ for some $\sigma_i \in \mathbb{R}$, then the three matrices are the identity, and only the minimal Rayleigh quotient remains.
If they are chosen only approximately orthogonal, then a corresponding multiplicative penalty is incurred.

The proofs of the preceding three propositions are deferred to Appendix \ref{appendix:variational}, \ref{appendix:split-cosine-singular} and \ref{appendix:approx-svd} respectively.

\section{Case studies}

\subsection{Linear models with quadratic loss, recovering known bounds}\label{sec:linear-models}
As a sanity check and simple first contact with the variational bound, we consider a model linear in its parameters, with quadratic loss, and recover the (known optimal) linear convergence rate.
This proposition is the continuous time form of \citet[Theorem~1]{karimi2016linear}.

\begin{proposition}[Convergence of quadratic-loss linear models]
Let $\mathcal{X} = \Theta= \mathbb{R}^d$, and $F : \Theta \to \mathcal{F},$ be the linear network map $F : \theta \mapsto f_\theta$
defined by $f_\theta(x) = \langle x, \theta \rangle$.
Let $f^* : \mathcal{X} \to \mathbb{R}$ be a linear function.
Let $\mathcal{L} : \Theta \to \mathbb{R}_+$ be the quadratic loss $\mathcal{L} : \theta \mapsto \lVert F(\theta) - f^* \rVert_\mathcal{D}^2$ where
$\mathcal{D}$ a distribution over $\mathcal{X}$ such that $\mathcal{L}$ is well-defined and finite.

If $\theta : \mathbb{R}_+ \to \Theta$ is a gradient flow of $\mathcal{L}$, then
for all $t\in \mathbb{R}_+$, it holds $\mathcal{L}(\theta_t) \leq \mathcal{L}(\theta_0) \, e^{ - 4 \, \lambda_{\min}^+(A) \, t}$,
where $A = \mathbb{E}_{x \sim \mathcal{D}}\left[ x x^T \right] \in \mathbb{R}^{d \times d}$ is the (uncentered) covariance matrix of the samples, and $\lambda_{\min}^+(A)$ its smallest non-null eigenvalue.
Moreover, there exists $\mathcal{D}$ such that this bound is an equality.
\end{proposition}

The idea is to apply Proposition~\ref{prop:KLoja-comp}. The functional Kurdyka-\Loja{} inequality is immediate, and we bound the Rayleigh quotient with Proposition~\ref{prop:cosine-singular-split} applied to the subspace $\Theta_0 = \mathrm{Ker}(A)^\bot$.

\begin{proof}
Let $\ell : \mathcal{F} \to \mathbb{R}_+, f \mapsto \lVert f - f^* \rVert_\mathcal{D}^2$ be the functional-space quadratic loss, whose gradient $\nabla \ell_f = 2 (f - f^*)$ satisfies the Polyak-\Loja{} inequality $\lVert \nabla \ell_f \rVert_\mathcal{D}^2 \geq 4 \, \ell(f)$.
Hence, let us show $\mathcal{L}(\theta) \neq 0 \Rightarrow \mathrm{R}(\overbar{K_\theta}; \nabla \ell_{F(\theta)}, \nabla \ell_{F(\theta)}) \geq \lambda_{\min}^+(A)$,
which is sufficient by applying Proposition~\ref{prop:KLoja-comp}.

Let $\theta^* \in \Theta$ be any parameter such that $f^* = f_{\theta^*}$, where existence is guaranteed by linearity of $f^*$.
Observe that the loss can be written $\mathcal{L}(\theta) = (\theta - \theta^*)^T A (\theta - \theta^*)$.
Let $\theta \in \Theta$ such that $\mathcal{L}(\theta) \neq 0$. In particular, $\theta - \theta^* \notin \mathrm{Ker}(A)$. Then, let $\Theta_0 = \mathrm{Ker}(A)^\bot$.
On one hand, it follows that
\[ \sup_{\nu \in \Theta_0 \setminus \{0\}} \frac{ \langle \d F_\theta \cdot \nu, F(\theta) - F(\theta^*) \rangle_\mathcal{D}^2}{\lVert \d F_\theta \cdot \nu \rVert_2^2 \, \lVert F(\theta) - F(\theta^*) \rVert_\mathcal{D}^2 }
= \frac{ (u^T A (\theta - \theta^*))^2 }{(u^T A u) ((\theta - \theta^*)^T A (\theta - \theta^*))}
= 1 \]
with the maximum attained for $u \in \operatorname{Ker}(A)^\bot \setminus \{0\}$ the orthogonal projection of $(\theta - \theta^*)$ to $\operatorname{Ker}(A)^\bot$, satisfying $A (\theta - \theta^*) = A u$ and $\langle \theta - \theta^*, u \rangle = 0$, thus $u^T A u = u^T A (\theta - \theta^*) = (\theta - \theta^*)^T A (\theta - \theta^*)$.

Then by definition
$ \inf_{\nu \in \Theta_0 \setminus \{0\}} \lVert \d F(\theta) \cdot \nu \rVert_\mathcal{D}^2 /\lVert \nu \rVert_2^2
= \inf_{\nu \in \Theta_0 \setminus \{0\}} (\nu^T A \nu)/ (\nu ^T \nu) = \lambda_{\min}^+(A) $.
Conclude by Proposition~\ref{prop:cosine-singular-split}, with $\mu=1$ and $\lambda=\lambda_{\min}^+(A)$.
Equality is recovered for $A = I_d$.
\end{proof}

This is to be contrasted with a direct proof of the Kurdyka-\Loja{} inequality, i.e. showing that
\[ \frac{\lVert \nabla \mathcal{L}(\theta) \rVert_2^2}{\mathcal{L}(\theta)} = 4\, \frac{(\theta - \theta^*)^T A^2 (\theta - \theta^*)}{(\theta - \theta^*)^T A (\theta - \theta^*)} \geq 4 \lambda_{\min}^+(A) \]

Although the proof seems a bit convoluted, the interesting part here is that the original bound can be split into two (hopefully simpler) subproblems, while still allowing the use of knowledge on $(f_\theta - f^*)$, leveraged here by the assumption $(\theta - \theta^*) \in \operatorname{Ker}(A)^\bot$.
Note that knowledge of a property such as $(\theta - \theta^*) \in \Theta_0 \subseteq \mathbb{R}^d$ for any subspace $\Theta_0$ could have been used to eliminate any eigenvalues of $A$ on $\Theta_0^\bot$, including strictly positive eigenvalues, there is nothing specific to $\operatorname{Ker}(A)^\bot$ other than the existence of the prior knowledge $(\theta - \theta^*) \notin \operatorname{Ker}(A)$ granted by $\mathcal{L}(\theta) \neq 0$.

\subsection{Lemniscate-constrained optimization, singular values}\label{sec:lemniscate}

\begin{figure}[H]
\centering
\begin{subfigure}{.5\textwidth}
\centering
\includegraphics[width=.8\textwidth]{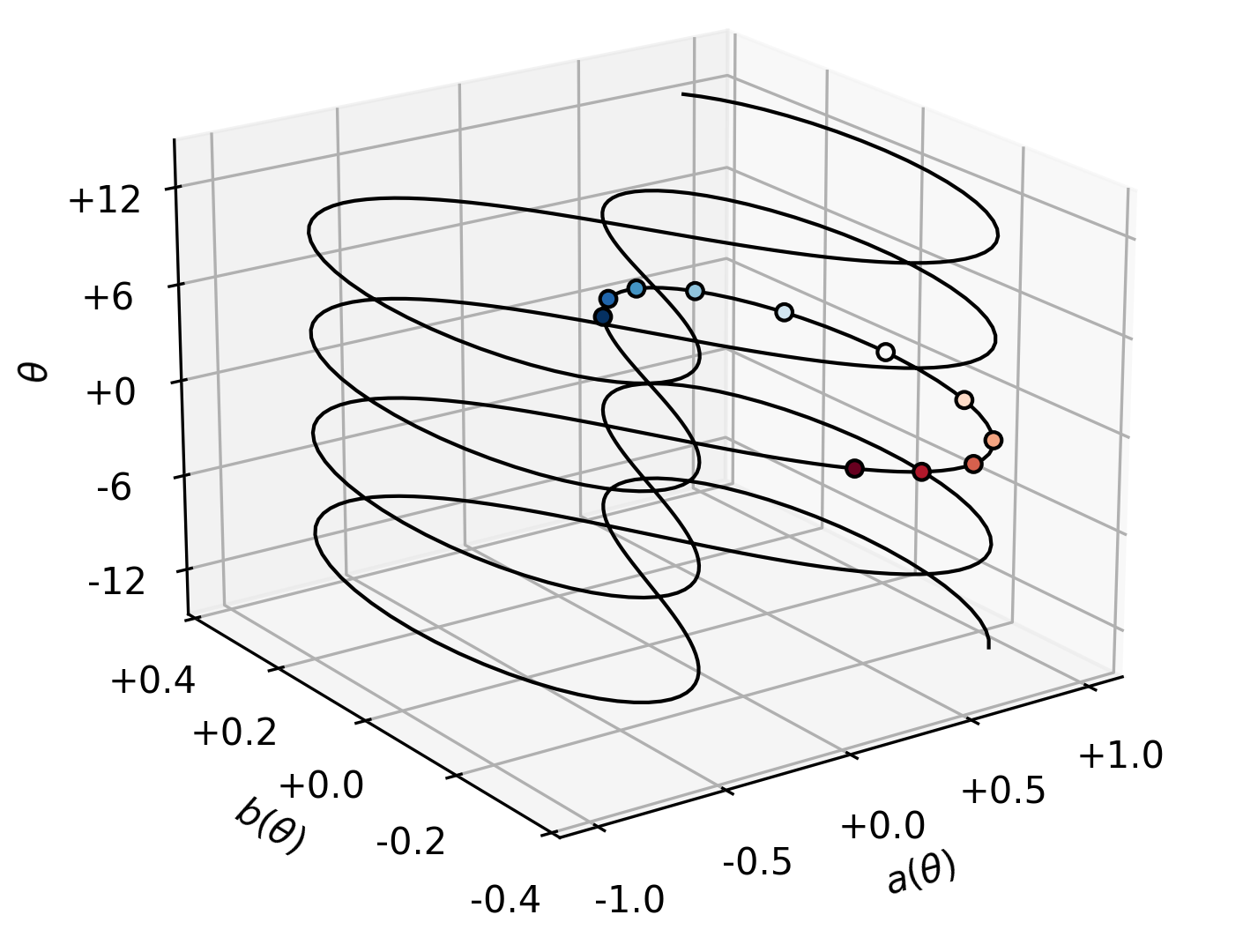}
\subcaption{Periodic lemniscate ($F_S$: sphere to lemniscate)}%
\label{fig:3d-lemniscate_S}
\end{subfigure}%
\begin{subfigure}{.5\textwidth}
\centering
\includegraphics[width=.8\textwidth]{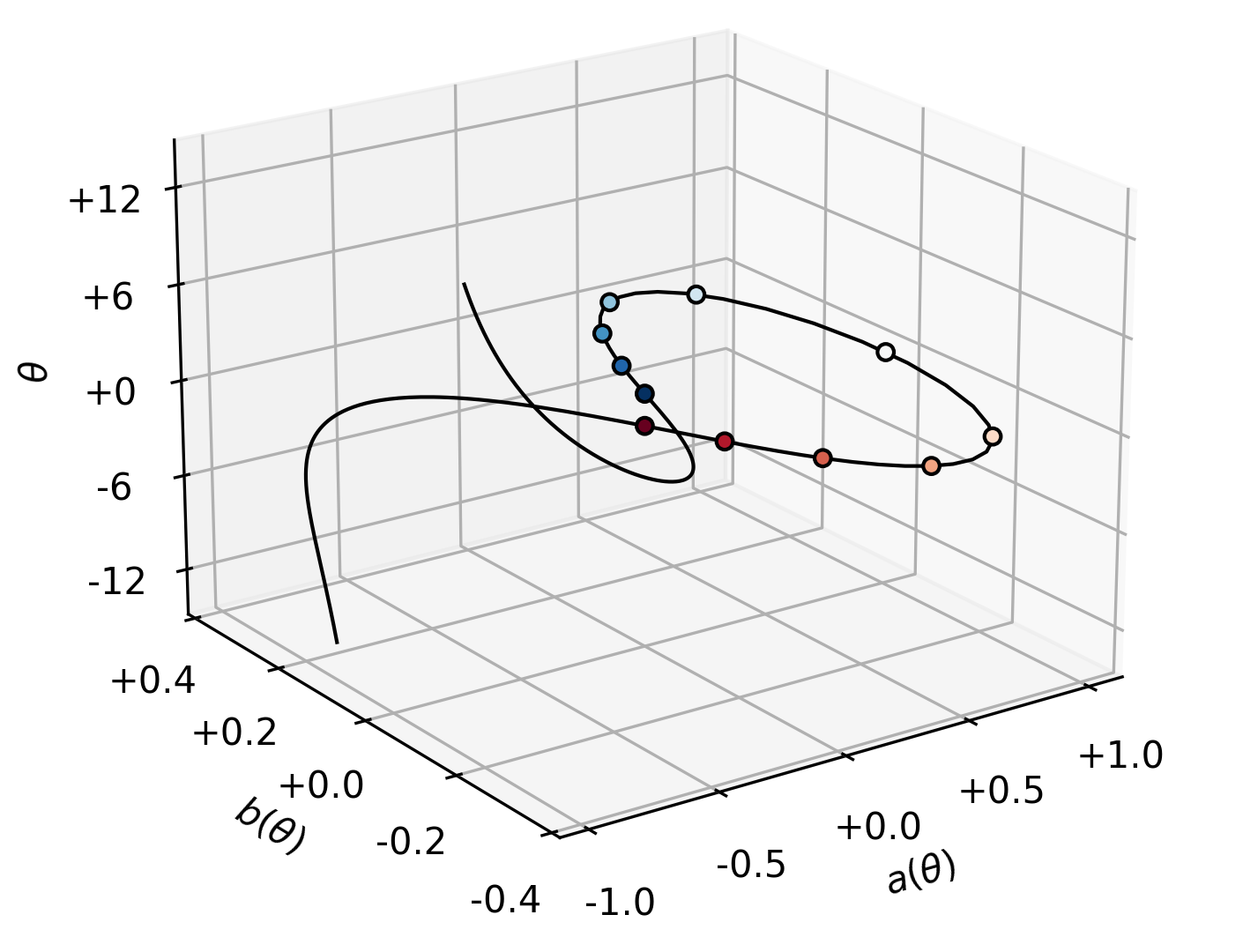}
\subcaption{Punctured lemniscate ($F_L$: line to lemniscate)}%
\label{fig:3d-lemniscate_L}
\end{subfigure}
\caption{Graph of the two parameterizations presented (with 11 dots regularly spaced on [-1,+1])}%
\label{fig:lemniscate-graph}
\end{figure}

\begin{figure}[H]
\centering
\begin{subfigure}{.5\textwidth}
\centering
\includegraphics[width=\textwidth]{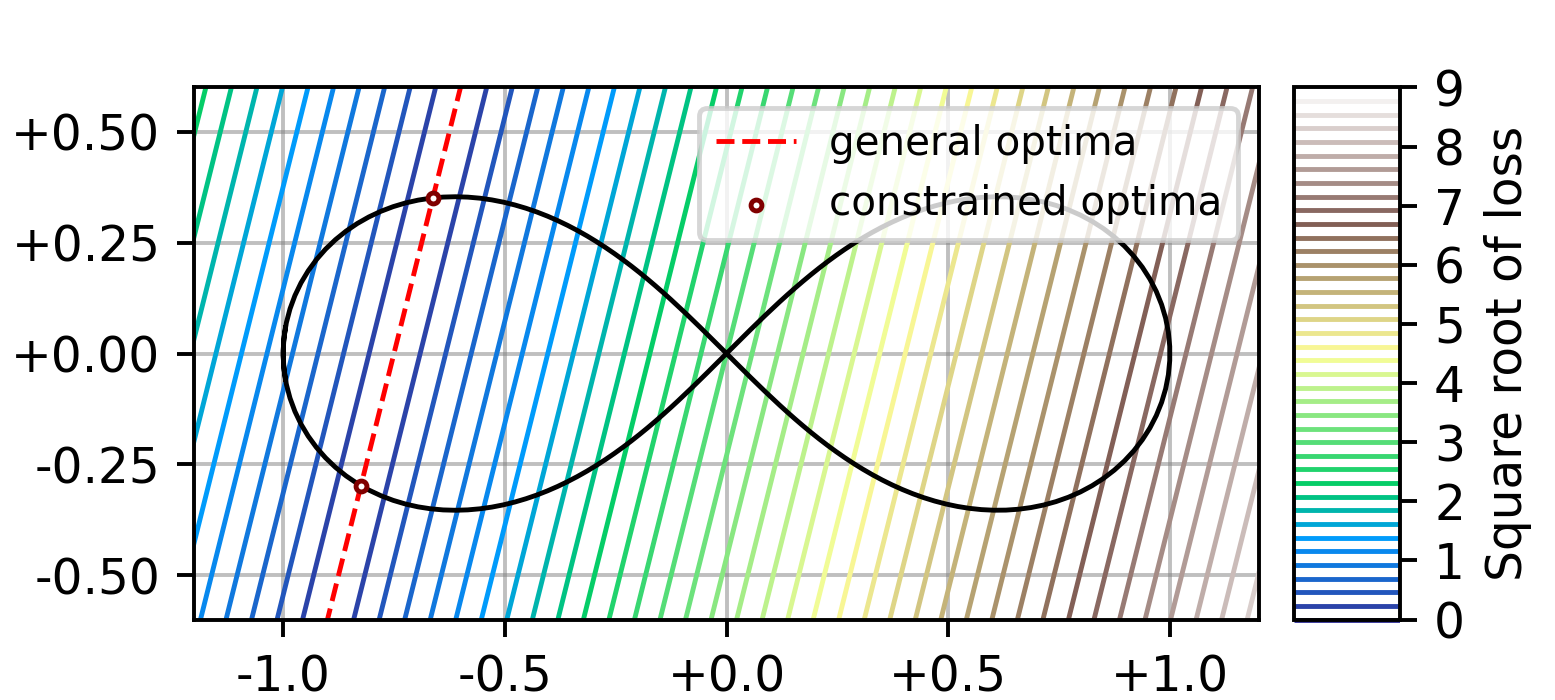}
\caption{Bernoulli's lemniscate $\mathcal{F}_0$ and level sets of $ \ell$}
\label{fig:2d-lemniscate}
\end{subfigure}%
\begin{subfigure}{.5\textwidth}
\centering
\includegraphics[width=6cm]{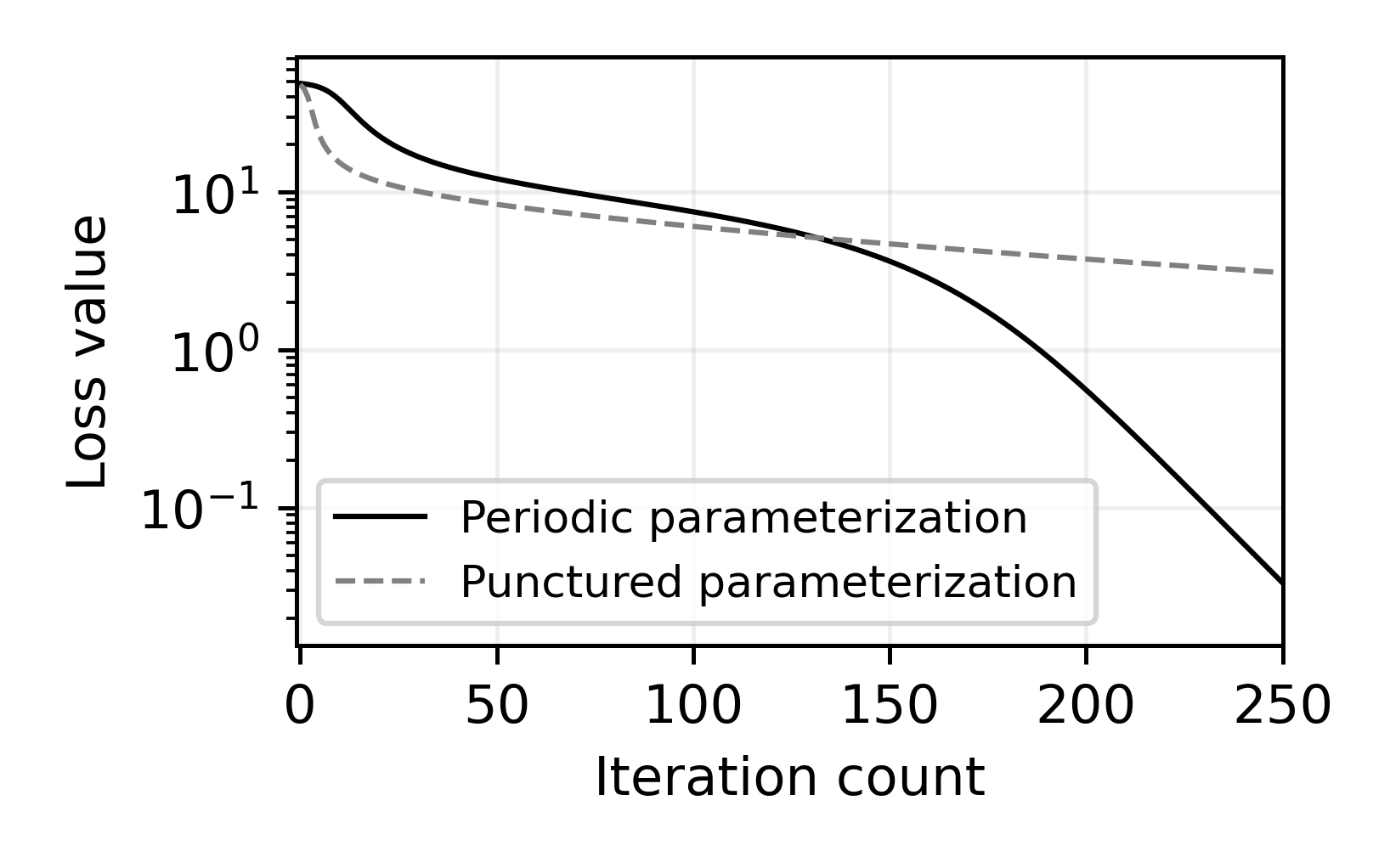}
\subcaption{Observed convergence speeds}%
\label{fig:gd-speed-short}
\end{subfigure}
\caption{Loss level sets with parameters $t=(4,-1)$ and $f^*(t) = -3$, corresponding to quadratic loss $\ell : (a,b) \mapsto (4a - b + 3)^2$ and convergence speed with step size ${10}^{-3}$ and initial estimate $\theta(0)=0$.
Both flows converge to the same functional minimum $(F_S(\theta^*_S) = F_L(\theta^*_L))$, the one depicted on the bottom in (a). Initializing at a different point could have led to a convergence to the other minimum. Proposition \ref{prop:lemniscate-convergence} only shows that the loss converges to zero, leaving unaddressed the question of \textit{which} minimum is reached.}
\end{figure}

We now present a toy example simple enough to allow for explicit computations and constructed to illustrate the importance of parametrization. We consider linear functions in two dimensions where the function $f_{(a,b)} : \mathbb{R}^2 \to \mathbb{R}, \, f_{(a,b)} : (x,y) \mapsto ax + by$ is simply identified with $(a,b) \in \mathbb{R}^2$. We will still consider a quadratic loss but we now assume that the target function $f^*=f_{(a^*,b^*)}$ is linear and with $(a^*,b^*) \in  \mathcal{F}_0 = \{ (a,b) \in \mathbb{R}^2 \,|\, (a^2 + b^2)^2 = a^2 - b^2 \}$. Although we are looking for a two dimensional linear functions $f^*$, knowing that $f^*\in \mathcal{F}_0$ reduces the "degrees of freedom". 
In such a scenario in machine learning, we typically incorporate this information in the parametrization. As a result, we now have only one parameter to estimate, i.e. $\Theta=\mathbb{R}$ and our network maps $F : \mathbb{R} \to \mathbb{R}^2$ will satisfy $\overline{\mathrm{Im}(F)} = \mathcal{F}_0$. 
Note that Bernoulli's lemniscate $\mathcal{F}_0$ (pictured in Fig~.\ref{fig:2d-lemniscate}) is neither a convex set, nor a manifold (due to the crossing at zero). There is no "natural" parametrization of $\mathcal{F}_0$ and as shown below, the chosen parametrization will matter.
For more clarity on the consequences of this parameterization, we use two parameterizations of the lemniscate $\mathcal{F}_0$: 
\[ F_S : \theta \mapsto \left( \frac{\cos(\theta)}{1 + \sin(\theta)^2}, \frac{\sin(\theta) \cos(\theta)}{1 + \sin(\theta)^2} \right)
\quad\text{ and, }\quad F_L : \theta \mapsto \left( \frac{1 - \theta^4}{1 + 6 \theta^2 + \theta^4}, \frac{2 \theta (1 - \theta^2)}{1 + 6 \theta^2 + \theta^4} \right). \]
The graph of these parameterizations $\{ (\theta, F(\theta)) \,\mid\, \theta \in \mathbb{R} \} \subseteq \mathbb{R}^3$ is depicted in Fig.~\ref{fig:lemniscate-graph}.
The first, $F_S$ is differentiable $2\pi$-periodic and surjective, satisfying $F_S([0,2\pi]) = \mathcal{F}_0$.
The second, $F_L$ is differentiable, but it is neither injective (since $F_L(-1) = (0,0) = F_L(+1)$) nor surjective. It is a punctured lemniscate $\mathrm{Im}(F_L) = \mathcal{F}_0 \setminus \{ (-1,0) \}$, it is only dense in the lemniscate $\overline{\mathrm{Im}(F_L)} = \mathcal{F}_0$.

Note that in both cases, the neural tangent kernel $K_\theta$ has rank one (because there is only one parameter), thus $\lambda_{\min}(\overbar{K_\theta}) = 0$ by rank deficiency but we can still prove convergence to zero loss.


To make things even more clear, we assume that all samples are lying on a line: $\mathcal{D}$ is a distribution supported on the one-dimensional subspace $\mathbb{R} \, t$ with $t =(u,v) \in \mathbb{R}^2 \setminus \{0\}$. In words, all the labeled samples are of the form $z(t, a^* u + b^* v) \in \mathbb{R}^2 \times \mathbb{R}$ for some $z\in \mathbb{R}$ and any function $f_{(a,b)}$ with $(a-a^*)u+(b-b^*)v=0$ will achieve a loss of zero. Indeed as shown in previous section, a standard linear regression in this case converges to a loss of zero but the parameters inferred will not be on the lemniscate $\mathcal{F}_0$. With the parametrization $F_S$  or $F_L$, we will find a solution living on $\mathcal{F}_0$, namely one of the two points in $\ell^{-1}(0) \cap \mathcal{F}_0$, as seen in Figure \ref{fig:2d-lemniscate}.

\begin{proposition}[Lemniscate convergence with varying speed]\label{prop:lemniscate-convergence}
Let $(u,v) \in \mathbb{R}^2$ such that $u > 0$ and $|v| < |u|$. Let $y \in \mathbb{R}_-$ such that the equation $(a u + b v = y)$ has exactly two solutions $(a,b) \in \mathcal{F}_0$.

Let $\ell : \mathbb{R}^2 \to \mathbb{R}_+$ be the quadratic loss $\ell (a,b) \mapsto (a u + bv - y)^2$.
Let $\theta_S : \mathbb{R}_+ \to \mathbb{R}$ (resp. $\theta_L : \mathbb{R}_+ \to \mathbb{R}$) be a gradient flow with respect to the loss $\ell \circ F_S$ (resp. $\ell \circ F_L$) such that $\theta(0) = 0$.
Then there exists a constant $\mu_0 \in \mathbb{R}_+^*$ such that it holds
$ \ell(F_S(\theta_S(t))) \leq \ell(0) \, \exp( - 4 \mu_0^2 \lambda_S^* t)$
and
$ \ell(F_L(\theta_L(t))) \leq \ell(0) \, \exp(- 4 \mu_0^2 \lambda_L^* t) $,
where $\lambda_S^* = \frac{1}{2}$ and $\lambda_L^* = \lVert \nabla F_L(\theta_L^*) \rVert_2^2$, for $\theta_L^* = \lim_t \theta_L(t)$.
\end{proposition}

The sketch of this proof is given in Appendix~\ref{appendix:lemniscate-proof}.
For the numerical values taken in Fig.~\ref{fig:2d-lemniscate}, we have $\lambda_L^* \approx 4.05 \times 10^{-3}$ showing that our bounds capture the speed of convergence.
The idea is as previously, to use the quadratic loss Polyak-\Loja{} property $(\lVert \nabla \ell \rVert_2^2 \geq 4 \ell)$ that will grant linear convergence provided we can show $\mathrm{R}(\overbar{K_\theta}; \nabla \ell_{F(\theta)}, \nabla \ell_{F(\theta)}) \geq \mu_0^2 \lambda_S^*$ for all $\theta \in \theta_S(\mathbb{R}_+)$ (resp. $\lambda_L^*$ for $\theta \in \theta_L(\mathbb{R}_+)$), achieved by a variational bound (Proposition~\ref{prop:variational}) split according to Proposition~\ref{prop:cosine-singular-split}.

\subsection{Cross-entropy minimization with linear models}\label{sec:logistic-regression}

We now consider a classification task with $c\geq 1$ classes. Let  $\Delta_c = \{ u \in (\mathbb{R}_+)^c \,\mid\, \sum_{i \in [c]} u_i = 1 \}$ be the set of distributions over those classes. The samples $x$ live in $\mathcal{X}= \mathbb{R}^d$ and the target function is $f^*:\mathcal{X}\to \Delta_c$.
Let $E : \mathbb{R}^c \to \Delta_c, u \mapsto ( \exp(u_i) / \sum_j \exp(u_j) )_i$ be the softargmax map.
Let $\Theta = \mathbb{R}^{c \times d}$ be the parameter space, and $X : \Theta \to (\mathcal{X} \to \mathbb{R}^c)$ be the operator mapping parameters to linear functions, such that $X(\theta) : x \mapsto \theta \cdot x$.
We use the parameterization $F : \theta \mapsto E(X(\theta))$ where $E$ is applied pointwise.
For any fixed sample $x \in \mathcal{X}$, 
we define the loss for this sample as $H_x : \Delta_c \to \mathbb{R}_+, p \mapsto - \sum_{i \in [c]} f^*_i(x) \log(p_i)$.
The complete loss used to train this model is then the logistic regression $\mathcal{L} : \theta \mapsto \mathbb{E}_{x \sim \mathcal{D}} \left[ H_x(F(\theta)(x)) \right]$, for which we give a new convergence bound.


\vspace{-.15in}
\[ \begin{tikzcd}
\Theta \arrow{r}{X} &(\mathcal{X} \to \mathbb{R}^{c}) \arrow{r}{E} &(\mathcal{X} \to \Delta_c) \arrow{r}{H} &(\mathcal{X} \to \mathbb{R}_+) \arrow{r}{\mathbb{E}_{x \sim \mathcal{D}}} & \mathbb{R}_+
\end{tikzcd} \]


\begin{definition}[Isolation]
A real-valued random variable $Y \in L^1$ is $\kappa$-isolated if $\mathbb{P}(Y \geq \mathbb{E}[Y]) \geq \kappa$.
\end{definition}
All $L^1$ variables are $\kappa$-isolated for some $\kappa > 0$, but we will need a notion of uniform isolation.
A random variable $Y$ with finite support, i.e. $\mathbb{P}(Y = y_i) = p_i$ for some $y \in \mathbb{R}^n$ and $p \in \Delta_n$ is $(\min_{i \in [n]} p_i)$-isolated, regardless of the values $y$.
This bounds the isolation of the maximal value in a sense.
Moreover, if $\psi : \mathbb{R} \to \mathbb{R}$ is increasing and $Y$ is $\kappa$-isolated, then it holds $\mathbb{E}[\psi(Y)] \geq \kappa \psi(\mathbb{E}[Y])$.
\RevDiff{We use $\kappa = 1/n$ in our experiments (see \ref{appendix:logistic-experiments}), where $n \in \mathbb{N}^*$ is the number of training points}.

\begin{definition}[Multi-class separating rays]
We say that a parameter $\zeta \in \mathbb{R}^{c \times d}$ is an $\varepsilon$-separating ray for the distribution $\mathcal{D}$ if it holds for $\mathcal{D}$-almost all $x \in \mathcal{X}$ that
\[ \exists i \in [c], \forall j \in [c] \setminus \{i \}, \> \langle \zeta_i, x \rangle_{\mathbb{R}^d} \geq \langle \zeta_j, x \rangle_{\mathbb{R}^d} + \varepsilon \, \lVert \zeta \rVert_2 \]
where $\zeta_i \in \mathbb{R}^d$ is the $i$-th row of $\zeta$,
i.e.\ if $(\zeta \cdot x) \in \mathbb{R}^c$ has a unique maximum (with a fixed margin).
\end{definition}

This property is invariant by rescaling of $\zeta$ and generalizes the notion of "separation margin" usual in two-class logistic regression.
If $\zeta$ is $\varepsilon$-separating for some $\varepsilon > 0$, then for $\mathcal{D}$-almost all inputs $x$,
the softargmax classifier $f = F(\zeta) : X \to \Delta_c$ induces a unique label $i \in [c]$ as $i = \argmax_j f(x)_j$.

\begin{proposition}[Convergence speed of logistic regression]\label{prop:logistic-convergence}
Let $\mathcal{D}$ be a distribution such that the point-loss random variable $\mathcal{L}_x = H_x(f(x))$, where $x \sim \mathcal{D}$, is $\kappa$-isolated for all $f \in F(\Theta)$.

Let $\mathcal{L} : \theta \in \Theta \mapsto \mathbb{E}_{x \sim \mathcal{D}} \left[ H_x(F(\theta)(x)) \right] \in \mathbb{R}_+$ be the multi-class cross-entropy loss.
If there exists an $\varepsilon$-separating ray $\zeta$ such that $\inf_{\lambda \in \mathbb{R}} \mathcal{L}(\lambda \zeta) = 0$, then for all non-trivial gradient flows $\theta : \mathbb{R}_+ \to \Theta$,
\[ \mathcal{L}(\theta_t) \leq \log\left(1 + \frac{1}{W_0 \left(\exp(\varepsilon^2 \kappa^2 t \RevDiff{\,-\, C}) \right)} \right)  \]
where $W_0 : \mathbb{R}_+ \to \mathbb{R}_+$ is the Lambert function,
 and $C = \log(e^{\mathcal{L}(\theta_0)} - 1) - (e^{\mathcal{L}(\theta_0)} - 1)^{-1} \in \mathbb{R}$.
\end{proposition}

The Lambert function $W_0$ is defined by $W_0(x) e^{W_0(x)} = x$, see \citet{corless96lambert}.
The proof is deferred to Appendix~\ref{appendix:logistic-regression}.
The idea is to prove a functional Kurdyka-\Loja{} inequality by leveraging the isolation property,
then bound the Rayleigh quotient by leveraging the separation and $\inf \mathcal{L} = 0$ hypotheses to obtain a parametric Kurdyka-\Loja{} inequality by Proposition~\ref{prop:KLoja-comp}.

\RevDiff{
Being a convex problem, the classical argument of \citet{boyd04convex} gives a bound $\mathcal{L}(\theta_t) \leq C_0 / t$ as long as there is a finite optimum $\theta^* \in \Theta$. This bound becomes vacuous ($C_0 \to +\infty$) in this setting with dirac labels, common in machine learning, because the infimum is located ``at infinity''. This assumption has been previously lifted (under separability in \citet{soudry18,nacson19convergence} , without separability in \citet{ji18implicit}) to recover the $\mathcal{O}(1/t)$ asymptotic behavior, but without explicit bounds for finite times.

This result is consistent (see Appendix \ref{appendix:logistic-bound-asymptote}) with the asymptotic $\mathcal{O}(1/t)$ bounds from \citet[Theorem 5]{soudry18} with similar hypotheses, this proposition only makes quantitative the non-asymptotic behavior of this system, and the characteristic quantities driving the convergence speed.
To do so, the separation assumption had to be made quantitative, hence the use of $\varepsilon$-separating rays for a fixed positive $\varepsilon$, where previous work used only non-quantified data separation (i.e.\ $\exists \varepsilon$, $\exists \zeta$ s.t. $\zeta$ is an $\varepsilon$-separating ray for the data), see Appendix \ref{appendix:logistic-assumption-discussion} for more details.
Similarly to the previous section, and contrary to the parameter-direction convergence theorems \citet[Theorem 5]{soudry18}, \citet[Theorem 3]{nacson19convergence}, and \citet[Theorem 1.1]{ji18implicit}, this proposition does not, on its own, yield any insights on implicit bias (which infimum is reached) towards max-margin rays, additional arguments are required for this purpose.
The focus here is on the precise quantification of convergence speed under separability assumptions, with continuous time.
}

\subsection{Overparameterized two-layer networks with quadratic loss}\label{sec:relu}

Let $\mathcal{X} = \mathbb{R}^d$, and $\sigma : \mathbb{R} \to \mathbb{R}$ be a non-polynomial Lipschitz map.
For $m \in \mathbb{N} \setminus \{0\}$ a number of neurons.
Let $\Theta^{(m)} = \mathbb{R}^{m \times d} \times \mathbb{R}^{m}$ be a parameter set
and $F^{(m)} : \Theta^{(m)} \to \mathcal{F}$ be the associated network map $F^{(m)}(w, a) : x \mapsto \sum_{i \in [m]} a_i \, \sigma(w_i \cdot x)$, i.e.\ a two-layer network\footnote{The bias term usually present in linear layers is omitted to lighten notations, without loss of generality since an additional dimension with non-null constant coordinate can be added to the input domain to compensate for it.} with non-linearity $\sigma$.

Let $\RevDiff{\mathcal{K}} \subseteq \mathcal{X}$ be compact, and $\mathcal{D}$ a distribution supported on $\RevDiff{\mathcal{K}}$.
Let $f^* \in \mathcal{F}$ be a continuous function.
Over $\Theta^{(m)} = \mathbb{R}^{m \times d} \times \mathbb{R}^m$, let $\mathcal{I}_m$ be the (usual in practice) iid normal rescaled initialization with density
$p(w,a) = \prod_{i \in [m], j \in [d]} \mathcal{N}(w_{i,j}; 0, 1) \prod_{k \in [m]} \mathcal{N}\left(a_k; 0, 1 / \sqrt{m}\right)$.
\RevDiff{We write $(x)_+ = \max(0,x)$}

\begin{proposition}\label{prop:relu-loja}
\RevDiff{
Let $\varepsilon \in \mathbb{R}_+^*$, and $\delta \in ]0,1[$.
There exists $c \in \mathbb{R}_+^*$ such that, for all radii $R \in \mathbb{R}_+^*$, there exists a neuron count $m \in \mathbb{N}$
such that with probability $(1-\delta)$ over initializations $\theta_0 \sim \mathcal{I}_m$, the quadratic
loss $\mathcal{L} : \theta \in \Theta^{(m)} \mapsto \lVert F^{(m)}(\theta) - f^* \rVert_\mathcal{D}^2 $ satisfies the inequality
\[ \forall \theta \in \mathcal{B}(\theta_0, R), \quad \lVert \nabla \mathcal{L}(\theta) \rVert_\Theta^2 \geq \frac{\FinDiff{1}}{(\lVert \theta - \theta_0 \rVert_2 + c)^2} (\mathcal{L}(\theta) - \varepsilon)_+^2 \]
Therefore, for any desired precision $\varepsilon_0 \in \mathbb{R}_+^*$, there exists $(m, \kappa) \in \mathbb{N}^* \times \mathbb{R}_+^*$ such that with probability at least $(1 - \delta)$ over initialization $\theta_0 \sim \mathcal{I}_m$, a gradient flow
$\theta : \mathbb{R}_+ \to \Theta$ of $\mathcal{L}$ with $\theta(0) = \theta_0$ satisfies
$\forall t \in \mathbb{R}_+, \> \mathcal{L}(\theta_t) \leq \varepsilon_0 + 1 / \sqrt[\FinDiff{3}]{\mathcal{L}(\theta_0)^{-\FinDiff{3}} + \kappa \, t}$.
}
\end{proposition}

Proof in Appendix~\ref{appendix:relu-proof}.
The idea for the proof is to use universal approximation property on compacts
\citep{cybenko1989approximation,barron93universal,leshno1993multilayer},
to get $\lVert F(\theta) + \d F_\theta \cdot \nu - f^* \rVert_\mathcal{D}^2 \leq \varepsilon$ for some $\nu \in \Theta$, then derive a Kurdyka-\Loja{} inequality from that with a variation of Proposition~\ref{prop:cosine-singular-split}.
\RevDiff{Knowledge of a Kurdyka-\Loja{} inequality in a ball around initialization alone is not sufficient to show loss convergence to arbitrary precision in general, but the \FinDiff{separable} form of this inequality makes it possible, \FinDiff{following \citet[Proposition~4.6]{scaman22a}}.
This proposition shows convergence outside the vastly overparameterized regime ($m$ is finite even with infinite data), but still relies heavily on a (very) large number of neurons.
In the next section, we give a partial convergence argument using similar techniques in a much more constrained regime.
}

\subsection{Periodic signal recovery}\label{sec:periodic}

Let $\mathcal{X} = \mathbb{R}$. Among functions $\mathcal{F} = (\mathbb{R} \to \mathbb{R})$, we are interested in continuous periodic antisymmetric functions, which we parameterize with
$\Theta = \mathbb{R}^m \times \mathbb{R}^m$, as $F : \Theta \to \mathcal{F}$, defined for $(a, \omega) \in \Theta$ as $F(a, \omega) : x \mapsto \sum_{i \in [m]} a_i \sin(\omega_i x)$,
and $K_{(a, \omega)}$ the associated NTK at the point $(a, \omega) \in \Theta$.


The central property of this application, separating it from the most common machine learning applications, is the inability to obtain good samples. Let $R \in \mathbb{R}_+^*$ be a finite window size, and define the training data distribution $\mathcal{D} = \mathcal{U}(-R, +R)$, the uniform distribution on the interval $[-R, +R]$.
Let $\mathcal{F}_0 \subseteq \mathcal{F}$ be the set of continuous periodic antisymmetric functions with period less than $R$.
Crucially, we are interested not just in learning the function on the interval, akin to just data retrieval, but rather in learning the function in $(\mathbb{R} \to \mathbb{R})$ as a whole.
This problem is well defined, i.e. if $f^* \in \mathcal{F}_0$, then $\argmin_{g \in \mathcal{F}_0} \lVert g - f^* \rVert_\mathcal{D}^2 = \{ f^* \}$.
The periodicity assumptions makes the data \textit{sufficient} to recover the target function among the hypotheses, however neither the assumption that the training and testing data distributions are identical, nor the assumption that the model has more parameters than there are data points are satisfied. There is infinite data, but there is bias in the sampling.

We will rely on two properties of frequency parameters to show bounds.
First, we say that $\omega \in \mathbb{R}^m$ is $\delta$-separated if $\inf_{i \neq j} |\omega_i - \omega_j| \geq \delta$ and $\inf_i | \omega_i | \geq \delta$.
Then, we say that the pair $(\omega, \omega^*) \in \mathbb{R}^m \times \mathbb{R}^m$ is $\varepsilon$-paired if $\sup_{i \in [m]} | \omega_i - \omega_i^* | \leq \varepsilon$.
Moreover, let $x_0 \in \mathbb{R}_+$ be the first zero of $\sinc''$. ($x_0 \approx 2.0815$).

\begin{proposition}[Polyak-\Loja{} region]\label{prop:sine-paired}
Let $(\eta, \mu) \in \mathbb{R}_+^* \times \mathbb{R}_+^*$ such that $\eta \leq x_0$ and $\eta < \frac{1}{2} \mu$.

Let $f^* \in \mathcal{F}$ be a target, and $\ell : f \in \mathcal{F} \mapsto \frac{1}{2} \lVert f - f^* \rVert_\mathcal{D}^2$ the quadratic loss, with gradient $\nabla \ell_f = f - f^*$.
Assume that there exists $(a^*, \omega^*) \in \Theta$ such that $f^* = F(a^*, \omega^*)$, and $\omega^*$ is $\frac{\mu}{R}$-separated.

Then for all $(a, \omega) \in \Theta$ such that $\ell(F(a, \omega)) \neq 0$, $(\omega, \omega^*)$ is $\frac{\eta}{R}$-paired, and $\exists \alpha \in [0,1], \forall k, a_k^2 \geq \alpha$,
\[ \mathrm{R}\left(\overbar{K_{(a, \omega)}}; \nabla \ell_{F(a, \omega)}, \nabla \ell_{F(a, \omega)} \right) \geq \alpha \left(\phi(\eta) - \frac{1}{\mu - \eta}\right) \frac{(\kappa_0 - \rho_0)^2}{1 + \rho_0} \]
where with $\psi = - \sinc'$, $\phi = - \sinc''$,
and $H = \sum_{k \leq m} \frac{1}{k} \leq 1 + \log(m) \in \mathbb{R}_+$, the constants are
\[ \kappa_0 = \frac{\phi(\eta) - \frac{1}{\mu - \eta}}{\phi(0) + \frac{1}{\mu - \eta}} \quad\quad \rho_0 = \frac{\psi(\eta) + \frac{1}{\mu - \eta} + \frac{4H}{\mu - 2 \eta}}{\phi(\eta) - \frac{1}{\mu - \eta}} \]
Moreover, $\exists \mu_0 \in \mathbb{R}_+$, $\forall \mu > \mu_0$, $\exists \eta > 0$, s.t. $\kappa_0 > \rho_0$.
(non degeneracy if enough periods observed)
\end{proposition}

Proof in Appendix~\ref{appendix:sine-paired-proof},
leveraging Prop.~\ref{prop:variational} (variational bound) and Prop.~\ref{prop:shattering}.
This shows that when each frequency present in the signal is correctly estimated, then a gradient flow is well-suited for fine-tuning both frequencies and amplitudes. There are sufficiently few interactions to allow each neuron $(a_i, \omega_i)$ to descend towards its target $(a_i^*, \omega_i^*)$.
If the modelling hypothesis is verified (the target is a sum of sine waves), there is a finite and small number of neurons giving a sufficiently-parameterized system, and no need to go for vast overparamterization.
Letting the number of neurons tend to infinity is one way to ensure there is at least one neuron in each bassin, but not the only way.



\FinDiff{
\section{Conclusion}

We have shown that Kurdyka-\Loja{} inequalities can be leveraged to prove convergence of gradient flows to a loss of zero, even when the convergence speed is not linear.
In contrast, Polyak-\Loja{} inequalities granted by positive-definiteness of the neural tangent kernel only covered least-squares losses enjoying linear convergence speed.
Furthermore, we have shown that by focusing on lowering-bounding Rayleigh quotients rather than all eigenvalues at once, one can prove convergence even when the neural tangent kernel is not positive-definite, the most striking example being the finite-width infinite-data regime, where the neural tangent kernel must have null eigenvalues by rank deficiency.
We have provided several simple examples of such convergence proofs outside the vastly over-parameterized regime where there are more parameters than samples, along with tools and preliminary results that lead us to believe that obtaining the crucial Kurdyka-\Loja{} inequalities is feasible in more reasonable machine learning settings.

\section{Acknowledgements}

The authors would like to thank Thomas Le Corre, Lucas Weber, and Luca Ganassali for their help with various details of the proofs presented here, along with the anonymous reviewers, for their corrections and help in improving the readability of this work.
The authors acknowledge support from the French government under the management of the Agence Nationale de la Recherche as part of the “Investissements d’avenir” program, reference ANR-19-P3IA-0001 (PRAIRIE 3IA Institute). 
}

\bibliography{bibliography}
\bibliographystyle{unsrtnat}

\newpage
\appendix

\begingroup
\allowdisplaybreaks
\section{Appendix}


\RevDiff{
\subsection{Notation summary}\label{appendix:notation}

\renewcommand{\arraystretch}{1.6} 
\begin{table}[H]
\caption{Notations used in the main text}
\begin{tabularx}{\textwidth}{@{}XX@{}}
\toprule
  $\mathcal{X}$ & Input of the neural network (viewed as a set with no particular structure) \\
  
  $\mathcal{D}$ & Distribution over $\mathcal{X}$ (may have infinite support) \\
  
  $\mathcal{F} = \mathcal{X}^\mathbb{R}$ & Set of $\mathbb{R}$-valued functions on $\mathcal{X}$ \\

\midrule
  $\Theta = \mathbb{R}^d$ & Parameter space of a neural network \\
  $ \theta \in \Theta$ & Parameters (i.e.\ weights) of the neural network \\
  $ \theta_t  \in \Theta$ & Parameters at time $t \in \mathbb{R}_+$ when considering a gradient flow $\theta : \mathbb{R} \to \Theta$ \\
  $ \partial_t \theta_t  \in \Theta$ & Time-derivative of the parameters at time $t \in \mathbb{R}_+$ when considering a gradient flow $\theta : \mathbb{R} \to \Theta$ \\
  
  $F : \Theta \to \mathcal{F}, \> \theta \mapsto f_\theta$ & Network map, takes weights $\theta$ as input and produces a prediction function $f_\theta : \mathcal{X} \to \mathbb{R}$ as output \\
  
  $\d F_\theta : \Theta \to \mathcal{F}, \> \nu \mapsto \d F_\theta \cdot \nu$ & Network map differential at $\theta \in \Theta$, takes weight derivative $\nu$ as input and produces functional derivative $(\d F_\theta \cdot \nu) : \mathcal{X} \to \mathbb{R}$ as output \\
\midrule
  
  $\langle \cdot, \cdot \rangle_\mathcal{D} : \mathcal{F} \times \mathcal{F} \to \mathbb{R}$ &
  $\langle f, g \rangle_\mathcal{D} = \mathbb{E}_{x \sim \mathcal{D}} \left[ f(x) g(x) \right]$, see Definition~\ref{def:seminorm}. \\
  
  $\lVert \cdot \rVert_\mathcal{D} : \mathcal{F} \to \mathbb{R}_+$ &
  $\lVert f \rVert_\mathcal{D} = \sqrt{ \mathbb{E}_{x \sim \mathcal{D}} \left[ f(x)^2 \right] }$, see Definition~\ref{def:seminorm}. \\
  
  $\ell : \mathcal{F} \to \mathbb{R}_+$ & Functional loss \\
  
  $\mathcal{L} : \Theta \to \mathbb{R}_+$ & Parametric loss, $\mathcal{L} = \ell \circ F$. \\
\midrule
  
  $ K_\theta : \mathcal{X} \times \mathcal{X} \to \mathbb{R}$ & Neural Tangent Kernel (primal), see Def~\ref{def:NTK-primal} \\
  $ \quad K_\theta : (x,x') \mapsto \sum_i \partial_{\theta_i} F(\theta)(x) \, \partial_{\theta_i} F(\theta)(x') $ \\
  
  $ \overbar{ K_\theta } : \mathcal{F} \times \mathcal{F} \to \mathbb{R} $ & Bilinear form associated with the NTK (dual) \\
  $ \quad \overbar{K_\theta} : (f, g) \mapsto \mathbb{E}_{x, x' \sim \mathcal{D}} \left[ f(x) K_\theta(x, x') g(x') \right]$ \\
  
  $ \d F_\theta^\star : \Theta \times \mathcal{F} \to \mathbb{R}$ & Bilinear form associated with $\d F_\theta$ and $\mathcal{D}$ \\
  $ \quad \d F_\theta^\star(\nu, g) = \langle \d F_\theta \cdot \nu, g \rangle_\mathcal{D}$ \\
  
\midrule
  $\varphi : \mathbb{R}_+ \to \mathbb{R}$ & Desingularizing function, eases analysis of loss convergence in Proposition~\ref{prop:KLoja-direct}. \\
  
  $\d \varphi : \mathbb{R}_+ \to \mathbb{R}, \> u \mapsto \d \varphi_u$ & Derivative of the desingularizing function \\ 
  
  $R(A; x,y) = \frac{A(x,y)}{ \lVert x \rVert_V \lVert y \rVert_W }$  & Rayleigh quotient at $(x,y) \in V \times W$ of a bilinear map $A : (V, \lVert \cdot \rVert_V) \times (W, \lVert \cdot \rVert_W) \to \mathbb{R}$ \\
\bottomrule
\end{tabularx}
\end{table}

}

\pagebreak

\subsection{Details omitted from the main text}

\subsubsection{Functional loss gradients}\label{appendix:loss-gradient}

The use of the semi-norm $\lVert \cdot \rVert_\mathcal{D}$ on the functional space $\mathcal{F}$ comes with some apparent problems, for instance the gradient of the functional loss is not always defined (see Definition~\ref{def:functional-gradient}).
One solution is to work on a quotient $L^2(\mathcal{D}, \mathbb{R})$ of functions $\mathcal{D}$-almost everywhere identical on which $\lVert \cdot \rVert_\mathcal{D}$ can be strengthened to a norm.
We find this change of space sometimes prone to confusions, for it discards information outside the training region.
In the example of the lemniscate from Sec.~\ref{sec:lemniscate}, taking the quotient amounts to considering $\mathcal{F}$ to be the line $\mathbb{R} v$ instead of the plane $\mathbb{R}^2$. In particular, the notion of which minimum is reached becomes void because both are identical in the quotient, and the angle between the loss gradient and the lemniscate's tangent is no longer defined.

Instead, we observe that in all reasonable machine learning settings, it seems that the loss has a well-defined gradient with respect to $\inner{\mathcal{D}}$ anyway, see e.g. the following proposition

\begin{lemma}
Let $\mathcal{U} \subseteq \mathbb{R}$ and $\mathcal{V} \subseteq \mathbb{R}$ be intervals of $\mathbb{R}$.
If $\psi : \mathcal{U} \times \mathcal{V} \to \mathbb{R}_+$ is twice continuously differentiable, with derivative with respect to its first variable $\frac{\partial \psi}{\partial u} : \mathcal{U} \times \mathcal{V} \to \mathbb{R}$,
and if $\mathcal{D}$ is a distribution over $\mathcal{X} \subseteq \mathbb{R}^d$ with compact support,
then for any continuous $f^* : \mathcal{X} \to \mathcal{V}$, the loss

\[  \ell : f \mapsto \mathbb{E}_{x \sim \mathcal{D}} \left[ \psi(f(x), f^*(x)) \right] \]

is $\mathcal{D}$-compatible, (defined on functions $\mathcal{X} \to \mathcal{U}$ s.t. this expectation is finite), 
and if $f : \mathcal{X} \to \mathcal{U}$ is continuous, then $\ell$ is differentiable at $f$ and
the following is a gradient of $\ell$ at $f$ with respect to $\inner{\mathcal{D}}$

\[ \nabla \ell_f : x \mapsto \frac{\partial \psi}{\partial u}(f(x), f^*(x)) \]
\end{lemma}

\begin{proof}
$\mathcal{D}$-compatibility is immediate.
Let $f : \mathcal{X} \to \mathcal{U}$ be continuous, and let $\mathcal{U}_0 \subseteq \mathcal{U}$ be a closed interval
such that $f(x) \in \mathcal{U}_0$ holds $\mathcal{D}$-almost surely.
Then for all $g : \mathcal{X} \to \mathcal{U}$ such that $f(x) + g(x) \in  \mathcal{U}_0$ holds $\mathcal{D}$-almost surely, there exists $R : \mathcal{X} \times [0,1] \to \mathcal{R}$ such that for all $\varepsilon > 0$,
\[\begin{aligned}
\ell(f + \varepsilon g)
   &= \mathbb{E}_{x \sim \mathcal{D}} \left[ \psi\left( f(x) + \varepsilon g(x), f^*(x)\right) \right]
\\ &= \mathbb{E}_{x \sim \mathcal{D}} \left[ \psi(f(x), f^*(x)) + \frac{\partial \psi}{\partial u}\left(f(x), f^*(x) \right) \, \varepsilon g(x) + R(x, \varepsilon) \right]
\\ &= \ell(f) + \varepsilon \langle \nabla \ell_f,  g \rangle_{\mathcal{D}} + \mathbb{E}_{x \sim \mathcal{D}} \left[ R(x, \varepsilon) \right]
\end{aligned}\]
Moreover, the residual satisfies $R(x, \varepsilon) = {o}(\varepsilon)$ for all $x$. And for fixed $\varepsilon$, $R(x,\varepsilon)$ is bounded $\mathcal{D}$-almost surely, thus $\mathbb{E}_x\left[ R(x,\varepsilon) \right] = {o}(\varepsilon)$.
Taking the limit when $\varepsilon \to 0$ concludes the proof.
\end{proof}

For instance $\psi : (u,v) \mapsto (u-v)^2$, or $\psi : (u,v) \mapsto - v \log(u)$ if $\mathcal{U} = ]0,1]$, are relatively common.

In this work, "differentiable" is not taken to imply that the derivative is bounded, for simplicity in the exposition, to avoid dealing with finiteness of involved expectations, since theses issues are entirely orthogonal to our claims, and it is sufficient that a gradient exists for computations to carry out.

\subsubsection{Commutation of evaluation and derivation}\label{appendix:eval-deriv-commutation}

For differentiable network map functions $F : \Theta \to \mathcal{F}$, derivation with respect to the parameters in $\Theta$ can be carried out before or after evaluation at $x \in \mathcal{X}$.
Formally, if $\bar{\partial_\theta}: (\Theta \to \mathcal{F}) \to (\Theta \to \mathcal{F})$ and $\partial_\theta : (\Theta \to \mathbb{R}) \to (\Theta \to \mathbb{R}))$ are the (functional-valued and real-valued) derivation operators with respect to $\theta \in \Theta$, and if $\delta_x : (\Theta \to \mathcal{F}) \to (\Theta \to \mathbb{R})$ is the evaluation operator at some $x \in \mathcal{X}$ (i.e. $\delta_x(F) : \theta \mapsto F(\theta)(x)$ for all $F : \Theta \to \mathcal{F}$), then $\partial_\theta \circ \delta_x = \delta_x \circ \bar{\partial_\theta}$.

In exponent notation, with $\Theta = \mathbb{R}^d$, the network differential at $\theta \in \Theta$ is a linear function with signature $\d F_\theta : \mathbb{R}^d \to \mathbb{R}^\mathcal{X}$.
For finite $\mathcal{X}$, it corresponds to a rectangular matrix $\nabla F_\theta \in \mathbb{R}^{\mathcal{X} \times d}$ acting by usual matrix multiplication, with entries $(\partial_{\theta_j} F_\theta (x_i) \in \mathbb{R})$ for $x_i \in \mathcal{X}$ and $j \in [m]$, the partial derivative of the output with respect to the $j$-th parameter, evaluated at the $i$-th point of the dataset.

\subsubsection{Kudyka-\Loja{} proof (Proposition~\ref{prop:KLoja-direct})}\label{appendix:KL-direct-proof}

\begin{proof}
Since $\mathcal{L}(\theta_0) \neq 0$, let
$I \subseteq \mathbb{R}_+$ be an interval with $0 \in I$, such that $\forall t \in I, \mathcal{L}(\theta_t) > 0$.
Over $I$, it holds $\partial_t \left( \varphi \circ \mathcal{L}(\theta) \right) = \d (\varphi \circ \mathcal{L})_\theta \cdot \partial_t \theta = \d \varphi_{\mathcal{L}(\theta)} \d \mathcal{L}_\theta \cdot \partial_t \theta = - \d\varphi_{\mathcal{L}(\theta)} \left\langle \nabla \mathcal{L}(\theta), \nabla\mathcal{L}(\theta) \right\rangle \leq - \mu$.
Thus by integration, $\forall t \in I, \, \varphi(\mathcal{L}(\theta_t)) - \varphi(\mathcal{L}(\theta_0)) \leq - \mu \, t$.
The result over $I$ follows by inverting $\varphi$, and is extended to times $t \in \mathbb{R}_+$ such that $\mathcal{L}(\theta_t) = 0$ by noticing that $\forall v \in \mathrm{Im}(\varphi), 0 < \varphi^{-1}(v)$.
\end{proof}

\subsubsection{Kurdyka-\Loja{} details}\label{appendix:KL-details}

Two assumptions are somewhat hidden in Proposition~\ref{prop:KLoja-direct}.
If $\mathcal{L} : \mathcal{U} \to \mathbb{R}_+$ satisfies the Kurdyka-\Loja{} inequality
$\d \varphi_\mathcal{L} \langle \nabla \mathcal{L}, \mathcal{L} \rangle_\Theta \geq \mu$,
and if there exists a gradient flow $\theta : \mathbb{R}_+ \to \mathcal{U}$, then
$\inf \mathcal{L} = 0$ and $\varphi(u) \underset{u \to 0}{\longrightarrow} -\infty$.

Let $J = \mathrm{Im}(\varphi) \subseteq \mathbb{R}$. $J$ is an interval, by continuity of $\varphi$.
By Proposition~\ref{prop:KLoja-direct}, for all $t \in \mathbb{R}_+$, it holds $\varphi(\mathcal{L}(\theta_t)) \leq \varphi(\mathcal{L}(\theta_0)) - \mu t$.
Therefore $\inf(J) \leq \varphi(\mathcal{L}(\theta_0)) - \mu t \rightarrow - \infty$ when $t \to + \infty$,
hence $\inf(J) = - \infty$.
Since $\varphi : \mathbb{R}^*_+ \to J$ is strictly increasing, this implies that $\varphi(u) \underset{u \to 0}{\longrightarrow} - \infty$.

By the same proposition, it follows that $\mathcal{L}(\theta_t) \leq \varphi^{-1}\left( \varphi(\mathcal{L}(\theta_0)) - \mu t \right)$. But since it holds that $\varphi(\mathcal{L}(\theta_0)) - \mu t \rightarrow - \infty$ when $t \to +\infty$, we conclude that $\mathcal{L}(\theta_t) \rightarrow 0$, in particular $(\inf \mathcal{L}) = 0$.

While these may be viewed as restrictions of the applicability of Proposition~\ref{prop:KLoja-direct}, we claim that the proof and general ideas are simple enough to be straightforwardly extended to any related setting, the most important part is that the statement is sufficiently clear to convey the idea for the proof.

\subsection{Proofs omitted from the main text}

\subsubsection{Proof of composition property (Proposition~\ref{prop:KLoja-comp})}\label{proof:composition}

\begin{proof}[Proof of Proposition~\ref{prop:KLoja-comp}]
Let $\theta \in \Theta$, and $f_\theta = F(\theta) \in \mathcal{F}$, such that $\mathcal{L}(\theta) \neq 0$.
Let us show that
$$ \langle \nabla \mathcal{L}(\theta), \nabla \mathcal{L}(\theta) \rangle_\Theta
= \langle \nabla \ell_{f_\theta}, \nabla \ell_{f_\theta} \rangle_\mathcal{D} \, \mathrm{R}\left(\overbar{K_\theta}; \nabla \ell_{f_\theta} , \nabla \ell_{f_\theta} \right) $$

First, the right-hand side is well-defined because $\mathcal{L}(\theta) \neq 0$ implies $\lVert \nabla \ell_{F(\theta)} \rVert_\mathcal{D}^2 \neq 0$. Indeed, $\ell(f_\theta) = \mathcal{L}(\theta) \neq 0$, therefore $\d \varphi_{\ell(f_\theta)} \lVert \nabla \ell_{f_\theta} \rVert_\mathcal{D}^2 \geq 1$, however $\varphi$ is strictly increasing, so $\d \varphi_{\ell(f_\theta)} > 0$.

Since $\mathcal{L} = \ell \circ F$, it follows that $\nabla \mathcal{L}(\theta) = \mathbb{E}_{x \sim \mathcal{D}} \left[ \nabla F_\theta(x) \, \nabla \ell_{F(\theta)}(x) \right]$.
Therefore
$$\begin{aligned}
\langle \nabla \mathcal{L}(\theta), \nabla \mathcal{L}(\theta) \rangle_\Theta
&= \left\langle \mathbb{E}_{x \sim \mathcal{D}} \left[ \nabla F_\theta(x) \nabla \ell_{F(\theta)} (x) \right], \mathbb{E}_{x' \sim \mathcal{D}} \left[ \nabla F_\theta(x') \nabla \ell_{F(\theta)} (x') \right] \right\rangle_\Theta
    & (1)
\\ &= \mathbb{E}_{x \sim \mathcal{D}, x' \sim \mathcal{D}} \left[  \nabla \ell_{F(\theta)} (x) \left\langle \nabla F_\theta(x), \nabla F_\theta(x') \right \rangle_\Theta \nabla \ell_{F(\theta)} \right]
& (2)
\\ &= \mathbb{E}_{x \sim \mathcal{D}, x' \sim \mathcal{D}} \left[  \nabla \ell_{F(\theta)} (x) K_\theta(x, x') \nabla \ell_{F(\theta)} \right]
& (3)
\\ &= \overbar{K_\theta}\left(\nabla \ell_{F(\theta)}, \nabla \ell_{F(\theta)}\right)
& (4)
\\ &= \frac{ \overbar{K_\theta}\left(\nabla \ell_{F(\theta)}, \nabla \ell_{F(\theta)}\right) }{\langle \nabla \ell_{F(\theta)}, \nabla \ell_{F(\theta)} \rangle_\mathcal{D}} \, \langle \nabla \ell_{F(\theta)}, \nabla \ell_{F(\theta)} \rangle_\mathcal{D}
& (5)
\\ &= R\left( \overbar{K_\theta}; \nabla \ell_{F(\theta)}, \nabla \ell_{F(\theta)} \right) \, \langle \nabla \ell_{F(\theta)}, \nabla \ell_{F(\theta)} \rangle_\mathcal{D}
& (6)
\end{aligned}$$
Where $(1)$ is by definition of $\nabla \ell$ (Def~\ref{def:functional-gradient}) as cited above, $(2)$ by linearity, $(3)$ by definition of $K(\theta)$ (Def~\ref{def:NTK-primal}a), $(4)$ by definition of $\overbar{K_\theta}$ (Def~\ref{def:NTK-primal}b), $(5)$ is well defined because $\mathcal{L}(\theta) \neq 0$, and $(6)$ is by definition of $\mathrm{R}$ (Def~\ref{def:rayleigh-quotient}).
The result follows immediately by multiplying both sides by $\d \varphi_{\mathcal{L}(\theta)}$.
\end{proof}

\subsubsection{Proof of variational bound (Proposition~\ref{prop:variational})}\label{appendix:variational}

\begin{proof}[Proof of Proposition~\ref{prop:variational}]
By the variational form of the $\ell_2$-norm induced by the inner product on $\Theta$.
$$\begin{aligned}
\overbar{K_\theta}( h, h )
&= \mathbb{E}_{x \sim \mathcal{D}, x' \sim \mathcal{D}} \left[ h(x) (\nabla F_\theta(x) \cdot \nabla F_\theta(x')) h(x') \right]
\\ &= \left\lVert \mathbb{E}_{x \sim \mathcal{D}} \left[ \nabla F_\theta(x) h(x) \right] \right\rVert_\Theta^2
\\ &= \sup_{\nu \in \Theta \setminus\{0\}} \frac{1}{ \lVert \nu \rVert_\Theta^2 } \langle \nu, \mathbb{E}_{x \sim \mathcal{D}} \left[ \nabla F_\theta(x) h(x) \right] \rangle_\Theta^2
\\ &= \sup_{\nu \in \Theta \setminus\{0\}} \frac{1}{ \lVert \nu \rVert_\Theta^2 } \langle \d F_\theta \cdot \nu, h \rangle_\mathcal{D}^2 
\end{aligned}$$
It then suffices to divide both sides by $\langle h, h\rangle_\mathcal{D} = \lVert h \rVert_\mathcal{D}^2 \neq 0$.
\end{proof}

\subsubsection{Proof of cosine-singular split (Proposition~\ref{prop:cosine-singular-split})}\label{appendix:split-cosine-singular}

\begin{proof}[Proof of Proposition~\ref{prop:cosine-singular-split}]
If $\lambda = \inf_{\nu \in \Theta_0 \setminus \{0\}} \lVert \d F_\theta \cdot \nu \rVert_\mathcal{D}^2 / \lVert \nu \rVert_\Theta^2 = 0$, then the result is immediate because $\mathrm{R}(\overbar{K_\theta}; h,h) \geq 0$ by positive semi-definiteness. Thus, assume $\lambda > 0$.
$$ \begin{aligned}
\mathrm{R}\left(\overbar{K_\theta}; h, h\right)
& = \sup_{\nu \in \Theta \setminus \{0\}} \mathrm{R}(\overbar{\d F_\theta} ; \nu, h)^2 &(1)
\\ &= \sup_{\nu \in \Theta \setminus \{ 0 \}} \frac{ \langle \d F_\theta \cdot \nu, h \rangle_\mathcal{D}^2}{\lVert \nu \rVert_\Theta^2 \lVert h \rVert_\mathcal{D}^2}
& (2)
\\ &\geq \sup_{\nu \in \Theta_0 \setminus \{ 0 \}} \frac{ \langle \d F_\theta \cdot \nu, h \rangle_\mathcal{D}^2}{\lVert \nu \rVert_\Theta^2 \lVert h \rVert_\mathcal{D}^2}
& (3)
\\ &= \sup_{\nu \in \Theta_0 \setminus \{ 0 \}} \frac{ \langle \d F_\theta \cdot \nu, h \rangle_\mathcal{D}^2}{\lVert \d F_\theta \cdot \nu \rVert_\mathcal{D}^2 \lVert h \rVert_\mathcal{D}^2} \frac{ \lVert \d F_\theta \cdot \nu \rVert_\mathcal{D}^2}{\lVert \nu \rVert_\Theta^2}
& (4)
\\ &\geq \left(\sup_{\nu \in \Theta_0 \setminus \{ 0 \}} \frac{ \langle \d F_\theta \cdot \nu, h \rangle_\mathcal{D}^2}{\lVert \d F_\theta \cdot \nu \rVert_\mathcal{D}^2 \lVert h \rVert_\mathcal{D}^2} \right) \left( \inf_{\nu \in \Theta_0 \setminus \{0\}} \frac{ \lVert \d F_\theta \cdot \nu \rVert_\mathcal{D}^2}{\lVert \nu \rVert_\Theta^2} \right)
\geq \mu^2 \lambda
&(5)
\end{aligned} $$
where $(1)$ is Prop~\ref{prop:variational}, $(2)$ the definition of $\mathrm{R}$, $(3)$ because the supremum is increasing with respect to inclusion, $(4)$ is well-defined because $\lambda > 0$, and $(5)$ is a uniform bound on the second factor.
\end{proof}

\subsubsection{Proof of approximate SVD (Proposition~\ref{prop:shattering})}\label{appendix:approx-svd}

\begin{proof}[Proof of Proposition~\ref{prop:shattering}]
Since $h \in \mathrm{Span}(g)$, let $u \in \mathbb{R}^k$ such that $h = \sum_i \frac{u_i}{\lVert g_i \rVert_\mathcal{D}} g_i$.
Then let $\rho = \min_i \lVert \mathrm{d} F_\theta \cdot a_i \rVert_\mathcal{D} / \lVert a_i \rVert_\Theta$. If $\rho = 0$ then the proposition is verified: let $\nu \in \mathrm{Span}(a) \setminus \{0\}$, observe either $\mathrm{R}(\overbar{\d F_\theta}; \nu, h) \geq 0$, which satisfies the property, or $\mathrm{R}(\overbar{\d F_\theta}; -\nu, h) = - \mathrm{R}(\overbar{\d F_\theta}; \nu, h) > 0$. Therefore assume in the following that $\rho > 0$.
Let $v \in \mathbb{R}^k$ be $v_i = u_i \lVert a_i \rVert_\Theta / \lVert \d F_\theta \cdot a_i \rVert_\mathcal{D}$.
$$ \begin{aligned}
&\max_{\nu \in \mathrm{Span}(a) \setminus \{0\}} \mathrm{R}( \overbar{\d F_\theta}; \nu, h)
\\ & \geq \mathrm{R}\left( \overbar{\d F_\theta}; \sum_i \frac{u_i}{\lVert \d F_\theta a_i \rVert_\mathcal{D}} a_i, h \right)
&(1)
\\ &= \frac{\sum_{i,j} \frac{u_i}{\lVert \d F_\theta \cdot a_i \rVert_\mathcal{D}} \frac{u_j}{\lVert g_j \rVert_\mathcal{D}} \langle \d F_\theta \cdot a_i, g_j \rangle_\mathcal{D}}{ \sqrt{ \sum_{i,j} \frac{u_i}{\lVert a_i \rVert_\Theta} \frac{\lVert a_i \rVert_\Theta}{\lVert \d F_\theta \cdot a_i \rVert_\mathcal{D}} \frac{u_j}{\lVert a_j \rVert_\Theta} \frac{\lVert a_j \rVert_\Theta}{\lVert \d F_\theta \cdot a_j \rVert_\mathcal{D}} \langle a_i, a_j \rangle_\Theta} \sqrt{ \sum_{i,j} \frac{u_i}{\lVert g_i \rVert_\mathcal{D}} \frac{u_j}{\lVert g_j \rVert_\mathcal{D}} \langle g_i, g_j \rangle_\mathcal{D}} } 
&(2)
\\ &= \frac{\sum_{i,j} u_i u_j \mathrm{R}(\inner{\mathcal{D}}; \d F_\theta \cdot a_i, g_j) }{ \sqrt{ \sum_{i,j}  v_i v_j \mathrm{R}(\inner{\Theta}; a_i, a_j)} \sqrt{ \sum_{i,j} u_i u_j \mathrm{R}(\inner{\mathcal{D}}; g_i, g_j)  } }
&(3)
\\ &\geq \frac{\lambda_{\min}(\mathrm{R}(\inner{\mathcal{D}}; \d F_\theta \cdot a_i, g_j)) \lVert u \rVert_2^2 }{ \sqrt{ \lambda_{\max}(\mathrm{R}(\inner{\Theta}; a_i, a_j)) \lVert v\rVert_2^2 } \sqrt{ \lambda_{\max}(\mathrm{R}(\inner{\mathcal{D}}; g_i, g_j)) \lVert u \rVert_2^2 } }
&(4)
\\ &\geq \frac{\lambda_{\min}(\mathrm{R}(\inner{\mathcal{D}}; \d F_\theta \cdot a_i, g_j)) \> \min_{i \in [k]} \lVert \mathrm{d} F_\theta \cdot a_i \rVert_\mathcal{D} / \lVert a_i \rVert_\Theta}{ \sqrt{ \lambda_{\max}(\mathrm{R}(\inner{\Theta}; a_i, a_j)) } \sqrt{ \lambda_{\max}(\mathrm{R}(\inner{\mathcal{D}}; g_i, g_j)) } }
&(5)
\end{aligned} $$
where $(1)$ is evaluation of the variational form, $(2)$ by definition of $\mathrm{R}$ and bilinearity, $(3)$ is a reorganization by bilinearity, $(4)$ by definition of $(\lambda_{\min}, \lambda_{\max})$, and $(5)$ by using $\lVert v \rVert_2 \leq \frac{1}{\rho} \lVert u \rVert_2$.
\end{proof}

\subsection{Computations for the lemniscate}

\subsubsection{Convergence on the lemniscate}\label{appendix:lemniscate-proof}

\paragraph{Proof sketch for Proposition~\ref{prop:lemniscate-convergence}.}
As previously, the quadratic loss satisfies a Polyak-\Loja{} property $(\lVert \nabla \ell \rVert_2^2 \geq 4 \ell)$ that will grant linear convergence provided we can show a lower bound for $\mathrm{R}(\overbar{K_\theta}; \nabla \ell_{F(\theta)}, \nabla \ell_{F(\theta)})$ for all $\theta(t)$ for each parameterization, which we will achieve by a variational bound (Proposition~\ref{prop:variational}), then splitting the variational term according to Proposition~\ref{prop:cosine-singular-split}.

We start by computing in closed form the differentials of each parameterization.
\[ \nabla F_S(\theta) = \left(- \sin(\theta) \frac{ \RevDiff{(}(1 + \sin^2(\theta)) \RevDiff{+} 2 \cos^2(\theta)\RevDiff{)}}{(1 + \sin^2(\theta))^2}, \frac{- \sin^4(\theta) - \sin^2(\theta) + (1 - \sin^2(\theta)) \cos^2(\theta)}{(1 + \sin^2(\theta))^2} \right)  \]

\[ \nabla F_L(\theta) = \frac{1}{(\theta^4 + 6 \theta^2 + 1)^2} \left( -4 \theta (3 \theta^4 + 2 \theta^2 + 3), 2(\theta^6 - 9 \theta^4 - 9 \theta^2 + 1) \right) \]

Without loss of generality, assume $v \geq 0$ (by symmetry).
Now for both parameterizations, we need to study several functions from $\mathbb{R}$ to $\mathbb{R}$.
By Proposition~\ref{prop:variational} then Proposition~\ref{prop:cosine-singular-split},
\[\begin{aligned}
\mathrm{R}(\overbar{K_{S,\theta}}; \nabla \ell_{F_S(\theta)}, \nabla \ell_{F_S(\theta)})
&= \sup_{\nu \in \mathbb{R} \setminus\{0\}} \mathrm{R}(\overbar{\d F_S(\theta)}; \nu, \nabla \ell_{F_S(\theta)})^2
\\ &= \sup_{\nu \in \mathbb{R} \setminus\{0\}} \mathrm{R}(\inner{\mathbb{R}^2}; \d F_s(\theta) \cdot \nu, \nabla \ell_{F_S(\theta)})^2 \times \frac{\lVert \d F_S(\theta) \cdot \nu \rVert_2^2}{\lVert \nu \rVert_2^2}
\\ &= \mathrm{R}(\inner{\mathbb{R}^2}; \nabla F_s(\theta), \nabla \ell_{F_S(\theta)})^2 \times \lVert \nabla F_S(\theta) \rVert_2^2
\end{aligned}\]

Observe that $\nabla \ell(a,b) = 2 (a u + b v - y) \left( u, v \right) \in \mathbb{R}^2$.
By hypothesis, if $(a,b) = \theta(t)$ then $(a u + bv - y) \geq 0$ (for both $\theta = \theta_S$ and $\theta = \theta_L$) because this quantity is
positive at initialization and cannot change signs (if it becomes null, the loss is null and the flow stops).

Let $\theta_S^* = \min \{ \theta \,|\, \ell(F_S(\theta)) = 0 \}$ and $\theta_L^* = \min \{ \theta \,|\, \ell(F_L(\theta)) = 0 \}$ be the first zeros of each loss on $\mathbb{R}_+$.
Let $\mu_S : \theta \mapsto 
\mathrm{R}(\inner{\mathbb{R}^2}; \nabla F_S(\theta), -\nabla\ell_{F(\theta)})$ (respectively $\mu_L$).
Show by computation that there exists $\mu_0 \in \mathbb{R}_+^*$ such that $\mu_S(\theta) \geq \mu_0$ for all $\theta \in [0, \theta_S^*[$. This constant is not dependent of the parameterization because if $F_S(\theta) = F_L(\nu)$ then $\d F_S(\theta) \in \mathbb{R}_+^* \cdot \d F_L(\nu)$. Thus $\mu_L(\theta) \geq \mu_0$ for all $\theta \in [0, \theta_L^*[$.
Moreover, $\partial_t \theta_S(t)$ and $\mu_S(\theta_S(t))$ have same sign, and $\theta_S(0) = 0$, so $\theta_S$ is increasing over time (respectively $\partial_t \theta_L \geq 0$ for the other parameterization). See Fig.~\ref{fig:lem-decomp} for an illustration.

Now let $\lambda_S : \theta \mapsto \lVert \mathrm{d} F_S(\theta) \rVert_2^2$ be the corresponding singular value for $F_S$ (respectively $\lambda_L : \theta \mapsto \lVert \mathrm{d} F_L(\theta) \rVert_2^2$ for $F_L$).
Observe that $\lambda_S$ is bounded below on $[0, \theta_s^*[$ (respectively $\lambda_L$ on $[0, \theta_L^*[$).
Conclude by lower-bounding the (positive) product with the product of the (positive) lower-bounds.

\begin{figure}[H]
\centering
\begin{subfigure}{.5\textwidth}
\centering
\includegraphics[width=\textwidth]{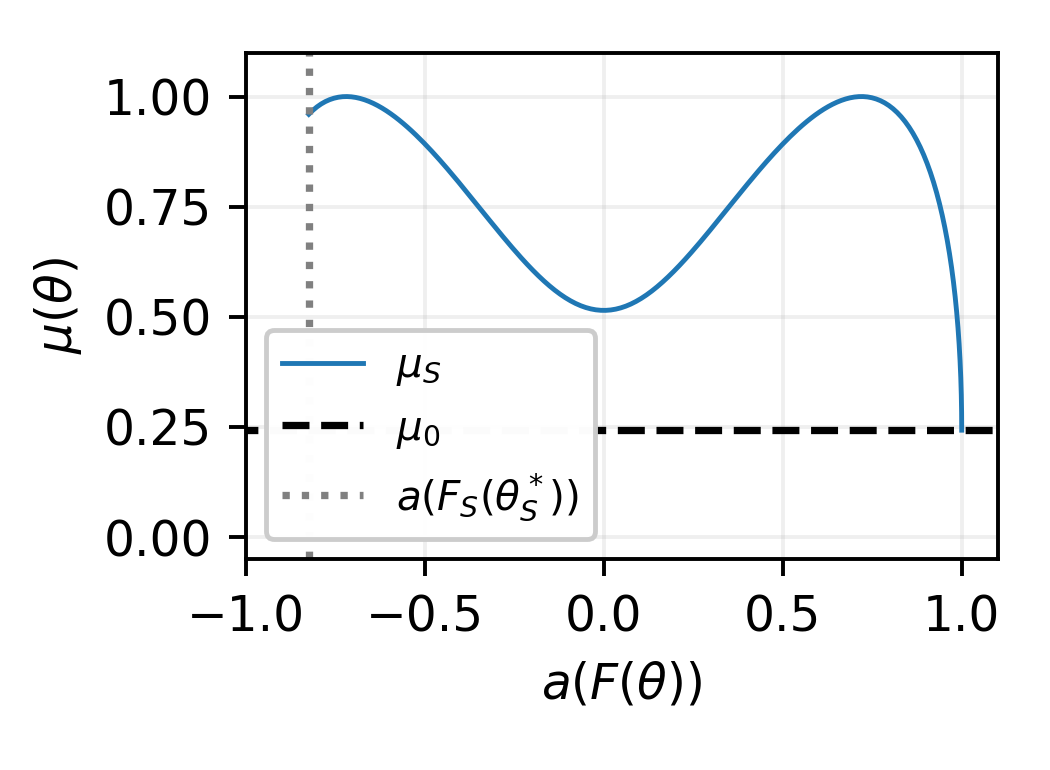}
\subcaption{Angle between tangent and gradient}%
\label{fig:lem-angles}
\end{subfigure}%
\begin{subfigure}{.5\textwidth}
\centering
\includegraphics[width=\textwidth]{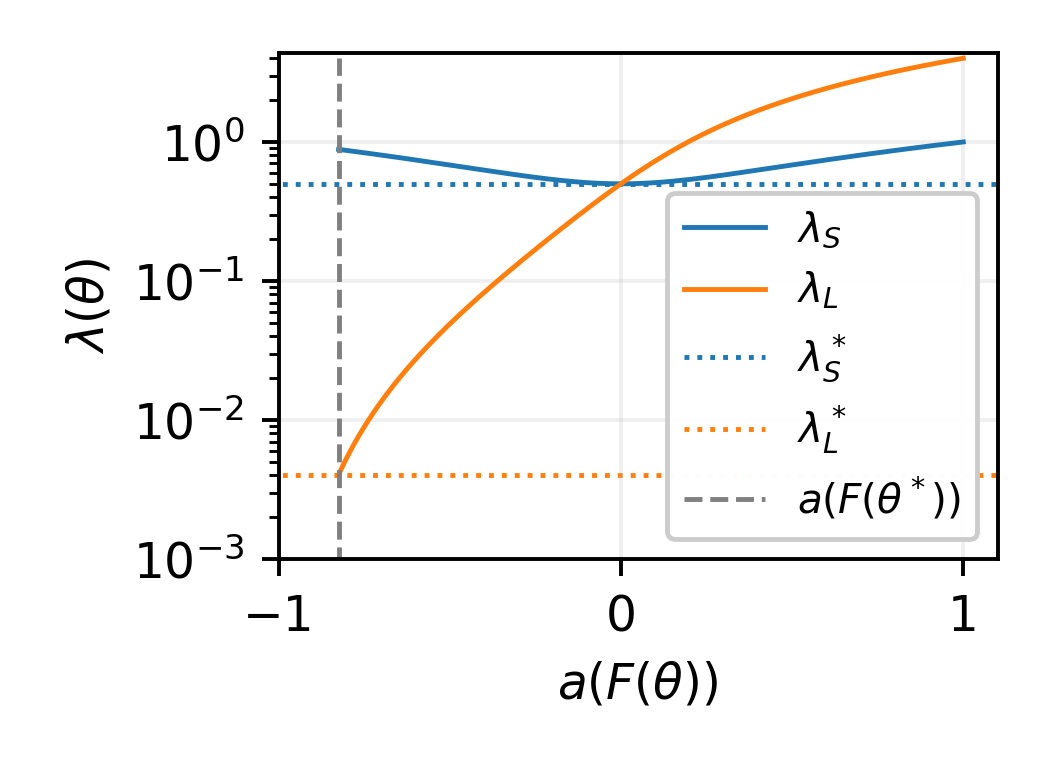}
\subcaption{Singular value of lemniscate parameterization}%
\label{fig:lem-singular}
\end{subfigure}
\caption{Decomposition of the bound for $(u,v,y) = (4,-1,-3)$, as for Fig.~\ref{fig:2d-lemniscate}, for $\theta \geq 0$}%
\label{fig:lem-decomp}
\end{figure}

The additional properties that $\lambda_S^* \geq \frac{1}{2}$ and $\lambda_L^* = \lVert \nabla F_L(\theta_L^*) \rVert_2^2$ are depicted on Fig.~\ref{fig:lem-singular}.

\subsubsection{Convergence speed details predictable from the Rayleigh quotient}

We depict in Fig.~\ref{fig:lem-triplot} the gradient flow for the "sphere to lemniscate" ($F_S$) parameterization
(already depicted in Fig.~\ref{fig:gd-speed-short}),
and show that the slowdowns observed in the decrease of the loss correspond to the points at which the gradient of the loss is less aligned with the lemniscate's tangent (corresponding to low values of $\mu_S$).
This is because the Rayleigh quotient is $\mathrm{R}(\overbar{K_\theta}; \nabla \ell_{F_S(\theta)}, \ell_{F_S(\theta)}) = \mu_S(\theta)^2 \, \lambda_S(\theta)$, and the singular-value factor $\lambda_S$ is almost constant, as can be seen on Fig.~\ref{fig:lem-singular}.

\begin{figure}[H]
    \centering
    \includegraphics[width=.9\textwidth]{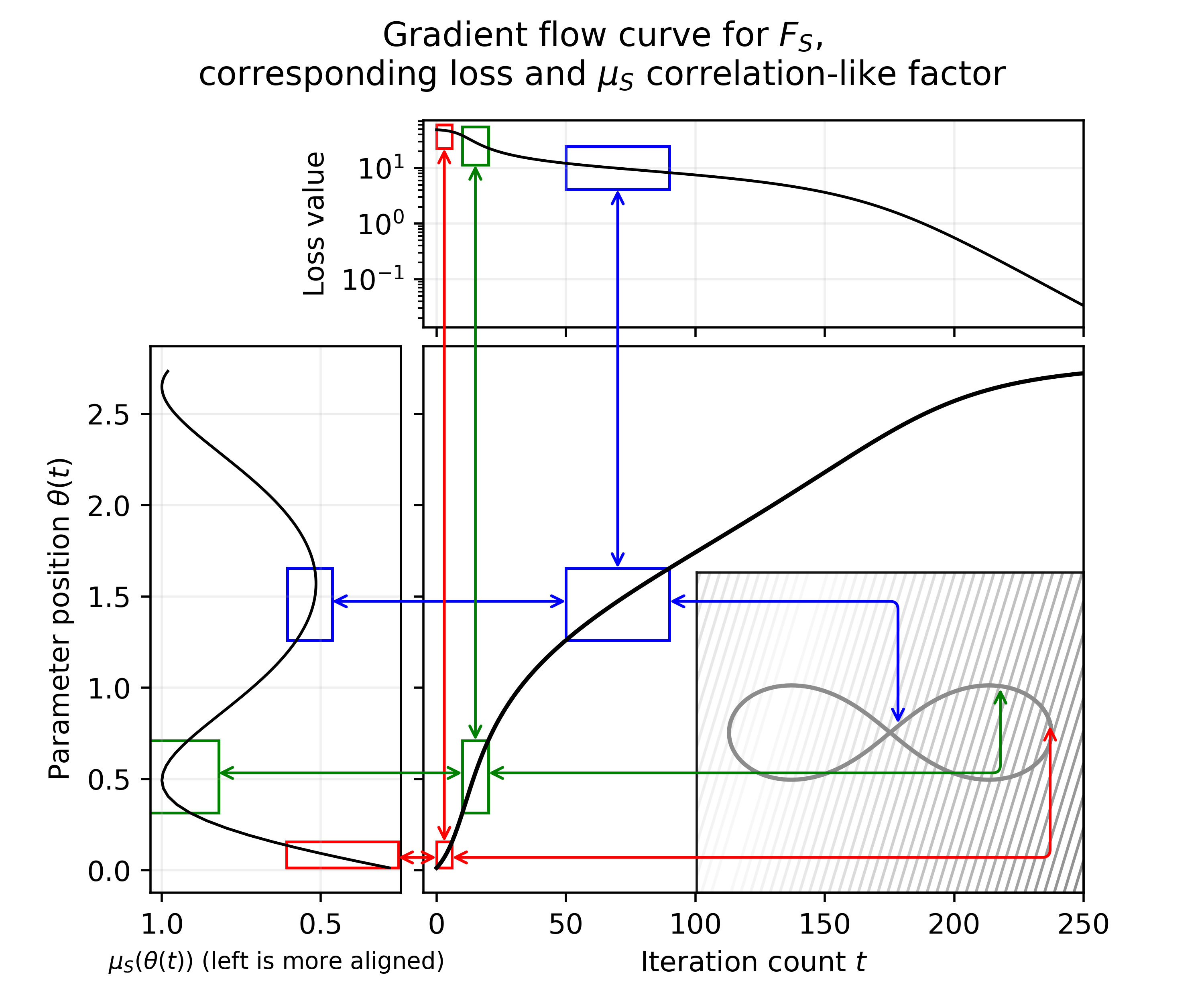}
    \caption{Alignement of gradient and lemniscate's tangent, with consequences on convergence speed
    (best viewed in color). Red (bottom-most) and blue (top-most) regions correspond to low $\mu_S$ and slowdowns in the loss decrease, Green (middle) region corresponds to higher $\mu_S$ and an acceleration.}%
    \label{fig:lem-triplot}
\end{figure}

\subsection{Computations for logistic regression}

\subsubsection{Pointwise gradients}\label{appendix:logistic-gradient}

We use the notations of section~\ref{sec:logistic-regression}.
Let $x \in \mathcal{X}$ and the corresponding label $y \in \Delta_c$.
Recall $\ell_x = H_x \circ E$, where $H_x : p \in \Delta_c \mapsto - \sum_i y_i \log (p_i) $.
Let $u \in \mathbb{R}^c$. Let us show that $\nabla \ell_x(u) = E(u) - y$.

\begin{proof}
The derivative of $H_x$ is straightforward
$$ \frac{\partial H_x}{\partial p_i}(p) = - \frac{y_i}{p_i} $$

The derivative of the $i$-th coordinate of softargmax $E_i : u \mapsto \exp(u_i) / \sum_j \exp(u_j)$ is
$$\begin{aligned}
\frac{\partial E_i}{\partial u_j}(u) = \frac{\delta_{i=j} \exp(u_i) \sum_k \exp(u_k) - \exp(u_i) \exp(u_j) }{\left( \sum_k \exp(u_k) \right)^2} = \delta_{i = j} E_i(u) - E_i(u) E_j(u)
\end{aligned}$$
The result follows by chain rule, using $\sum_{i \in [c]} y_i = 1$
$$ \frac{\partial \ell_x}{\partial u_i}(u) = \sum_j \frac{\partial H_x}{\partial p_j}(E(u)) \frac{\partial E_j}{\partial u_i}(u) = \sum_j - \frac{y_j}{E_j(u)} \left( \delta_{i = j} E_j(u) - E_i(u) E_j(u) \right) = E_i(u) - y_i $$
\end{proof}

\subsubsection{Separating ray with zero loss implies dirac labels }\label{appendix:logistic-delta}

As a first step, consider the following lemma.
Let $x \in \mathcal{X}$ and $H_x : p \in \Delta_c \mapsto - \sum_i y_i \log(p_i)$.
If $p : \mathbb{N} \to \Delta_c$ is a sequence converging to $q \in \Delta_c$ such that $H_x(p(k)) \to 0$, then $q = y$ and $\exists i, \forall j,  y_j = \delta_{i = j}$.
To prove this by contradiction, assume there exists $i \neq j$ such that $y_i \neq 0$ and $y_j \neq 0$.
Since $H_x(q) < +\infty$, it holds $q_i \neq 0$ and $q_j \neq 0$, thus $\max(q_i, q_j) < 1$ and $H_x(q) \geq - y_i \log(q_i) - y_j \log(q_j) \geq - (y_i + y_j) \log(\max(q_i, q_j)) > 0$ which contradicts $H_x(p(k)) \to 0$, thus $y$ is a dirac. Finally, if $i \in [c]$ is such that $y_i = 1$, then $H_x(p_x(k)) \to 0$ implies $q_i = 1$.

It remains to show that this holds for $\mathcal{D}$-almost all responses $y$.
Let $\zeta \in \mathbb{R}^{c \times d}$ be an $\varepsilon$-margin separating ray satisfying $\inf_\lambda \mathcal{L}(\lambda \zeta) = 0$.
Let $\lambda : \mathbb{N} \to \mathbb{R}$ be a sequence such that $\mathcal{L}(\lambda_k \zeta) \underset{k \to +\infty}{\longrightarrow} 0$.

For $x \in \mathcal{X}$, the sequence $k \mapsto E(\lambda_k \zeta \cdot x)$ has values in $\Delta_c$, which is compact. Hence extract from it a convergent sequence $(p_x(k))_{k \in \mathbb{N}}$.
Then, $(H_x(p_x(k)))_k$ is a sequence of positive random variables converging in expectation to zero, therefore up to extraction of another subsequence, it converges almost surely to zero \RevDiff{\citep[see e.g.][Theorem 3.4, page 212]{gut2013probability}}.
Thus, it holds almost surely that $y$ is a dirac and $p_x(k) \to y$.
Moreover, for $\mathcal{D}$-almost all $x \in \mathcal{X}$, there exists $i^* \in [c]$ such that for all $j \in [c]$, $\langle \zeta_{i^*}, x \rangle \geq \langle \zeta_j, x \rangle$, hence $p_{x,i^*}(k) \geq p_{x,j}(k)$, which implies $y = (\delta_{j = i^*})_{j \in [c]}$.

\subsubsection{Proof of convergence speed for logisitic regression}\label{appendix:logistic-regression}

For $\mathcal{D}$-almost all $x \in \mathcal{X}$, let $\ell_x = H_x \circ E : \mathbb{R}^c \to \mathbb{R}_+$. For $u \in \mathbb{R}^c$, by a simple calculation, this has gradient $\nabla \ell_x(u) = E(u) - y$ (see appendix~\ref{appendix:logistic-gradient}).
Then, define $\ell : (\mathcal{X} \to \mathbb{R}^c) \to \mathbb{R}_+$, as $\ell(u) = \mathbb{E}_{x \sim \mathcal{D}}[ \ell_x(u(x)) ]$. Observe that $\nabla \ell(u) : x \mapsto \nabla \ell_x(u(x))$ is a gradient for $\ell$, and $\mathcal{L} = \ell \circ X$.
Therefore, we can apply the variational bound to try to get a Kurdyka-\Loja{} property.

$$ \lVert \nabla \mathcal{L}(\theta) \rVert_\Theta^2 = \sup_{\nu \in \Theta} \, \langle \nu, \nabla \mathcal{L}(\theta) \rangle_\Theta^2 / \lVert \nu \rVert_\Theta^2 = \sup_{\nu \in \Theta} \, \langle X(\nu), \nabla \ell(u) \rangle_\mathcal{D}^2 / \lVert \nu \rVert_\Theta^2 $$

We can then evaluate at a well-chosen point ($\nu = \zeta \in \Theta$).
For $\mathcal{D}$-almost all $x \in \mathcal{X}$, define $i^* = \argmax_i \langle \zeta, x \rangle$, together with $M_x = \langle \zeta_{i^*}, x \rangle$ and $m_x = \max_{i \neq i^*} \langle \zeta_i, x \rangle$. By the $\varepsilon$-margin separability assumption, it holds $M_x \geq m_x + \varepsilon \lVert \zeta \rVert_\Theta$. Therefore, with the notation $p_{x,i} = E(u(x))_i$

$$ \begin{aligned}
\langle X(\zeta), y - p \rangle_\mathcal{D}
   &= \mathbb{E}_x \left[ \sum_{i \in [c]} \langle \zeta_i, x \rangle \, (y_{x,i} - p_{x,i}) \right]
   = \mathbb{E}_x \left[ M_x (1 - p_{x,i^*}) - \sum_{i \neq i^*} \langle \zeta_i, x \rangle \, p_{x,i} \right]
   &(1)
\\ &\geq \mathbb{E}_x \left[ M_x (1 - p_{x,i^*}) - m_x \sum_{i \neq i^*} p_{x,i} \right]
   = \mathbb{E}_x \left[ (M_x - m_x) (1 - p_{x,i^*}) \right]
\\ &\geq \varepsilon \lVert \zeta \rVert_\Theta \mathbb{E}_x \left[ 1 - p_{i^*} \right]
   = \varepsilon \lVert \zeta \rVert_\Theta \mathbb{E}_x \left[ 1 - e^{- \ell_x(u(x))} \right]
   \geq \varepsilon \kappa \lVert \zeta \rVert_\Theta \left( 1 - e^{- \ell(u)} \right)
   &(2)
\end{aligned} $$
where $(1)$ is because $(\inf_\lambda \mathcal{L}(\lambda \zeta) = 0)$ implies $y_{x,i} = \delta_{i = i^*}$ (see appendix \ref{appendix:logistic-delta}), and $(2)$ is by $\varepsilon$-margin separability assumption then definition of $\ell_x$ and finally $\mathbb{E}[\psi(Z)] \geq \mathbb{P}(Z \geq \mathbb{E}[Z])\, \psi(\mathbb{E}[Z])$ for any non-negative random variable $Z$ since $\psi : z \in \mathbb{R}_+ \mapsto 1 - e^{-z}$ is increasing and non-negative.
The final result follows from $ \d \varphi_{\mathcal{L}(\theta)} \, \lVert \nabla \mathcal{L}(\theta) \rVert_\Theta^2 \geq \varepsilon^2 \kappa^2$,
by integration of $\d \varphi_z = \left( 1-e^{-z}\right)^{-2}$ to get $\varphi : z \in \mathbb{R}_+^* \mapsto \log(e^{+z} - 1) - (e^{+z} - 1)^{-1}$ and thus $\varphi^{-1} : u \in \mathbb{R} \mapsto \log(1 + 1 / W_0(e^{-u}))$.

\subsubsection{Logistic bound asymptotic behavior}\label{appendix:logistic-bound-asymptote}

\RevDiff{
We show here that the convergence bound for the logistic regression presented in Proposition~\ref{prop:logistic-convergence} is consistent with the previously-known asymptotic $\mathcal{O}(1/t)$ behavior.

Let $(C, \tau) \in \mathbb{R} \times \mathbb{R}_+^*$ and $f : t \mapsto \log\left( 1 + \dfrac{1}{ W_0\left( \exp\left( t / \tau - C \right) \right)} \right)$.
Let us show $f(t) \underset{+\infty}{=} \mathcal{O}(1/t)$.

As warmup, note that $\exp(t / \tau - C) \underset{t \to +\infty}{\longrightarrow} + \infty$, and $W_0(x) \underset{x \to +\infty}{\longrightarrow} + \infty$, therefore $f(t) \underset{t \to +\infty}{\longrightarrow} 0$.

From \citet[Theorem 2.1]{Hoorfar08inequalities}, for $x \geq e$ it holds $W_0(x) \geq \log(x) - \log(\log(x))$. Therefore, for $t$ sufficiently large, it holds

\[
\log\left( 1 + \frac{1}{W_0\left( \exp\left( t / \tau - C \right) \right)} \right)
\leq \frac{1}{W_0( \exp( t / \tau - C ) )}
\leq \frac{1}{t / \tau - C - \log(t / \tau - C)} = \mathcal{O}(1/t)
\]
}

\subsubsection{Discussion of assumptions for the logistic bound}\label{appendix:logistic-assumption-discussion}

\RevDiff{
\paragraph{Separation assumption.}
The existence of an $\varepsilon$-separating ray for some $\varepsilon > 0$ in Proposition~\ref{prop:logistic-convergence} is identical to the separation assumption \citet[Assumption 4]{soudry18} (multi-class version, which itself recovers \citet[Assumption 1]{soudry18} in the two-class case, which is the standard notion of ``linear separability''). Then $\inf_\lambda \mathcal{L}(\lambda \zeta) = 0$ is consistency of the ray $\zeta$ with the class labels.

Indeed, the linear separability assumption is that for a dataset $(x_i, y_i) \in \mathbb{R}^d \times [c]$ for $i \in [n]$, there exists a vector $w \in \mathbb{R}^{c \times d}$ such that for all $i \in [n]$, and for all $k \in [c]$, if $k \neq y_i$, then $w_k \cdot x_i - w_{y_i} \cdot x_i < 0$.
Let $\varepsilon = \inf_{i} \inf_{k \neq y_i} - (w_k \cdot x_i - w_{y_i} \cdot x_i)$. 
Since the number of training points is finite and the number of classes is finite, this infimum is a minimum, and thus $\varepsilon > 0$. It follows immediately that $w$ is an $(\varepsilon / \lVert w \rVert_2)$-separating ray, and satisfies $\inf_\lambda \mathcal{L}(\lambda w)= 0$.

The difference is only that our assumption is quantified, because $\varepsilon$ appears explicitly in our bound, whereas it was previously abstracted away by the Landau asymptotic notation.
To properly quantify this notion of separation margin, one must be careful with the fact that the unquantified separation assumption is invariant by positive rescaling of the separating vector. We have chosen to define the ray $\zeta$ only up to a positive constant, whereas in \citep{soudry18}, a cancellation of the norm of the separating vector is chosen instead (convergence to $w^* / \lVert w^* \rVert$), but the two viewpoints are equivalent.

\paragraph{Isolation assumption.} Previous works operating in the finite-data regime did not explicitly have a mention of an isolation assumption. Indeed, for a finitely supported distribution $p$, one can simply take $\kappa = \min_i p_i$, as noted in Section~\ref{sec:logistic-regression}. For a dataset of size $n$ with equally-weighted samples, this reduces to $\kappa = 1/n$ and can again be abstracted away in asymptotic notation.
Since we have chosen to give explicit bounds, we must make that constant appear, hence the existence of the assumption.

We could have used $1/n$ in place of the introduction of the notion of isolation, but this would have forced a vacuous bound in the infinite-data regime, whereas a positive isolation constant guarantees convergence even with continuous distributions. We try hereafter to give a better intuition of why such a positive isolation might be proven in typical machine learning scenarii.

The use of $\kappa$ in the proof is $\mathbb{E}[ \psi(Y) ] \geq \kappa \, \psi(\mathbb{E}[Y])$ when $Y$ is $\kappa$-isolated and $\psi$ increasing. This is because we measure only the average loss, the pointwise loss averaged over points in the dataset, which could be driven by the loss on a single point. This happens precisely when there remains exactly one misclassified point $i_0$, while other points are correctly classified, i.e.\ $\ell = \frac{1}{n} \sum_i \ell_i \approx \frac{1}{n} \ell_{i_0}$.
This local misclassification is possible because there is 1 point which is sufficiently ``isolated'', hence the $1/n$, however if the dataset came with point-pairs very close to each other and identical labels, then it would become essentially impossible for a sufficently regular classifier to misclassify exactly one point, leading to a factor of $2/n$ instead (the corresponding amount of mass ``isolated''). \newline
For a fixed number of training points $n \in \mathbb{N}$, there always exists a dataset with a single isolated point, thus the bound $\kappa \geq 1/n$ is tight without assumptions on the data generation process.
However, there is typically an assumption in machine learning that we have not leveraged here: as the size of the dataset increases, the distribution of the data does not change, for instance all samples are taken independently identically distributed with a fixed distribution.
Thus, $\kappa$ need not vanish as $n \to +\infty$. The regularity of the underlying distribution and the regularity of the classifier (obtained by finiteness of $\lVert \theta \rVert_2$) could be analyzed together to derive a positive limit for $\kappa$.
Should a proof for such a property become available in the future, it could be chained with Proposition~\ref{prop:logistic-convergence} as-is directly to obtain a better convergence speed.
The use of $\kappa$ rather than $1/n$ in our bound is meant to highlight this possibility explicitly.
We otherwise use $\kappa = 1/n$ in experiments.
}

\subsubsection{Experiments for logistic regression}\label{appendix:logistic-experiments}

The following figures show examples of the convergence speed observed with gradient descent and step size 0.1 in different scenarios.
In Fig.~\ref{fig:logistic-E1} we depict a configuration where the bound we presented in Proposition~\ref{prop:logistic-convergence} accurately describes the observed evolution of the loss, including the flat startup, the sudden drop and its position, and the asymptotic regime $\mathcal{L}(\theta_t) \leq \frac{1}{\alpha + \beta t}$.
In Fig.~\ref{fig:logistic-E2}, we depict a more realistic configuration, where the general behavior observed is similar, but the bound's constants are off by several orders of magnitude.
In both cases, we take as isolation constant $\kappa = 1 / n$.

\begin{figure}[H]
\centering
\includegraphics[width=\textwidth]{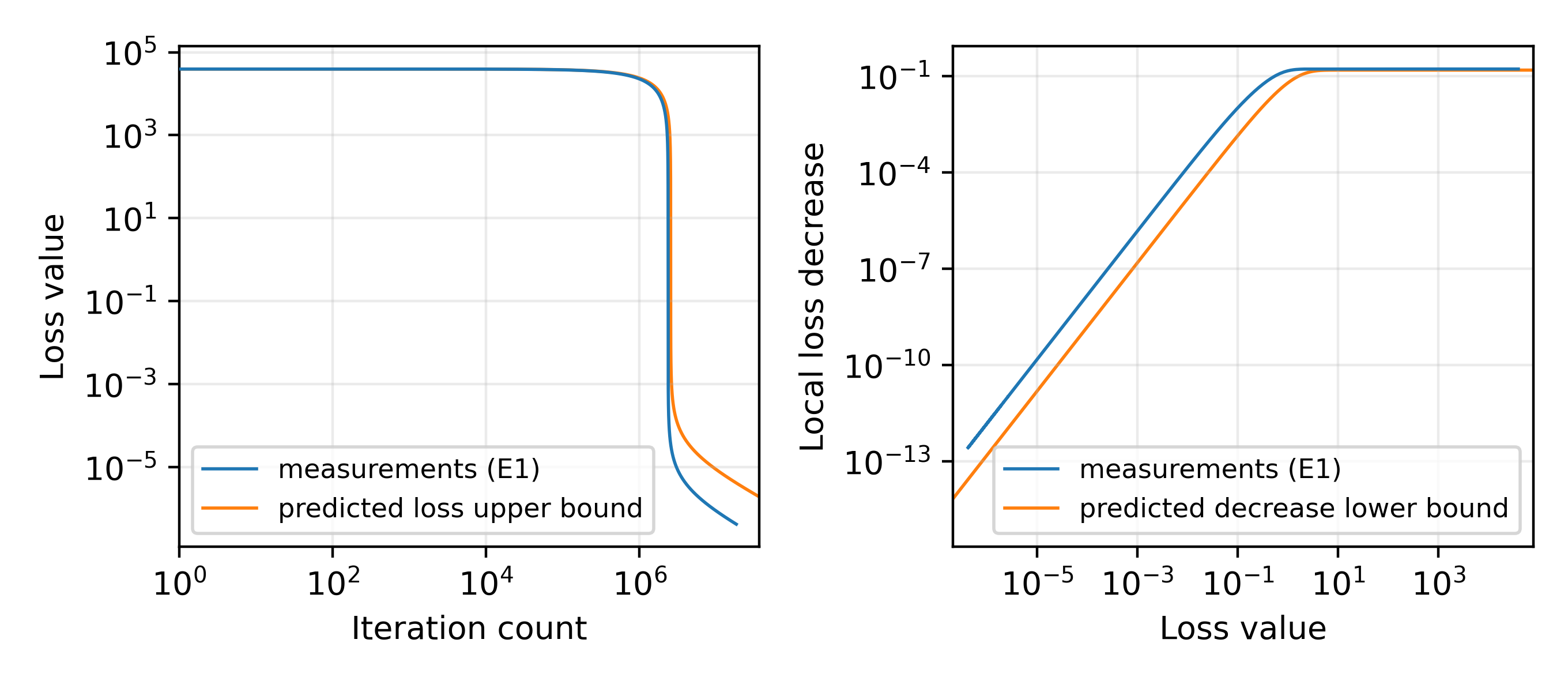}
\caption{Logistic regression on $n=3$ samples in dimension $d=4$ with $c=3$ classes.
The data is hand-picked to show a tight regime of the bound. Measurements and predicted curves overlap at first.}%
\label{fig:logistic-E1}
\end{figure}

\begin{figure}[H]
\centering
\includegraphics[width=\textwidth]{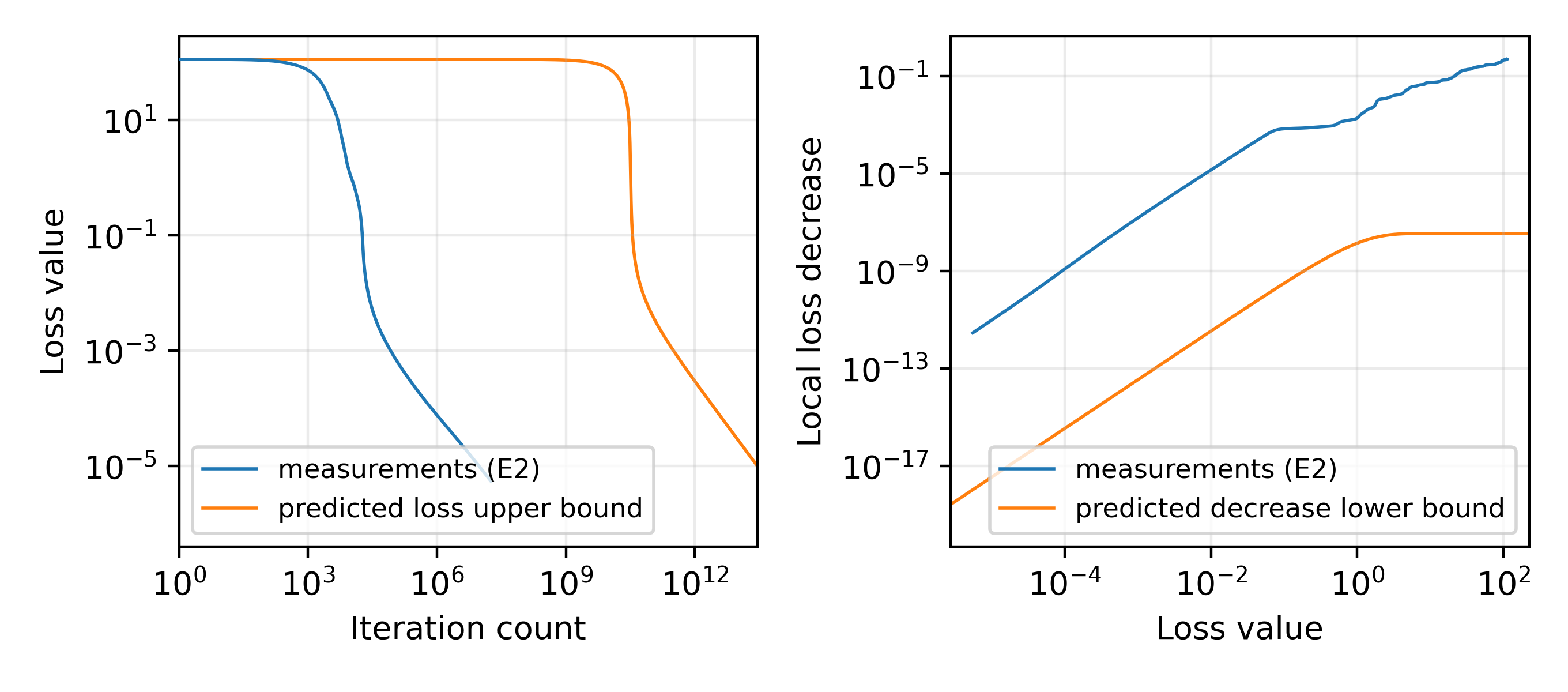}
\caption{Logistic regression on $n=100$ samples in dimension $d=5$ with $c=4$ classes.
The data points, optimal direction and initial point are drawn at random from gaussian distributions. }%
\label{fig:logistic-E2}
\end{figure}

These two experiments were conducted in parallel on an Intel i7 CPU, for a total running time of 14h.

\subsection{Periodic signal recovery, paired subcase}\label{appendix:sine-paired-proof}

For shortness in the following proof, for any $\omega \in \mathbb{R}$, let $e_\omega \in \mathcal{F}$ be the function $e_\omega : x \mapsto \sin(\omega x)$, and $e'_\omega = \frac{\partial}{\partial \omega} e_\omega \in \mathcal{F}$ its derivative, $e'_\omega : x \mapsto x \cos(\omega x)$.
Moreover, in all the following, we let $\psi = - \sinc'$ and $\phi = - \sinc''$ to avoid writing apostrophes and additional negative signs everywhere. Recall that we let $x_0$ be the first zero of $x \mapsto \phi(x)$, that is to say $x_0 \approx 2.0815$.

\begin{proof}[Proof of Proposition~\ref{prop:sine-paired}]
\looseness=-1
Let $\varepsilon = \eta / R$ and $\delta = \mu / R$.
Note that it holds $R \varepsilon \leq x_0$ and $\varepsilon < \frac{1}{2} \mu$.
Let $\theta = (a, \omega) \in \Theta$ such that $(\omega, \omega^*)$ is $\frac{\eta}{R}$-paired.
Let $h = F(a, \omega) - f^*$.
By the assumption $\ell(F(a,\omega)) \neq 0$, we know $\lVert h \rVert_\mathcal{D} \neq 0$.
We will show that $\mathrm{R}\left(\overbar{K_\theta}; h,h \right)$ is bounded below by some constant.
We defer the proof that this constant is positive (non-degeneracy of the bound) to a later section.

Let $g_{0,k} = e_{\omega_k}$ and $g_{1,k} = e_{\omega_k^*} - e_{\omega_k}$.
Observe that $h \in \mathrm{Span}(g)$ because
$$ h = \sum_k a_k e_{\omega_k} - a^*_k e_{\omega^*_k} = \sum_k (a_k - a^*_k) e_{\omega_k} + a^*_k (e_{\omega_k} - e_{\omega^*_k}) $$

Let $b_{0,k} = (\delta_{i=k}, 0)_{i \in [m]} \in \Theta$, and $b_{1,k} = \left(0, \delta_{i=k} \frac{1}{a_k} (\omega_k - \omega_k^*)\right)_{i \in [m]} \in \Theta$, so that it holds
$\d F_\theta \cdot b_{0,k} = e_{\omega_k}$ and $\d F_\theta \cdot b_{1,k} = (\omega_k - \omega_k^*) \, e'_{\omega_k}$.

$$\begin{aligned}
\mathrm{R}\left( \overbar{K_\theta}; h, h \right)
&= \sup_{\nu \in \Theta} \mathrm{R}\left(\overbar{\d F_\theta}; \nu, h \right)^2
&(1)
\\ &\geq \frac{\lambda_{\min}(\mathrm{R}\left(\inner{\mathcal{D}}; \d F_\theta \cdot b_i, g_j \right)_{i,j})^2 \min_{(i,u) \in [m] \times [2]} \lVert \d F_\theta \cdot b_{i,u} \rVert_\mathcal{D}^2 / \lVert b_{i,u} \rVert_\Theta^2 }{ \lambda_{\max}\left(\mathrm{R}\left( \inner{\Theta}; b_i, b_j\right)_{i,j} \right) \lambda_{\max}\left( \mathrm{R}\left( \inner{\mathcal{D}}; g_i, g_j \right)_{i,j} \right) }
&(2)
\end{aligned}$$
where $(1)$ is Proposition~\ref{prop:variational}, and $(2)$ is Proposition~\ref{prop:shattering}.
In the above expression, the indices $(k,0) \in ([m] \times \{0\})$ such that $a_k = a^*_k$ and the indices $(k,1) \in ([m] \times \{1\})$ such that $\omega_k = \omega_k^*$ have been omitted (since the corresponding derivative is zero), and thus the matrices are all well-defined.
For simplicity in the notation, and since the correction would just amount to selecting the corresponding subsets of $b$ and $g$ without altering the final result, we will just assume that $\forall k, a_k \neq a_k^*$ and $\forall k, \omega_k \neq \omega_k^*$ in the following, so the index set remains $([m] \times [2])$.

The first factor in the denominator is the largest eigenvalue of the identity, i.e.\ one.
The second factor of the numerator, corresponding to the singular value, is
$ \min_{(i,u)} \lVert \d F_\theta \cdot b_{i,u} \rVert_\mathcal{D}^2 / \lVert b_{i,u} \rVert_\Theta = \min_k (\lVert e_{\omega_k} \rVert_\mathcal{D}^2, a_k^2 \lVert e'_{\omega_k} \rVert_\mathcal{D}^2)  \geq \min(1, a_k^2) \left( \phi(R \varepsilon) - {1} / {(R (\delta - \varepsilon))} \right) \geq \alpha (\phi(\eta) - 1/(\mu - \eta)) $ (see Lemma~\ref{appendix-prop:sine-integration} for the lower bounds on the seminorms).

There remains only two matrices whose eigenvalues we need to bound.

We will proceed using Gershgorin's discs theorem \citep{gerschgorin1931} for the control of eigenvalues:
$$ \forall X \in \mathbb{R}^{k \times k}, \, \lambda_{\min}(X) \geq \inf_{i \in [k]} X_{i,i} - \frac{1}{2} \sum_{j \neq i} | X_{i,j} | + | X_{j,i} | $$
$$ \forall X \in \mathbb{R}^{k \times k}, \, \lambda_{\max}(X) \leq \sup_{i \in [k]} X_{i,i} + \frac{1}{2} \sum_{j \neq i} | X_{i,j} | + | X_{j,i} | $$

Starting with the denominator $z_0 = \lambda_{\max}(\mathrm{R}(\inner{\mathcal{D}}; g_i, g_j)_{i,j})$,
$$\begin{aligned}
z_0
&= \lambda_{\max}(\mathrm{R}(\inner{\mathcal{D}}; g_i, g_j)
\leq \sup_{(i,u) \in [m] \times [2]} 1 + \sum_{(j,v) \neq (i,u)} | \mathrm{R}(\inner{\mathcal{D}}; g_i, g_j) |
&(1)
\\ & \leq \sup_{i \in [m]} 1 + \frac{\psi(R \varepsilon) + \frac{1}{R (\delta - \varepsilon)}}{ \phi(R \varepsilon) - \frac{1}{R (\delta - \varepsilon)}} + 4 \sum_{j \neq i} \frac{ \frac{1}{R (|j-i| \delta - 2 \varepsilon)} + \frac{1}{R ((i+j+2) \delta - 2 \varepsilon)} }{ \phi(R \varepsilon) - \frac{1}{R (\delta - \varepsilon)} }
&(2)
\\ & \leq \sup_{i \in [m]} 1 + \frac{1}{ \phi(R \varepsilon) - \frac{1}{R (\delta - \varepsilon)}}\left( \psi(R \varepsilon) + \frac{1}{R (\delta - \varepsilon)} + \frac{4}{R (\delta - 2 \varepsilon)} \sum_{j \neq i} \left( \frac{1}{|j-i|} + \frac{1}{i+j+2} \right) \right)
&(3)
\\ &\leq 1 + \frac{1}{ \phi(R \varepsilon) - \frac{1}{R (\delta - \varepsilon)}}\left( \psi(R \varepsilon) + \frac{1}{R (\delta - \varepsilon)} + \frac{4}{R (\delta - 2 \varepsilon)} 4 H_m  \right)
= 1 + \rho_0
&(4)
\end{aligned} $$
where $(1)$ is Gershgorin's disc upper bound with a symmetric matrix, $(2)$ is Proposition~\ref{appendix-prop:off-diagonal} proven below, $(3)$ is factorization of the terms depending on $(i,j)$ in the denominators (because if $j \neq i$, then $|j-i| \geq 1$), and $(4)$ is
$\sum_{j \neq i} \frac{1}{|j-i|} + \frac{1}{j+i+2} \leq \sum_{j < i} \frac{2}{|j-i|} + \sum_{j > i} \frac{2}{|j-i|} \leq 2 \sum_{k=1}^{m} \frac{2}{k} = 4 H_m$.

For the numerator $z_1 = \lambda_{\min}(\mathrm{R}(\overbar{\d F_\theta}; b_i, g_j)$ now, we start by simplifying each entry
$$ \mathrm{R}(\d F_\theta; b_i, g_j)
= \frac{ \langle \d F_\theta \cdot b_i, g_j \rangle_\mathcal{D} }{ \lVert b_i \rVert_\Theta \lVert g_j \rVert_{\mathcal{D}} }
= \frac{ \langle \d F_\theta \cdot b_i, g_j \rangle_\mathcal{D} }{ \lVert b_i \rVert_\Theta \lVert g_j \rVert_{\mathcal{D}} }
= \mathrm{R}(\inner{\mathcal{D}}; h_i, g_j)
$$
Then apply Gershgorin's disc lower bound
$$\begin{aligned}
z_1
&= \lambda_{\min}(\mathrm{R}(\overbar{\d F_\theta}; b_i, g_j)_{i,j})
= \lambda_{\min}(\mathrm{R}(\inner{\mathcal{D}}; h_i, g_j)_{i,j})
\\ &\geq \inf_{(i,u) \in [m] \times [2]} \mathrm{R}(\inner{\mathcal{D}}; h_{i,u}, g_{i,u}) - \frac{1}{2} \sum_{(j,v) \neq (i,u)} | \mathrm{R}(\inner{\mathcal{D}}; h_{i,u}, g_{i,v}) | + | \mathrm{R}(\inner{\mathcal{D}}; h_{j,v}, g_{i,u}) |
\\&\geq \inf_{i\in [m]}  \frac{\phi(R \varepsilon) - \frac{1}{R (\delta - \varepsilon)}}{ \phi(0) + \frac{1}{R (\delta - \varepsilon)}} - \left(\frac{\psi(R \varepsilon) + \frac{1}{R (\delta - \varepsilon)}}{\phi(R \varepsilon) - \frac{1}{R (\delta- \varepsilon)}} + 4 \sum_{j \neq i} \frac{ \frac{1}{R (|j-i| \delta - 2\varepsilon)} + \frac{1}{R ((i+j+2) \delta - 2 \varepsilon)} }{ \phi(R \varepsilon) - \frac{1}{R (\delta - \varepsilon)} } \right)
&(1)
\\ &\geq \inf_{i \in [m]}  \frac{\phi(R \varepsilon) - \frac{1}{R (\delta - \varepsilon)}}{ \phi(0) + \frac{1}{R (\delta - \varepsilon)}} - \rho_0
= \kappa_0 - \rho_0
&(2)
\end{aligned}$$
where $(1)$ is Proposition~\ref{appendix-prop:on-diagonal} for the left term and Proposition~\ref{appendix-prop:off-diagonal} for the right term (with a symmetric upper-bound),
and $(2)$ is the same upper-bound for off-diagonal terms as calculated above.
\end{proof}

\begin{proof}[Proof of non-degeneracy for Proposition~\ref{prop:sine-paired}]
Recall the definition of the constants
\[ \kappa_0(\mu, \eta) = \frac{\phi(\eta) - \frac{1}{\mu - \eta}}{\phi(0) + \frac{1}{\mu - \eta}}
\quad\quad
\rho_0(\mu, \eta) = \frac{\psi(\eta) + \frac{1}{\mu - \eta} + \frac{4 H}{\mu - 2 \eta}}{\phi(\eta) - \frac{1}{\mu - \eta}} \]
By continuity, to show $\exists \eta > 0$ s.t. $\kappa_0(\mu,\eta) > \rho_0(\mu,\eta)$, it is sufficient to show $\kappa_0(\mu,0) > \rho_0(\mu, 0)$. These values are
$\kappa_0(\mu, 0) = \left(\phi(0) - \frac{1}{\mu}\right) / \left( \phi(0) + \frac{1}{\mu}\right)$
and $\rho_0(\mu, 0) = \left(\frac{1}{\mu} + \frac{4H}{\mu}\right) / \left(\phi(0) - \frac{1}{\mu}\right)$.

Using $\phi(0) = 1/3$ and reorganizing terms, the equation $\kappa_0(\mu, 0) > \rho_0(\mu, 0)$ is satisfied if and only if
\[ \left( \frac{\mu}{3} - 1 \right)^2 - \left( 1 + 4H \right) \left( \frac{\mu}{3} + 1 \right) > 0 \]

This is a polynomial in $\mu$ of degree two, and positive at infinity, therefore letting $\mu_0$ be its largest root,
it holds for all $\mu > \mu_0$ that $\kappa_0(\mu, 0) > \rho_0(\mu,0)$.
\end{proof}

\begin{proposition}[Auxiliary for on-diagonal control]\label{appendix-prop:on-diagonal}
If $(\omega, \omega^*)$ is $\varepsilon$-paired and $\omega^*$ is $\delta$-separated,
where it holds $R \varepsilon \leq x_0$, and $\varepsilon < \frac{1}{2} \delta$, and $\forall i, \omega_i \neq \omega_i^*$, then
$$ \forall (i,u) \in [m] \times [2], \> \mathrm{R}(\inner{\mathcal{D}}; g_{i,u}, h_{i,u}) \geq \frac{\phi(R \varepsilon) - \frac{1}{R (\delta - \varepsilon)} }{ \phi(0) + \frac{1}{R (\delta - \varepsilon)} } $$
where $g,h \in \mathcal{F}^{m \times 2}$ satisfy $g_{i,0} = h_{i,0} = e_{\omega_i}$ and $g_{i,1}, h_{i,1} \in \{ e_{\omega_k} - e_{\omega_k^*}, (\omega_k - \omega_k^*) e'_{\omega_k} \}$.
\end{proposition}

\begin{proposition}[Auxiliary for off-diagonal control]\label{appendix-prop:off-diagonal}
If $(\omega, \omega^*)$ is $\varepsilon$-paired and $\omega^*$ is $\delta$-separated and ordered $(i \leq j \Rightarrow \omega^*_i \leq \omega^*_j)$,
where it holds $R \varepsilon \leq x_0$, and $\varepsilon < \frac{1}{2} \delta$,
and $\forall i, \omega_i \neq \omega_i^*$,
then
$$ \forall i \in [m], \> \sup_{\substack{(u,v) \in [2] \times [2] \\ u \neq v}}  \left| \mathrm{R}\left( \inner{\mathcal{D}}; g_{i,u}, h_{i,v} \right) \right| \leq \frac{\psi(R \varepsilon) + \frac{1}{R (\delta - \varepsilon)}}{\phi(R \varepsilon) - \frac{1}{R (\delta - \varepsilon)}} $$
$$ \forall (i,j) \in [m] \times [m], \, i \neq j \Rightarrow \> \sup_{(u,v) \in [2] \times [2]} \left| \mathrm{R}\left( \inner{\mathcal{D}}; g_{i,u}, h_{j,v}\right) \right| \leq \frac{\frac{1}{R (|j -i | \delta - 2 \varepsilon)} + \frac{1}{R ((i+j+2) \delta - 2 \varepsilon)}}{\phi(R \varepsilon) - \frac{1}{R (\delta - \varepsilon)}} $$
where $g,h \in \mathcal{F}^{m \times 2}$ satisfy $g_{i,0} = h_{i,0} = e_{\omega_i}$ and $g_{i,1}, h_{i,1} \in \{ e_{\omega_k} - e_{\omega_k^*}, (\omega_k - \omega_k^*) e'_{\omega_k} \}$.
\end{proposition}

The proof for both propositions is a case disjunction. We will state an intermediate lemma first.

\begin{lemma}[Cosine bound by cross-ratio control]\label{appendix-prop:cross-ratio}
\,\newline
Let $a : [0,1] \to \mathcal{F}$ and $b : [0,1] \to \mathcal{F}$.
If there exists $\alpha \in [0,1]$ such that it holds
$$ \forall (s,t,u,v) \in [0,1]^4, \> \frac{\langle a(s), b(u) \rangle_\mathcal{D} \langle a(t), b(v) \rangle_\mathcal{D}}{ \langle a(s), a(t) \rangle_\mathcal{D} \langle b(u), b(v) \rangle_\mathcal{D}} \leq \alpha^2 $$
then $ \left| \mathrm{R}\left(\inner{\mathcal{D}}; \int_0^1 a(s) \d s, \int_0^1 b(t) \d t \right) \right| \leq \alpha$.
\end{lemma}

\begin{proof}
By expanding the definition $(1)$, bilinearity $(2)$, then applying the hypothesis pointwise $(3)$.
$$\begin{aligned}
\mathrm{R}\left(\inner{\mathcal{D}}; \int_0^1 a(s) \d s, \int_0^1 b(t) \d t\right)^2
& = \frac{
\left\langle \int_0^1 a(s) \d s, \int_0^1 b(t) \d t \right\rangle_\mathcal{D} \left\langle \int_0^1 a(u) \d u, \int_0^1 b(v) \d v \right\rangle_\mathcal{D}
}{
\left\langle \int_0^1 a(s) \d s, \int_0^1 a(u) \d u \right\rangle_\mathcal{D} \left\langle \int_0^1 b(t) \d t, \int_0^1 b(v) \d v \right\rangle_\mathcal{D}
}
&(1)
\\ &= \frac{
\int_0^1 \int_0^1 \int_0^1 \int_0^1 \langle a(s) , b(t) \rangle_\mathcal{D} \langle a(u) , b(v) \rangle_\mathcal{D} \d s \d t \d u \d v
}{
\int_0^1 \int_0^1 \int_0^1 \int_0^1 \langle a(s) , a(u) \rangle_\mathcal{D} \langle b(t) , b(v) \rangle_\mathcal{D} \d s \d t \d u \d v
}
&(2)
\\ &\leq \alpha^2
&(3)
\end{aligned}$$
An upper bound on the cross-ratio allows interversions under the integral, thus the result.
\end{proof}

\begin{proof}[Proof of Proposition~\ref{appendix-prop:off-diagonal}]
For shortness, let $I_i = \{ (1-t) \omega_i + t (\omega_i^*), t \in [0,1] \} \subseteq \mathbb{R}$.
By observing that $e_{\omega_k} - e_{\omega_k^*} = \int_0^1 e'_{q(t)} (\omega_k - \omega_k^*) \d t$ for $q(t) = (1-t)\omega_k^* + t \omega_k$ on one hand, and $(\omega_k - \omega_k^*) e'_{\omega_k} = \int_0^1 e'_{r(t)} (\omega_k - \omega_k^*) \d t$ for $r(t) = \omega_k$ on the other hand, we reduce to cross-ratio upper bounds only.

For the first part of the proof, let $i \in [m]$. By symmetry, it is sufficient to consider $(u = 0, v = 1)$.
$$ \forall p, q \in I_i, \frac{ \langle e'_p, e_{\omega_i} \rangle_\mathcal{D} \langle e'_q, e_{\omega_i} \rangle_\mathcal{D} }{ \langle e'_p, e'_q \rangle_\mathcal{D} \langle e_{\omega_i}, e_{\omega_i} \rangle_\mathcal{D} } \leq \frac{\left( \psi(R \varepsilon) + \frac{1}{R (\delta - \varepsilon)} \right)^2 }{\left( 1 - \frac{1}{R (\delta - \varepsilon)} \right) \left( \phi(R \varepsilon) - \frac{1}{R (\delta - \varepsilon)} \right)} \leq \left( \frac{\psi(R \varepsilon) - \frac{1}{R (\delta - \varepsilon)}}{\phi(R \varepsilon) - \frac{1}{R (\delta - \varepsilon)}} \right)^2 $$
by Lemma~\ref{appendix-prop:sine-integration}, where the last step is $\phi(R \varepsilon) \leq \phi(0) = 1/3 \leq 1$. Conclude by Lemma~\ref{appendix-prop:cross-ratio}.

For the second part of the proposition, let $(i,j) \in [m] \times [m]$, and proceed by case disjuction on $(u,v) \in [2] \times [2]$. Let $(p_i,q_i) \in I_i \times I_i$ and $(p_j,q_j) \in I_j \times I_j$, and observe that
$$ (u=0, v=0) \quad  \frac{ \langle e_{\omega_i}, e_{\omega_j} \rangle_\mathcal{D} \langle e_{\omega_i}, e_{\omega_j} \rangle_\mathcal{D}}{ \langle e_{\omega_i}, e_{\omega_i} \rangle_\mathcal{D} \langle e_{\omega_j}, e_{\omega_j} \rangle_\mathcal{D}} \leq \frac{\left(\frac{1}{R (|j-i| \delta - 2 \varepsilon)} + \frac{1}{R((i+j+2) \delta - 2 \varepsilon)} \right)^2}{\left( 1 - \frac{1}{R (\delta - \varepsilon)} \right)^2} $$
$$ (u=0, v=1) \quad \frac{ \langle e_{\omega_i}, e'_{p_j} \rangle_\mathcal{D} \langle e_{\omega_i}, e'_{q_j} \rangle_\mathcal{D} }{ \langle e_{\omega_i}, e_{\omega_i} \rangle_\mathcal{D} \langle e'_{p_j}, e'_{q_j} \rangle_\mathcal{D}}  \leq \frac{\left(\frac{1}{R (|j-i| \delta - 2 \varepsilon)} + \frac{1}{R((i+j+2) \delta - 2 \varepsilon)} \right)^2}{\left( 1 - \frac{1}{R (\delta - \varepsilon)} \right) \left( \phi(R \varepsilon) - \frac{1}{R (\delta - \varepsilon)} \right) } $$
$$ (u=1, v=1) \quad \frac{ \langle e'_{p_i}, e'_{p_j} \rangle_\mathcal{D} \langle e'_{q_i}, e'_{q_j} \rangle_\mathcal{D} }{ \langle e'_{p_i}, e'_{q_i} \rangle_\mathcal{D} \langle e'_{p_j}, e'_{q_j} \rangle_\mathcal{D} } \leq \frac{\left(\frac{1}{R (|j-i| \delta - 2 \varepsilon)} + \frac{1}{R((i+j+2) \delta - 2 \varepsilon)} \right)^2}{\left( \phi(R \varepsilon) - \frac{1}{R (\delta - \varepsilon)} \right)^2 } $$
by Lemma~\ref{appendix-prop:sine-integration}.
The case $(u=1,v=0)$ is identical to $(u=0,v=1)$ by symmetry, and conclusion follows as above by $\phi(R \varepsilon) \leq 1$ then Lemma~\ref{appendix-prop:cross-ratio}.
\end{proof}

\begin{proof}[Proof of Proposition~\ref{appendix-prop:on-diagonal}]
Let $i \in [m]$. We will prove $\mathrm{R}(\inner{\mathcal{D}}; g_{i,u}, h_{i,u}) \geq \kappa_0$ by case disjunction on $u \in [2]$.
For the case $u=0$, observe that $\mathrm{R}(\inner{\mathcal{D}}; e_{\omega_k}, e_{\omega_k}) = 1$.
Since $\phi$ is decreasing on $[0, R \varepsilon]$ (see \ref{appendix:sinc-bounds}), and $\kappa_0 \leq 1$, the conclusion is immediate.

For the case $u=1$, identically to the proof of Prop~\ref{appendix-prop:off-diagonal} above,
let $(p,q,r,s) \in I_i^4$.
$$ \frac{ \langle e'_p, e'_q \rangle_\mathcal{D} \langle e'_r, e'_s \rangle_\mathcal{D} }{ \langle e'_p, e'_r \rangle_\mathcal{D} \langle e'_q, e'_s \rangle_\mathcal{D} } \geq \left( \frac{\phi(R \varepsilon) - \frac{1}{R (\delta - \varepsilon)}}{ \phi(0) - \frac{1}{R (\delta - \varepsilon)} } \right)^2 = \kappa_0^2 $$
by Lemma~\ref{appendix-prop:sine-integration}. Thus by expanding integrals as for Lemma~\ref{appendix-prop:cross-ratio}, the cross-ratio lower bound implies
$$\mathrm{R}\left(\inner{\mathcal{D}}; \int g, \int h \right) = \frac{ \langle \int g, \int h \rangle_\mathcal{D} }{\sqrt{ \langle \int g, \int g \rangle_\mathcal{D} \langle \int h, \int h \rangle_\mathcal{D} } } \geq \kappa_0$$
\end{proof}

\begin{lemma}\label{appendix-prop:sine-integration}
If $(\omega, \omega^*)$ is $\varepsilon$-paired and $\omega^*$ is $\delta$-separated and ordered, $R \varepsilon \leq x_0$ and $\varepsilon < \frac{1}{2} \delta$, then
$$\begin{aligned}
&\forall i,&& \forall u \in I_i, &\langle e_u, e_u \rangle_\mathcal{D} &\geq \frac{1}{2} - \frac{1}{2 R (\delta - \varepsilon)}
\\ &\forall i,&& \forall (u,v) \in I_i \times I_i, &\frac{1}{R} \left| \langle e'_u, e_v \rangle_\mathcal{D} \right| &\leq \frac{\psi(R \varepsilon)}{2} + \frac{1}{2 R (\delta - \varepsilon)}
\\ &\forall i,&& \forall (u,v) \in I_i \times I_i, &\frac{1}{R^2} \langle e'_u, e'_v \rangle_\mathcal{D} &\in \left[ \frac{1}{2} \phi(R \varepsilon) - \frac{1}{2 R (\delta - \varepsilon)}, \frac{1}{2} \phi(0) + \frac{1}{2 R ( \delta - \varepsilon)} \right]
\end{aligned}$$

Additionally, if $i \neq j$, then for all $(u,v) \in I_i \times I_j$, it holds
$$
\max\left( \left|\langle e_u, e_v \rangle_\mathcal{D} \right|, \frac{1}{R} \left|\langle e'_u, e_v \rangle_\mathcal{D} \right|, \frac{1}{R^2} \left| \langle e'_u, e'_v \rangle_\mathcal{D} \right| \right) \leq \frac{1}{2 R (|i -j| \delta - 2 \varepsilon)} + \frac{1}{2 R ((i+j+2) \delta - 2 \varepsilon)}
$$
where $I_i = \{ (1-t) \, \omega_i + t \, \omega_i^*, t \in [0,1] \} \subseteq \mathbb{R}$.
\end{lemma}

\begin{proof}
The idea is to first compute dot products in closed forms to make the cardinal sine function $(\sinc : x \mapsto \sin(x) / x)$ appear, then rely on properties of the cardinal sine and its derivatives to prove each property by case disjunction.
Therefore, for any $(u,v) \in \mathbb{R}_+ \times \mathbb{R}_+$, compute the integral (assuming $u \neq v$ and completing by continuity) using $\left[ 2 \sin(a) \sin(b) = \cos(a-b) - \cos(a+b) \right]$,
$$\begin{aligned}
\langle e_u, e_v \rangle_\mathcal{D} &= \frac{1}{2 R} \int_{-R}^{+R} \sin(u x) \sin (v x) \d x
\\ &= \frac{1}{4 R} \int_{-R}^{+R} \cos((u-v) x) - \cos((u+v) x)
\\ &= \frac{1}{4 R} \left[ \frac{\sin((u-v) x)}{u-v} - \frac{\sin((u+v) x)}{u+v} \right]_{-R}^{+R}
\\ &= \frac{1}{2} \left( \sinc(Ru - Rv) - \sinc(Ru + Rv) \right)
\end{aligned}$$
Compute the others by derivation
$$\begin{aligned}
&\langle e'_u, e_v \rangle_\mathcal{D} =& \frac{\partial}{\partial u} \langle e_u, e_v \rangle_\mathcal{D} &= \frac{R}{2} \left( \sinc'(R u - R v) - \sinc'(R u + R v) \right)
\\ &\langle e'_u, e'_v \rangle_\mathcal{D} =& \frac{\partial}{\partial u} \frac{\partial}{\partial v} \langle e_u, e_v \rangle_\mathcal{D} &= \frac{R^2}{2} \left( - \sinc''(R u - R v) - \sinc''(R u + R v) \right)
\end{aligned}$$

The proof of all statements will then follow from a couple of properties of $\sinc$ and its derivatives:
\begin{enumerate}
    \item $\forall x \in \mathbb{R}, \max \left\{ \left| \sinc(x) \right|, \left| \sinc'(x) \right|, \left| \sinc''(x) \right| \right\} \leq \frac{2}{|x|}$
    \item $(- \sinc'')$ is non-negative decreasing on $[0, x_0]$, where $x_0 \approx 2.0815$ is its first zero.
    \item $(- \sinc')$ is non-negative increasing on $[0, x_0]$.
\end{enumerate}
These properties are depicted in Figure~\ref{fig:sinc-derivatives}, and proven in Appendix~\ref{appendix:sinc-bounds}.

\begin{figure}[H]
\centering
\includegraphics[width=\textwidth]{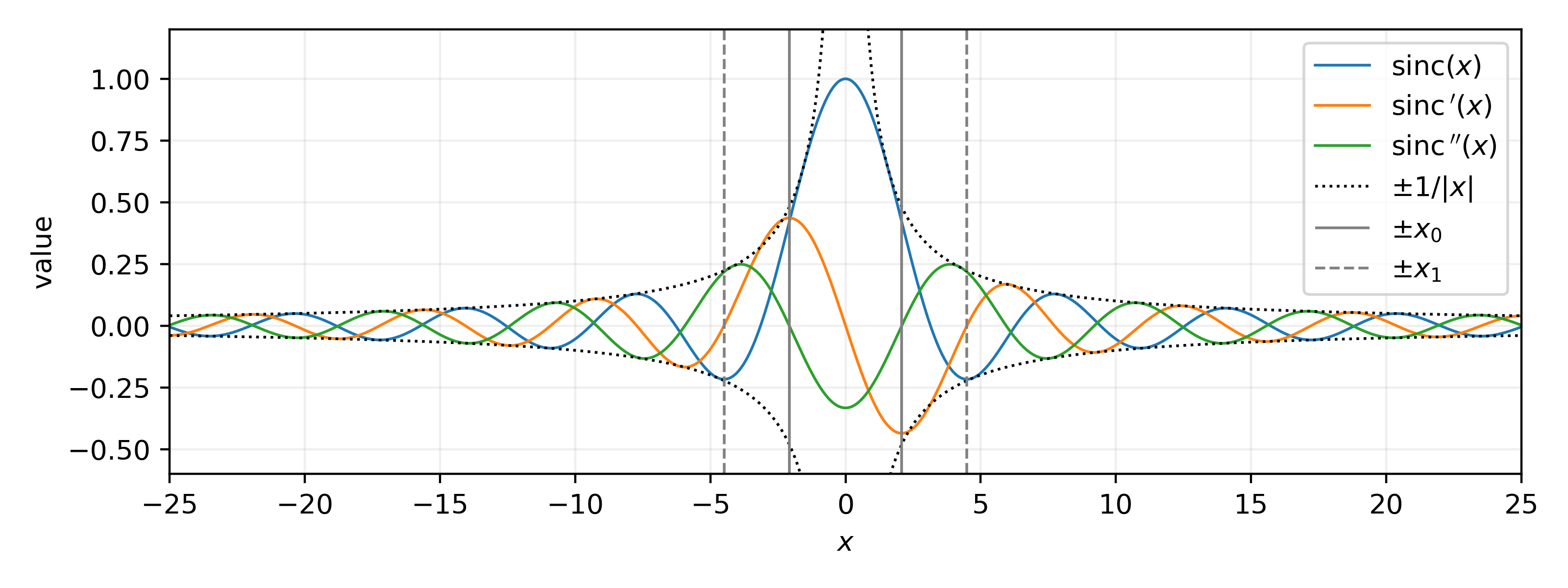}
\caption{$\mathrm{sinc}$ and derivatives, first zeros $\sinc'(x_1) = 0$, $\sinc''(x_0) = 0$, $\frac{1}{|x|}$ envelope on $\pm [x_1, +\infty[$}%
\label{fig:sinc-derivatives}
\end{figure}

For the first property, let $i \in [m]$, let $u \in I_i$, and observe that
$$ \langle e_u, e_u \rangle_\mathcal{D} = \frac{1}{2} \left( \sinc(0) - \sinc(2 R u) \right) \geq \frac{1}{2} \left(1 - \frac{2}{2 R |u|} \right) \geq \frac{1}{2} - \frac{1}{2 R (\delta - \varepsilon)} $$
since $|u| \geq |\omega_i^*| - |\omega_i - \omega_i^*| \geq \delta - \varepsilon$.

For the second property, let $i \in [m]$, and $(u,v) \in I_i \times I_i$. Without loss of generality, assume $u \leq v$,
$$\begin{aligned}
\frac{1}{R} \left| \langle e'_u, e'_v \rangle_\mathcal{D} \right| &= \frac{1}{R} \left| \frac{R}{2} \left( \sinc'(R u - R v) - \sinc'(R u + R v) \right) \right|
\\ &\leq \frac{1}{2} \left( \left| \sinc'(R u - R v) \right| + \left| \sinc'(R u + R v) \right| \right)
\\ &\leq \frac{1}{2} \left( \psi(R \varepsilon) + \frac{2}{R | u + v |} \right)
\\ &\leq \frac{\psi(R \varepsilon)}{2} + \frac{1}{2 R (\delta - \varepsilon)}
\end{aligned}$$
since $| \sinc'(R u - R v) | = - \sinc'(R u - R v) \leq - \sinc'(R \varepsilon)$ by increase of $\psi = - \sinc'$ and $|u - v| \leq \varepsilon$ for the first term, and $\min(|u|,|v|) \geq \delta - \varepsilon$ for the second term.

For the third property, let $i \in [m]$ and $(u,v) \in I_i \times I_i$, without loss of generality $u \leq v$.
$$\begin{aligned}
\frac{1}{R^2} \langle e'_u, e'_v \rangle_\mathcal{D}
&= \frac{1}{R^2} \frac{R^2}{2} \left( - \sinc''(R u - R v) - \sinc''(R u + R v) \right)
\\ &\in \left[ \frac{1}{2}\phi(R \varepsilon) - \frac{1}{2 R (\delta - \varepsilon)}, \frac{1}{2} \phi(0) + \frac{1}{2 R (\delta - \varepsilon)} \right]
\end{aligned}$$
since $\phi(R u - R v) \in \left[ \phi(R \varepsilon), \phi(0) \right]$ by decrease of $\phi$ and since $| u - v| \leq \varepsilon$ for the first term, and $\phi(R u + R v) \in \left[- 1 / {(R (\delta - \varepsilon))}, + 1 / {(R (\delta - \varepsilon))} \right]$ since $\min(|u|, |v|) \geq \delta - \varepsilon$ for the second term.

Finally, for the last property, let $(i,j) \in [m] \times [m]$ such that $i \neq j$, and $(u,v) \in I_i \times I_j$.
$$
\max\left( \left|\langle e_u, e_v \rangle_\mathcal{D} \right|, \frac{1}{R} \left|\langle e'_u, e_v \rangle_\mathcal{D}\right|, \frac{1}{R^2} \left|\langle e'_u, e'_v \rangle_\mathcal{D}\right| \right) \leq \frac{1}{2} \left( \frac{1}{R (|i -j| \delta - 2 \varepsilon)} + \frac{1}{R ((i+j+2) \delta - 2 \varepsilon)} \right)
$$
Because it holds $|u - v| \geq |j-i| \delta - 2 \varepsilon$ and $|u| + |v| \geq (i+1) \delta  - \varepsilon + (j+1) \delta - \varepsilon$.
\end{proof}

\subsubsection{Summary of the periodic signal recovery convergence argument}

The proof is a little involved and the computations hard to follow, but the interesting part is that the proof is broken down, by relatively easy steps, into smaller statements that can be checked independently of each other.
First, by Prop~\ref{prop:KLoja-comp}, convergence proofs on the quadratic loss can be reduced to control of a Rayleigh quotient away from zero. Secondly, by Prop~\ref{prop:shattering}, Rayleigh quotient control is reduced to some easy singular values computation and an eigenvalue control of a matrix of simpler Rayleigh quotients in a well-chosen basis. Thirdly, by Gershgorin's disc theorem, the eigenvalue control is reduced  to a number of upper bounds on cosine similarities. Then, by Lemma~\ref{appendix-prop:cross-ratio}, the numerous upper bounds on cosine similarities are reduced to upper bounds on cross-ratios to reduce the number of distinct cases to consider. Finally, by Lemma~\ref{appendix-prop:sine-integration}, each cross-ratio bound is reduced to the analysis of a real-valued function on a small interval.

\subsubsection{Properties of the cardinal sine and derivatives}\label{appendix:sinc-bounds}

By definition, $\sinc(x) = \sin(x) / x$, thus $\left| \sinc(x) \right| \leq 1 / |x| \leq 2 / |x|$.
Moreover, we will show $\sinc(x) \leq 1$. By symmetry (since $\sinc(-x) = \sinc(x)$), it is sufficient to show that $\sin(x) \leq x$ for all $x \in \mathbb{R}_+$, which holds because $\rho : x \in \mathbb{R}_+ \mapsto x - \sin(x)$ satisfies $\rho(0) = 0$ and is decreasing because $\rho'(x) = 1 - \cos(x) \leq 0$.

By derivation of the quotient,
$\sinc' : x \mapsto \frac{\cos(x)}{x} - \frac{\sin(x)}{x^2}$,
thus by triangular inequality and the above,
$$ \left| \sinc'(x) \right| = \left| \frac{\cos(x)}{x} - \frac{\sinc(x)}{x} \right| \leq \frac{|\cos(x)|}{|x|} + \frac{|\sinc(x)|}{|x|} \leq \frac{2}{|x|} $$
Now let us show that $|\sinc'(x)| \leq \frac{1}{2}$.
As previously, it is sufficient by antisymmetry (since $\sinc'(-x) = - \sinc'(x)$) to prove the result on $ x \in \mathbb{R}_+$. We proceed by studying the function $\rho : x \in \mathbb{R}_+ \mapsto \frac{1}{2} x^2 - x \cos(x) + \sin(x) $, null at zero, whose derivative is $\rho'(x) = x + x \sin(x) - \cos(x) + \cos(x) = x (1 + \sin(x)) \geq 0$.
Hence $\rho(x) \geq \rho(0) = 0$ and thus $\frac{1}{2}x^2 \geq x \cos(x) - \sin(x)$, therefore $\frac{1}{2} \geq \sinc'(x)$ for $x \in \mathbb{R}_+$.
Similarly for the other inequality,
study $\tau : x \in\mathbb{R}_+ \mapsto - \frac{1}{2} x^2 - x \cos(x) + \sin(x)$, whose derivative is $\tau'(x) = -x (1 - \sin(x)) \leq 0$, hence $\tau(x) \leq \tau(0) = 0$, thus $- \frac{1}{2} x^2 \leq  x\cos(x) - \sin(x)$, therefore $\sinc'(x) \geq - \frac{1}{2}$ for $x \in\mathbb{R}_+$.
This concludes the proof that $x \geq 0 \Rightarrow |\sinc'(x)| \leq \frac{1}{2}$.

Computing the derivative again,
$$ \sinc'' : x \mapsto \frac{2 \sin(x)}{x^3} - \frac{2 \cos(x)}{x^2} - \frac{\sin(x)}{x} $$

Using the recently proven fact $| \sinc'(x) | \leq 1/2$,
$$ \left| \sinc''(x) \right| = \left| - 2 \frac{\sinc'(x)}{x} - \frac{\sin(x)}{x} \right|
\leq 2 \left| \frac{ \sinc'(x) }{x} \right| + \left| \frac{\sin(x)}{x} \right| \leq \frac{2}{|x|} $$

It remains to show that $\psi$ is increasing on $[0, x_0]$ and $\phi$ is decreasing on the same interval.
Since $\phi$ is continuous, $\phi(0) = \frac{1}{3}$ and $x_0$ is the first zero of $\phi$ by definition, it follows that $\psi'(x) = \phi(x) \geq 0$ for $x \in [0,x_0]$, which proves the first statement.

It remains to show that $\phi = - \sinc''$ is decreasing on $[0, x_0]$,
for which it is sufficient to show that $\sinc'''$ is positive on this interval.
Using the form $\sinc''(x) = - 2 \sinc'(x) / x - \sinc(x)$,
\[\begin{aligned}
\sinc'''(x) &= -2 \frac{\sinc''(x)}{x} + 2 \frac{\sinc'(x)}{x^2} - \sinc'(x)
\\ &= -2 \frac{1}{x} \left(- 2 \frac{\sinc'(x)}{x} - \sinc(x) \right) + 2 \frac{\sinc'(x)}{x^2} - \sinc'(x)
\\ &= \left( \frac{6}{x^2} - 1 \right) \sinc'(x) + 2 \frac{\sinc(x)}{x}
\end{aligned}\]
on the interval $[0,x_0]$, $\sinc(x) \geq 0$, $\sinc'(x) \geq 0$ and $(6 / x^2 - 1) \geq 0$.

\pagebreak
\subsection{Kurdyka-\Loja{} region for two-layer networks}\label{appendix:relu-proof}

\RevDiff{We split the proof of Proposition~\ref{prop:relu-loja} in two parts, first the inequality satisfied in high probability (a), and then how to leverage this inequality to get a convergence speed (b).}

\begin{proof}[Proof of Proposition~\ref{prop:relu-loja} \RevDiff{(a)}]
For the first part of the proof (the inequality), note that by density of $\mathrm{Im}(F)$ in $L^1(K)$ \citep[Proposition~1]{leshno1993multilayer}, there exists $m^* \in \mathbb{N} \setminus\{0\}$ and $\theta^* \in \Theta^{(m^*)}$ such that $\sup_{x \in K} \lVert F(\theta^*)(x) - f^*(x) \rVert \leq \sqrt{\varepsilon} / 2$.
We will write $\lVert g \rVert_\infty = \sup_{x \in K} \left| g(x) \right|$ for shortness.

Let $(w^*,a^*) = \theta^*$.
We will show that for any $(w,a)$ such that there is at least one $w_i$ in a bassin around $w^*_j$ for all $j \in [m^*]$, \RevDiff{$F$ has a first-order approximation that is an $\varepsilon$-approximation of $f^*$} (\RevDiff{i.e.\ }it is sufficient to roughly approximate features to get relatively good gradients far from the optimum).

Formally, let $\eta = \sqrt{\varepsilon} / (2 \lVert a^* \rVert_{\RevDiff{1}} L_\sigma D)$, where $L_\sigma$ is the Lipschitz constant of $\sigma$ and $D = \sup_{x \in K} \lVert x \rVert_2$.
Let $\mathcal{P}_{\theta^*}^{(m)} = \{ (w,a) \in \Theta^{(m)} \,|\, \forall i \in [m^*], \exists j \in [m], \lVert w_j - w^*_i \rVert_2 \leq \eta \}$.

Let us show that if $\theta \in \mathcal{P}_{\theta^*}^{(m)}$, then 
$\exists \nu \in \Theta^{(m)}$ such that $\lVert F(\theta) + \d F_{\theta} \cdot \nu - f^* \rVert_{\infty} \leq \sqrt{\varepsilon}$.

Let $m \in \mathbb{N}$ and $(w,a) \in \mathcal{P}_{\theta^*}^{(m)}$.
For all $i \in [m\FinDiff{^*}]$, define $j_i \in \argmin_{k \in [m]} \lVert w^*_i - w_k \rVert_{2}$.
\FinDiff{In words, $j_i \in [m]$ is the index of the (learned) neuron closest to target neuron $i \in [m^*]$}.
Then, let $\nu_0 = (-a_k + \sum_{i \in [m\FinDiff{^*}]} \delta_{k=j_i} a_i^* , 0)_{k \in [m]} \in \Theta^{(m)}$.
Observe that for $x \in \mathbb{R}^d$,

\[\begin{aligned}
(F(w,a) + \d F_{(w,a)} \cdot \nu_0)(x)
&= \sum_{k \in [m]} a_k \sigma(w_k \cdot x) + \sum_{k \in [m]} \left(- a_k + \sum_{i \in [m\FinDiff{^*}]} \delta_{k = j_i} a_i^* \right) \sigma(w_k \cdot x)
\\ &= \sum_{i \in [m\FinDiff{^*}]} a_i^* \, \sigma(w_{j_i} \cdot x)
\end{aligned}\]

Therefore, \RevDiff{using the Lipschitz property of $\sigma$, then $\lVert w_{j_i} - w_i^* \rVert_2 \leq \eta$},
\[\begin{aligned}
\lVert F(w,a) + \d F_{(w,a)} \cdot \nu_0 - f^* \rVert_\infty
& \leq \lVert F(w,a) + \d F_{(w,a)} \cdot \nu_0 - F(w^*, a^*) \rVert_\infty + \lVert F(w^*, a^*) - f^* \rVert_\infty
\\ &\leq \sup_{x \in K} \left| \sum_{i \in [m\FinDiff{^*}]} a_i^* \, \left( \sigma(w_{j_i} \cdot x) - \sigma(w^*_i \cdot x) \right) \right| + \frac{\sqrt{\varepsilon}}{2}
\\ &\leq  \sum_{i \in [m\FinDiff{^*}]} \left| a_i^* \right| \, L_\sigma \lVert w_{j_i} - w_i^* \rVert_{2} \sup_{x \in K} \lVert x \rVert_2 + \frac{\sqrt{\varepsilon}}{2}
\\ &\leq  \lVert a^* \rVert_1 \, L_\sigma \frac{\sqrt{\varepsilon}}{2 \lVert a^* \rVert_1 L_\sigma D} D + \frac{\sqrt{\varepsilon}}{2}
\\ &\leq \sqrt{\varepsilon}
\end{aligned}\]

Moreover, observe that
$ \lVert \nu_0 \rVert_2 \leq \lVert -a \rVert_2 + \lVert a^* \rVert_2 = \lVert a \rVert_2 + \lVert a^* \rVert_2 $\RevDiff{.}

Then, similarly to the linear cases, define the functional quadratic loss $\ell : f \mapsto \lVert f - f^* \rVert_\mathcal{D}^2$, which satisfies the Polyak-\Loja{} inequality $\lVert \nabla \ell_f \rVert_\mathcal{D}^2 \geq 4 \ell(f)$. It remains to transfer it to $\mathcal{L}$.
Unfortunately, we will not be able to lower-bound the Rayleigh quotient by a constant, so we perform a slightly different manipulation to obtain a Kurdyka-\Loja{} inequality on $\mathcal{L}$ \FinDiff{anyway}.

\[\begin{aligned}
\lVert \nabla \mathcal{L}(w, a) \rVert_2^2
&= \mathrm{R}\left( \overbar{K_\theta}; \nabla \ell_{F(w,a)}, \nabla \ell_{F(w,a)} \right) \lVert \nabla \ell_{F(w,a)} \rVert_\mathcal{D}^2
\\ &= \sup_{\nu \in \Theta^{(m)} \setminus \{0\} } \mathrm{R}\left( \overbar{\d F_{(w,a)}}; \nu, \nabla \ell_{F(w,a)} \right)^2 \lVert \nabla \ell_{F(w,a)} \rVert_\mathcal{D}^2
\\ &= \sup_{\nu \in \Theta^{(m)} \setminus \{0\} } \frac{ \langle \d F_\theta \cdot \nu, \nabla \ell_{f} \rangle_\mathcal{D}^2}{ \lVert \nu \rVert_2^2}
\\ &\geq \frac{ \langle \d F_\theta \cdot \nu_0, \, 2(F(\theta) - f^*) \rangle_\mathcal{D}^2}{ \lVert \nu_0 \rVert_2^2}
\end{aligned}\]
These computations are similar to the other cases, but since we're unable to obtain a lower bound multiplicatively by bounding the Rayleigh quotient directly, we instead split it to accept an additive $\varepsilon$ term (leading to convergence to $\varepsilon$ instead of convergence to zero as in the other examples).

\[\begin{aligned}
\lVert \nabla \mathcal{L}(w, a) \rVert_2^2
&\geq \frac{\FinDiff{1}}{ \lVert \nu_0 \rVert_2^2} \left( \lVert F(\theta) + \d F_\theta \cdot \nu_0 - f^* \rVert_\mathcal{D}^2 - \lVert \d F_\theta \cdot \nu_0 \rVert_\mathcal{D}^2 - \lVert F(\theta) - f^* \rVert_\mathcal{D}^2 \right)^2
&\FinDiff{(1)}
\\ &= \frac{\FinDiff{1}}{ \lVert \nu_0 \rVert_2^2} \left( \mathcal{L}(\theta) + \lVert \d F_\theta \cdot \nu_0 \rVert_\mathcal{D}^2 - \lVert F(\theta) + \d F_\theta \cdot \nu_0 - f^* \rVert_\mathcal{D}^2 \right)^2
\\ &\geq \frac{\FinDiff{1}}{ \lVert \nu_0 \rVert_2^2} \left( \mathcal{L}(\theta) + 0 - \varepsilon \right)_\RevDiff{+}^2
&\RevDiff{(2)}
\\ &\geq \frac{\FinDiff{1}}{\RevDiff{ \left(\lVert a \rVert_2 + \lVert a^* \rVert_2 \right)^2 }} \left( \mathcal{L}(\theta) - \varepsilon \right)_\RevDiff{+}^2
\end{aligned}\]

\FinDiff{Where $(1)$ is the parallelogram identity for the $\ell_2$ norm, $2 \langle u, v \rangle = \lVert u + v \rVert_2^2 - \lVert u \rVert_2^2 - \lVert v \rVert_2^2$, and}
\RevDiff{where $(2)$ is $(u \geq v \geq 0) \Rightarrow (u^2 \geq v^2)$ when $\mathcal{L}(\theta) \geq \varepsilon$.}
This almost concludes the first part of the proof (the inequality), \RevDiff{though it remains to show that  $\left(\lVert a \rVert_2 + \lVert a^*\rVert_2 \right)$ is bounded by $(\lVert a - a_0 \rVert_2 + C)$ for some constant $C \in \mathbb{R}_+^*$ independent of $m$}, with high probability. For reasons that will become apparent later, let $\delta_0 = \delta / 2 \in \,]0,1[$.
We now focus on the high probability part of the proof.

We have proved so far that under some condition on $\theta$, $\mathcal{L}$ satisfies a Kurdyka-\Loja{} inequality at $\theta$. It remains to prove that this condition is satisfied with high probability near initialization.
For any $\theta_0 \in \Theta^{(m)}$, let $\mathcal{B}(\theta_0, R) = \{ \theta \in \Theta^{(m)} \,|\, \lVert \theta - \theta_0 \rVert_2 \leq R \}$ be the $R$-radius ball around $\theta_0$.
We would like to show that for some $m \in \mathbb{N}$, it holds
\[ \mathbb{P}_{\theta_0 \sim \mathcal{I}_m} \left( \mathcal{B}(\theta_0, R) \subseteq \mathcal{P}_{\theta^*}^{(m)} \right)  \geq 1 - \delta_0 \]
To prove this statement, we will need a stronger property than just $\theta_0 \in \mathcal{P}_{\theta^*}^{(m)}$ with high probability. Namely, let $\mathcal{Q}_{\theta^*}^{(m)} = \{ \theta \in \Theta^{(m)} \,\mid\, \forall i \in [m^*], |\{ j \in [m], \lVert w_j - w_i^* \rVert_2 \leq \frac{\eta}{2} \}| \geq k \}$ be the set of parameters such that there are at least $k$ neurons in each (half smaller) feature bassin, for some yet unspecified value of $k \,\FinDiff{\in \mathbb{N}^*}$. In the set $\mathcal{P}_{\theta^*}$ we only required that $k=1$ and allow\FinDiff{ed} larger bassins.

Let $(H_u \subseteq [m])_{u \in [k]}$ be any partition of $[m]$ into $k$ sets, each of size at least $\lfloor m / k \rfloor$.
\FinDiff{For $(2)$ hereafter, note that if a set $S \subseteq [m]$ has size $\lvert S \rvert < k$, then $\exists u \in [k], S \cap H_u = \emptyset$, by the pigeonhole principle.}

\[\begin{aligned}
\mathbb{P}_{\theta_0 \sim \mathcal{I}_m} \left( \theta_0 \notin \mathcal{Q}_{\theta^*}^{(m)} \right)
&= \mathbb{P}_{(w,a) \sim \mathcal{I}_m} \left( \exists i \in [m^*], \left| \left\{ j \in [m], \lVert w_j - w_i^* \rVert_2 \leq \frac{\eta}{2} \right\} \right| < k \right)
\\ &\leq \sum_{i=1}^{m^*} \mathbb{P}\left( \left| \left\{ j, \lVert w_j - w_i^* \rVert_2 \RevDiff{\leq} \frac{\eta}{2} \right\} \right| < k \right) &(1)
\\ &\leq \sum_{i=1}^{m^*} \mathbb{P}\left( \exists u \in [k], \forall j \in H_u,  \lVert w_j - w_i^* \rVert_2 > \frac{\eta}{2} \right) &\FinDiff{(2)}
\\ &\leq \sum_{i=1}^{m^*} \sum_{u \in [k]} \prod_{j \in H_u} \mathbb{P}\left(\lVert w_j - w_i^* \rVert_2 > \frac{\eta}{2} \right) &(\FinDiff{3})
\\ &\leq \sum_{i=1}^{m^*} k \, {\left(\mathbb{P}_{y \sim \mathcal{N}(0_d,1_d)} \left( \lVert y - w_i^* \rVert_2 > \frac{\eta}{2} \right)\right)}^{\lfloor m / k \rfloor}
\end{aligned}\]

Where $(1)$ and $(\FinDiff{3})$ are union bounds, followed by independent identical distribution of $w_{j}$.

For all $i \in [m^*]$, it holds $\mathbb{P}_{y \sim \mathcal{N}(0_d,1_d)} \left( \lVert y - w_i^* \rVert > \eta/2 \right) < 1$ (i.e.\ full support), therefore for \FinDiff{any} fixed constant $k$, there exists an $m$ sufficiently large such that it holds $\mathbb{P}_{\theta_0 \sim \mathcal{I}_m}(\theta_0 \notin \mathcal{Q}_{\theta^*}^{(m)}) \leq \delta_0$.

Let $\theta_0 \in \mathcal{Q}_{\theta^*}^{(m)}$.
Let us show that $\mathcal{B}(\theta_0, R) \subseteq \mathcal{P}_{\theta^*}^{(m)}$.
Let $\theta \in \mathcal{B}(\theta_0, R)$, and $i \in [m^*]$.
We write $(w^{(0)}, a^{(0)}) = \theta_0$ the two components of $\theta_0$.
By assumption, there is a subset $J \subseteq [m]$ of size $|J| = k$ such that $\forall j \in J, \lVert w^{(0)}_j - w_i^* \rVert_2 \leq \frac{\eta}{2}$.

\[\begin{aligned}
\min_{j \in [m]} \lVert w_j - w_i^* \rVert_2
   &\leq \min_{j \in J} \lVert w_j - w_i^* \rVert_2
    \leq \frac{1}{k} \sum_{j \in J} \lVert w_j - w_i^* \rVert_2
\\ &\leq \frac{1}{k} \sum_{j \in J} \lVert w_j^{(0)} - w_i^* \rVert_2 + \lVert w_j - w_j^{(0)} \rVert_2
\\ &\leq \frac{\eta}{2} + \frac{1}{k} \sum_{j \in J} \lVert w_j - w_j^{(0)} \rVert_2
\\ & \RevDiff{\leq \frac{\eta}{2} + \sqrt{\frac{1}{k} \sum_{j \in J} \lVert w_j - w_j^{(0)} \rVert_2^2} }
\\ &\leq \frac{\eta}{2} + \frac{R}{\RevDiff{\sqrt{k}}}
\end{aligned}\]

Thus if $\RevDiff{\sqrt{k}} \geq 2 R / \eta$, it holds that $\forall m, \, \theta_0 \in \mathcal{Q}_{\theta^*}^{(m)} \Rightarrow \mathcal{B}(\theta_0, R) \subseteq \mathcal{P}_{\theta^*}^{(m)}$.
In particular, there exists $m$ such that $\mathbb{P}_{\theta_0 \sim \mathcal{I}_m}\left(\theta_0 \in \mathcal{Q}_{\theta^*}^{(m)} \right) \geq 1 - \delta_0$, thus $\mathbb{P}_{\theta_0 \sim \mathcal{I}_m}\left( \mathcal{B}(\theta_0, R) \subseteq \mathcal{P}_{\theta^*}^{(m)} \right) \geq 1 - \delta_0$, as claimed.

\RevDiff{As previously noted, it remains to show that $(\lVert a \rVert_2 + \lVert a^* \rVert_2) \leq \lVert a - a^{(0)} \rVert_2 + C$, for a constant $C$ independent of $m$. Let $C = \sqrt{1/\delta_0} + \lVert a^* \rVert_2$. The norm $\lVert a^* \rVert_2$ depends on $\varepsilon$, and thus the ``optimal'' number of neurons $m^*$, but not on the number of ``training'' neurons $m$.}
To reach the conclusion, let us show that $\mathbb{P}_{\theta_0 \sim \mathcal{I}_m}\left( \sup_{(w,a) \in \mathcal{B}(\theta_0, R)} \RevDiff{ \lVert a \rVert_2 - \lVert a - a^{(0)} \rVert_2 \leq \sqrt{1/\delta_0} } \right) \geq 1 - \delta_0$.

\[\begin{aligned}
\mathbb{P}_{\theta_0 \sim \mathcal{I}_m} &\left( \sup_{(w,a) \in \mathcal{B}(\theta_0, R)} \RevDiff{ \lVert a \rVert_2 - \lVert a - a^{(0)} \rVert \geq \sqrt{\frac{1}{\delta_0}} } \right)
\\ &\leq \mathbb{P}_{a^{(0)} \sim \mathcal{N}\left(0_m, I_m / \sqrt{m}\right)} \left( \sup_{a \in \mathcal{B}(a^{(0)}, R)} \RevDiff{ \lVert a \rVert_2 - \lVert a - a^{(0)} \rVert_2 \geq \sqrt{\frac{1}{\delta_0}} } \right)
\\ &\leq \mathbb{P}_{a^{(0)} \sim \mathcal{N}\left(0_m, I_m / \sqrt{m}\right)} \left( \RevDiff{ \lVert a^{(0)} \rVert_2 \geq \sqrt{\frac{1}{\delta_0}} } \right) &\FinDiff{(1)}
\\ &= \RevDiff{ \mathbb{P}_{a^{(0)} \sim \mathcal{N}\left(0_m, I_m / \sqrt{m} \right)} \left( \RevDiff{ \sum_{i \in [m]} \left( a_i^{(0)} \right)^2 \geq \frac{1}{\delta_0} } \right) }
\\ &\leq \RevDiff{ \frac{ \mathbb{E}_{a^{(0)}} \left[ \sum_{i \in [m]} \left( a_i^{(0)} \right)^2 \right] }{ 1 / \delta_0 }}
= \RevDiff{ \delta_0 \sum_{i \in [m]} \frac{1}{m} =\delta_0 }
& \RevDiff{ (2) }
\end{aligned}\]

\FinDiff{Where $(1)$ is because $\lVert a^{(0)}\rVert_2 \geq \lVert a \rVert_2 - \lVert a - a^{(0)} \rVert_2$ by triangular inequality, therefore for all constants $M \in \mathbb{R}_+^*$, it holds
$ \{ \lVert a \rVert_2 - \lVert a - a^{(0)} \rVert_2 \geq M \} \subseteq \{ \lVert a^{(0)} \rVert_2 \geq M \} $, and}
\RevDiff{where \FinDiff{$(2)$} is Markov's inequality.}
Now, tying all pieces together,
\[\begin{aligned}
& \mathbb{P}_{\theta_0 \sim \mathcal{I}_m} \left( \forall \theta \in \mathcal{B}(\theta_0, R), \lVert \nabla \mathcal{L}(\theta) \rVert_2^2 \geq \RevDiff{\frac{1}{\left( \lVert \theta - \theta_0 \rVert_2 + C \right)^2}} \left( \mathcal{L}(\theta) - \varepsilon \right)_\RevDiff{+}^2 \right)
\\ &\geq \mathbb{P}_{\theta_0 \sim \mathcal{I}_m} \left(\left( \theta_0 \in  \mathcal{Q}_{\theta^*}^{(m)} \right) \cap \left( \sup_{(w,a) \in \mathcal{B}(\theta_0, R)} \RevDiff{\lVert a \rVert_2 - \lVert a- a^{(0)} \rVert_2 \leq \sqrt{\frac{1}{\delta_0}} } \right) \right)
\\ &\RevDiff{\geq 1
- \mathbb{P}_{\theta_0 \sim \mathcal{I}_m} \left( \theta_0 \notin  \mathcal{Q}_{\theta^*}^{(m)} \right) - \mathbb{P}_{(w^{(0)},a^{(0)}) \sim \mathcal{I}_m} \left( \sup_{a \in \mathcal{B}(a^{(0)}, R)} \lVert a \rVert_2 - \lVert a - a^{(0)} \rVert_2 > \sqrt{\frac{1}{\delta_0}} \right) }
\\ &\RevDiff{\geq 1 - \delta_0 - \delta_0}
= 1 - \delta
\end{aligned}\]

This completes the proof that the Kurdyka-\Loja{} inequality holds on a ball near the initialization with high probability over the initialization when the number of neurons is sufficiently large.
%
%
\end{proof}

\RevDiff{
The idea for the second part of the proof is to put the Kurdyka-\Loja{} inequality in separable form, then integrate it (following \citet[Proposition 4.6]{scaman22a}, but we will reproduce the proof for shortness).
This will yield one upper bound on the loss if the weights remain in the ball, and an other bound on the loss if the weights escape the ball, which we can force into coinciding with the desired precision by adjusting the chosen radius $R$.
}

\RevDiff{
\begin{proof}[Proof of Proposition~\ref{prop:relu-loja} (b)]
Part (a) of this proof has established the following proposition w.h.p:
\[ \exists c \in \mathbb{R_+}, \forall R \in \mathbb{R}_+, \exists m \in \mathbb{N}^*, 
\forall \theta \in \mathcal{B}(\theta_0, R), \quad \lVert \nabla \mathcal{L}(\theta) \rVert_\Theta^2 \geq  \frac{\left( \mathcal{L}(\theta) - \varepsilon \right)_+^2}{(\lVert \theta - \theta_0 \rVert_2 + c)^2}  \]
Where the probability is taken over initializations $\theta_0 \sim \mathcal{I}_m$. Moreover, with high probability as well, $\mathcal{L}(\theta_0) \leq L_0 \in \mathbb{R}_+$ (independently of $m \in \mathbb{N}$, see Lemma~\ref{lemma:bounded-loss} below for details).

Let $\varepsilon_0 \in \mathbb{R}_+^*$ be any target precision.
Let \FinDiff{$R > 2 c L_0 / \varepsilon_0$} and apply Proposition~\ref{prop:relu-loja} (a) with $\varepsilon = \frac{\varepsilon_0}{2}$.

Let $\theta : \mathbb{R}_+ \to \Theta$ be a gradient flow of $\mathcal{L}$ with $\theta(0) = \theta_0$.
Since $t \mapsto \mathcal{L}(\theta_t)$ is a non-negative non-increasing function of time, it must converge to a non-negative real value $\mathcal{L}(\theta_t) \to_t \eta \in \mathbb{R}_+$ (by monotone convergence). Therefore, let us show that it will reach a loss below $\varepsilon_0$, which is sufficient to obtain $\eta \leq \varepsilon_0$.
If $\mathcal{L}(\theta_0) \leq \varepsilon_0$ then the proof is concluded, otherwise
let us define $T = \inf \left( \{ t \in \mathbb{R}_+ \mid \theta_t \in \mathcal{B}(\theta_0, R) \} \cap \{ t \in \mathbb{R}_+ \mid \mathcal{L}(\theta_t) \geq \varepsilon_0 \} \right) \in \mathbb{R}_+^* \cup \{ + \infty \}$.
We will now focus our attention to the interval $I = [0,T[$, where the Kurdyka-\Loja{} inequality is satisfied (by definition of $T$).
We start by weakening the inequality to get rid of $\lVert \theta - \theta_0 \rVert_2$ by separability.

Define $r : [0,T[ \to \mathbb{R}_+$, as $r : t \mapsto \int_0^t \lVert \partial_t \theta(u) \rVert \d u$. Observe that for all $t < T$, it holds $\lVert \theta_t - \theta_0 \rVert \leq r_t$ by triangular inequality. Additionally, using the square root of the Kurdyka-\Loja{} inequality,
\[\begin{aligned}
\partial_t r_t = \lVert \partial_t \theta \rVert_2 = \lVert \nabla \mathcal{L}(\theta_t) \rVert_2 = \frac{ \lVert \nabla \mathcal{L}(\theta_t) \rVert_2^2}{\lVert \nabla \mathcal{L}(\theta_t) \rVert_2} \leq \frac{\lVert \nabla \mathcal{L}(\theta_t) \Vert_2^2}{\frac{\FinDiff{1}}{r_t + c} \left(\mathcal{L}(\theta_t) - \varepsilon \right)}
= (r_t + c) \frac{ -\partial_t \mathcal{L}(\theta) }{\mathcal{L}(\theta_t) - \varepsilon}
\end{aligned}\]
This corresponds to the inequality $\partial_t (\psi \circ r) \leq \partial_t (\varphi \circ \mathcal{L})$, with desingularizers $\varphi : u \mapsto - \log( u - \varepsilon)$ and $\psi : u \mapsto \log(u + c)$.
Integrating between $0$ and $t < T$, this yields the inequality
\[
\log\left( \frac{r_t + c}{c} \right) = \left[ \log\left( r_u + c \right) \right]_0^t \leq \Bigl[ - \log\left( \mathcal{L}(\theta_u) - \varepsilon \right) \Bigr]_0^t = \log\left( \frac{ \mathcal{L}(\theta_0) - \varepsilon}{\mathcal{L}(\theta_t) - \varepsilon } \right)
\]
\begin{equation}\label{eq:radius-loss-bound}
r_t + c \leq c \, \frac{\mathcal{L}(\theta_0) - \varepsilon}{\mathcal{L}(\theta_t) - \varepsilon}
\end{equation}

Define the $\varepsilon$-discounted loss $\mathcal{L}^{\varepsilon} : \Theta \to \mathbb{R}_+$ as $\mathcal{L}^\varepsilon : u \mapsto \left( \mathcal{L}(u) - \varepsilon \right)_+$.
For all $t < T$, it holds $\mathcal{L}(\theta_t) = \mathcal{L}^\varepsilon(\theta_t) + \varepsilon$, thus
$\nabla \mathcal{L}(\theta_t) = \nabla \mathcal{L}^\varepsilon(\theta_t)$.
Therefore the restriction $\theta_I : [0,T[ \to \Theta$ is a gradient flow of $\mathcal{L}^\varepsilon$.
Moreover, injecting inequality $(\ref{eq:radius-loss-bound})$ above into the previous Kurdyka-\Loja{} inequality, we get the more easily understood inequality
\[
\forall t < T, \quad \lVert \nabla \mathcal{L}^\varepsilon(\theta_t) \rVert_2^2 \geq \frac{\FinDiff{1}}{ {\left( c \, \mathcal{L}^\varepsilon(\theta_0) \right)}^\FinDiff{2}} \left(\mathcal{L}^\varepsilon(\theta_t) \right)^\FinDiff{4}
\]
Setting $\kappa = \FinDiff{3 / (c \, \mathcal{L}^\varepsilon(\theta_0))^2 } \in \mathbb{R}_+^*$, and the desingularizer $\varphi : u \mapsto -1 / u^\FinDiff{3}$, this corresponds to the inequality $\d \varphi_{\mathcal{L}^\varepsilon} \lVert \nabla \mathcal{L}^\varepsilon \rVert_2^2 \geq \kappa$.
Integrating between $0$ and $T$ according to Proposition~\ref{prop:KLoja-direct}, this gives at all times $t < T$, the inequality
$\mathcal{L}^\varepsilon(\theta_t) \leq ( \mathcal{L}^\varepsilon(\theta_0)^{-\FinDiff{3}} + \kappa t )^{-1/\FinDiff{3}}$.
We are now ready to conclude by case disjunction.
If $T = + \infty$ then it is immediate that the convergence speed holds for $t \in \mathbb{R}_+$.
If $T < + \infty$, there are two cases to tackle.
If $\mathcal{L}(\theta_T) \leq \varepsilon_0$, then since the loss is decreasing, it holds for all $t \geq T$ that $\mathcal{L}(\theta_t) \leq \mathcal{L}(\theta_T) \leq \varepsilon_0$, therefore the bound holds for $t \geq T$ as well which concludes this case.
If $\lVert \theta_T - \theta_0 \rVert_2 = R$, then equation $(\ref{eq:radius-loss-bound})$ gives
\FinDiff{$\mathcal{L}(\theta_T) - \varepsilon \leq  \mathcal{L}(\theta_0) \, c / (R + c) < \varepsilon$}
(by definition of $R$), thus $\mathcal{L}(\theta_T) \leq \varepsilon + \varepsilon = \varepsilon_0$, and therefore by the same argument, it holds for $t \geq T$ that $\mathcal{L}(\theta_t) \leq \mathcal{L}(\theta_T) \leq \varepsilon_0$ and thus the bound is extended to $t \in \mathbb{R}_+$, which concludes the proof.
\end{proof}
}
\endgroup

\pagebreak
\RevDiff{
\begin{lemma}[Bounded initial loss with high probability]\label{lemma:bounded-loss}
Under the hypotheses of Proposition~\ref{prop:relu-loja}, for all $\delta \in ]0,1[$, there exists $L_0 \in \mathbb{R}_+$, such that for all $m \in \mathbb{N}^*$, it holds $\mathbb{P}_{\theta_0}\left( \mathcal{L}(\theta_0) \leq L_0 \right) \geq 1 - \delta$.
\end{lemma}

\begin{proof}
For an $m \in \mathbb{N}^*$, let $a_i \sim \mathcal{N}(0, 1 / \sqrt{m})$ for $i \in [m]$ and $w_{i,j} \sim \mathcal{N}(0, 1)$ for $(i,j) \in [m] \times [d]$ be independent random variables, so that $(a,w) \sim \mathcal{I}_m$. We wish to prove that $\left(\mathcal{L}(a, w) \leq L_m\right)$ holds with high probability for some constant $L_m \in \mathbb{R}_+$.
First, observe that

\[
\mathcal{L}(a,w) = \mathbb{E}_{x \sim \mathcal{D}} \left[ \left(\sum_{i \in [m]} a_i \sigma(w_i \cdot x) -  f^*(x) \right)^2 \right]
= \lVert f_\theta - f^* \rVert_\mathcal{D}^2 \leq 2 \left( \lVert f_\theta \rVert_\mathcal{D}^2 + \lVert f^* \rVert_\mathcal{D}^2 \right)
\]

Since $\lVert f^* \rVert_\mathcal{D}^2$ is a constant, it is sufficient to show that $\lVert f_\theta \rVert_\mathcal{D}^2$ is bounded with high probability.

In order to proceed by Markov's inequality, let us show that $\mathbb{E}_{a,w} \left[ \mathbb{E}_x \left[ \left( \sum_i a_i \sigma(w_i \cdot x) \right)^2 \right] \right]$ is finite.
First, if $i \neq j$, then $\mathbb{E}_{a,w,x}\left[ a_i a_j \sigma(w_i x) \sigma(w_j x) \right] = \mathbb{E}[a_i] \mathbb{E}[a_j] \mathbb{E}\left[ \sigma(w_i \cdot x) \sigma(w_j \cdot x) \right]$ = 0, by independence of $a$ an $(w,x)$, and independence of $a_i$ and $a_j$.
Therefore

\[\begin{aligned}
\mathbb{E}_{a,w,x} \left[ \left( \sum_{i \in [m]} a_i \sigma(w_i \cdot x) \right)^2 \right]
&= \sum_{i \in [m]} \mathbb{E}_{a,w,x} \left[ a_i^2 \sigma(w_i \cdot x)^2 \right]
\\ &= \sum_{i \in [m]} \mathbb{E}_{a}\left[ a_i^2 \right] \mathbb{E}_{w,x} \left[ \sigma(w_i \cdot x)^2 \right]
&(1)
\\ &\leq \sum_{i \in [m]} \frac{1}{m} \mathbb{E}_{w,x} \left[ \left(\lvert \sigma(0) \rvert + L_\sigma \lvert w_i \cdot x\rvert \right)^2 \right]
&(2)
\\ &\leq \frac{1}{m}\sum_{i \in [m]} \mathbb{E}_{w,x} \left[ \left(\lvert \sigma(0) \rvert + L_\sigma \lVert w_i \rVert_2 \lVert x\rVert_2 \right)^2 \right]
&(3)
\\ &\leq \frac{1}{m}\sum_{i \in [m]} \mathbb{E}_{w,x} \left[ \left(\lvert \sigma(0) \rvert + L_\sigma \lVert w_i \rVert_2 D \right)^2 \right]
&(4)
\\ &\leq \frac{1}{m}\sum_{i \in [m]} 2 \left( \sigma(0)^2 + L_\sigma^2 D^2 \mathbb{E}_{w} \left[ \lVert w_i \rVert_2^2 \right] \right)
&(5)
\\ &= \frac{1}{m}\sum_{i \in [m]} 2 \left( \sigma(0)^2 + L_\sigma^2 D^2 \mathbb{E}_{w} \left[ \sum_j w_{i,j}^2 \right] \right)
\\ &= \frac{1}{m}\sum_{i \in [m]} 2 \left( \sigma(0)^2 + L_\sigma^2 D^2 d \right) = 2 \left( \sigma(0)^2 + L_\sigma^2 D^2 d \right)
\end{aligned}\]

Where $(1)$ is independence, $(2)$ is because $\sigma$ is $L_\sigma$-Lipschitz, $(3)$ is Cauchy-Schwarz, $(4)$ is bounded input radius $D = \sup_{x \in K} \lVert x \Vert_2$ by compact-support assumption, $(5)$ is $(u+v)^2 \leq 2 (u^2 + v^2)$, and the remaining is evaluation in closed form.

Let $K = 2 \left( \sigma(0)^2 + L_\sigma^2 D^2 d \right) / \delta \in \mathbb{R}_+^*$.
By Markov's inequality, $\mathbb{P}_\theta \left( \lVert f_\theta \rVert_\mathcal{D}^2 \geq K \right) \leq \mathbb{E}_\theta\left[ \lVert f_\theta \rVert_\mathcal{D}^2 \right] / K = \delta$.
This constant does not depend on $m$, therefore the choice of bound $L_0 = 2 \left( K + \lVert f^* \rVert_\mathcal{D}^2 \right)$ concludes the proof.
\end{proof}
}


\end{document}